%% file: main.tex
\documentclass[10pt]{article} 
\usepackage[preprint]{tmlr}

\input{math_commands.tex}

\usepackage{hyperref}
\usepackage{url}

\usepackage{graphicx}
\usepackage{wrapfig}
\usepackage{tcolorbox}
\usepackage{booktabs}
\usepackage{amsthm}
\usepackage{algorithm}
\usepackage{algpseudocode}
\usepackage{algcompatible}
\usepackage{float}
\usepackage{multirow}
\usepackage{amssymb}
\usepackage{dsfont}
\usepackage[toc,page,header]{appendix}
\usepackage{minitoc}

\title{What Matters in Hierarchical Search for Combinatorial \\ 
Reasoning Problems?}

\author{\name Michał Zawalski$^{*}$
    \email m.zawalski@uw.edu.pl \\
    \addr University of Warsaw \\
            UC Berkeley
    \AND
    \name Gracjan Góral$^{*}$
    \email gp.goral@uw.edu.pl  \\
    \addr University of Warsaw \\
            IDEAS NCBR
    \AND
    \name Michał Tyrolski \\
    \addr Independent Researcher
    \AND
    \name Emilia Wiśnios \\
    \addr Independent Researcher
    \AND
    \name Franciszek Budrowski \\
    \addr Independent Researcher
    \AND 
    \name Marek Cygan \email M.Cygan@mimuw.edu.pl \\
    \addr University of Warsaw \\
        Nomagic 
    \AND 
    \name Łukasz Kuciński \email lkucinski@mimuw.edu.pl\\
    \addr University of Warsaw \\
        IDEAS NCBR \\
        Institute of Mathematics, Polish Academy of Sciences 
    \AND 
    \name Piotr Miłoś \email pmilos@mimuw.edu.pl \\
    \addr University of Warsaw \\
        IDEAS NCBR \\
        Institute of Mathematics, Polish Academy of Sciences}

\newcommand{\FUNCTION}[2]{\State \textbf{function} \textsc{#1}(#2)}

\newcommand{\takeaways}[2]{%
  \begin{tcolorbox}[colback=gray!5!white, boxrule=0.2mm]
    \paragraph{Takeaway #2} \textit{#1}
  \end{tcolorbox}%
}

\newtheorem{theorem}{Theorem}
\newtheorem{lemma}{Lemma}

\newcommand{\BestFS}{$\rho$-BestFS}
\newcommand{\Astar}{$\rho$-A*}
\newcommand{\MCTS}{$\rho$-MCTS}

\def\month{MM}  
\def\year{YYYY} 
\def\openreview{\url{https://openreview.net/forum?id=XXXX}} 

\begin{document}

\doparttoc 
\faketableofcontents 

\part{} 

\maketitle

\input{content/abstract}
\input{content/introduction}
\input{content/related-work}
\input{content/method}
\input{content/experiments}
\input{content/conclusions}

\bibliography{main}
\bibliographystyle{tmlr}

\newpage
\appendix
\addcontentsline{toc}{section}{Appendix} 
\part{Appendix} 
\parttoc 
\input{content/appendix}

\end{document}

%% file: math_commands.tex
\usepackage{amsmath,amsfonts,bm}

\def\eqref#1{equation~\ref{#1}}

\def\1{\bm{1}}

\DeclareMathAlphabet{\mathsfit}{\encodingdefault}{\sfdefault}{m}{sl}
\SetMathAlphabet{\mathsfit}{bold}{\encodingdefault}{\sfdefault}{bx}{n}



%% file: content/abstract.tex
\begin{abstract}
Combinatorial reasoning problems, particularly the notorious NP-hard tasks, remain a significant challenge for AI research. A common approach to addressing them combines search with learned heuristics. Recent methods in this domain utilize hierarchical planning, executing strategies based on subgoals. Our goal is to advance research in this area and establish a solid conceptual and empirical foundation. 
Specifically, we identify the following key obstacles, whose presence favors the choice of hierarchical search methods: \emph{hard-to-learn value functions, complex action spaces, presence of dead ends in the environment,} or \emph{training data collected from diverse sources}. Through in-depth empirical analysis, we establish that hierarchical search methods consistently outperform standard search methods across these dimensions, and we formulate insights for future research. On the practical side, we also propose a consistent evaluation guidelines to enable meaningful comparisons between methods and reassess the state-of-the-art algorithms.
\end{abstract}

%% file: content/introduction.tex
\section{Introduction}

\begin{wrapfigure}[24]{r}{0.44\textwidth}
    \vspace{-14pt}
    \centering
    \includegraphics[width=0.95\linewidth]{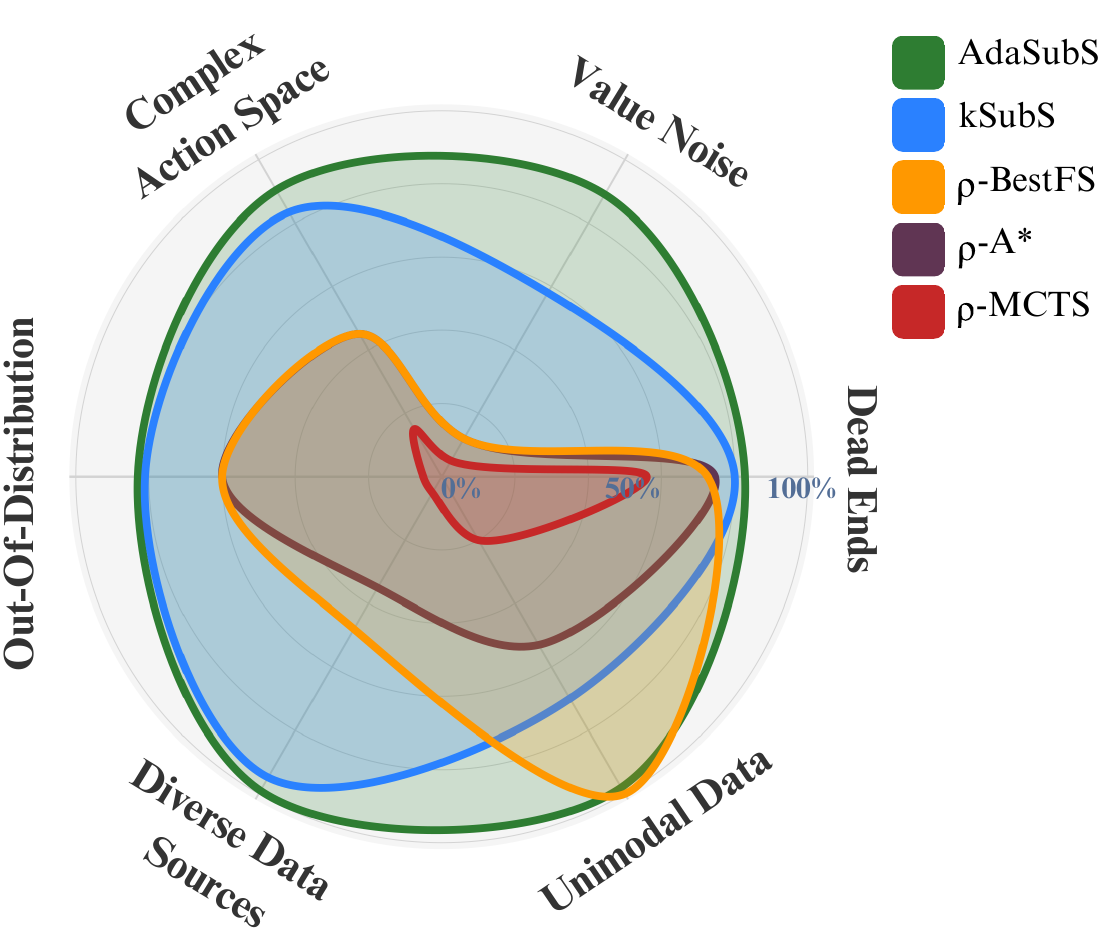} 
    \caption{Schematic performance comparison of hierarchical methods (AdaSubS, kSubS) and low-level methods (\BestFS, \Astar, \MCTS) across six dimensions: \textit{handling data collected from diverse sources}, \textit{learning from clean unimodal demonstrations}, \textit{avoiding dead ends}, \textit{performance under high value approximation errors}, \textit{handling complex action space}, and \textit{generalizing to out-of-distribution instances}.
    }
    \label{fig:center_image}
\end{wrapfigure}

The ability to solve discrete tasks that require sophisticated reasoning, particularly those involving \mbox{NP-hard} problems, is essential for advancing AI \citep{bengio_combinatorial_optimization}. 
These include complex problems like theorem proving \citep{int, AlphaGeometryTrinh2024}, constraint satisfaction problem \citep{DBLP:conf/icml/AchiamHTA17}, molecule alignment \citep{needleman1970general, smith1981identification}, social network analysis \citep{DBLP:conf/iclr/KipfW17}, or navigation \citep{lavalle2006planning, choset2005principles}.

Even driving a car, which typically involves continuous control of steering and speed, requires high-level discrete decision-making, e.g., when to overtake, when to change lanes, or how to navigate through traffic \citep{DBLP:journals/tits/KiranSTMSYP22}. 

Addressing that kind of tasks, known as combinatorial reasoning problems, requires efficient planning strategies due to the vast and complex search spaces involved \citep{DBLP:conf/nips/BruckG87}. A promising approach to this challenge, inspired by how humans plan their actions \citep{Hull1932, Fishbach2005, Kool2014}, is hierarchical search. This method breaks down a problem into manageable subproblems, or subgoals, making the overall task more tractable, in contrast to low-level methods that rely on atomic actions for planning. Hierarchical search has been successfully applied to a variety of combinatorial reasoning tasks, as evidenced by methods like Subgoal Search (kSubS) \citep{ksubs}, and further advanced by approaches such as Adaptive Subgoal Search (AdaSubS) \citep{ada}, Hierarchical Imitation Planning with Search (HIPS) \citep{hips}, and HIPS-$\varepsilon$ \citep{hipseps}.

Our goal in this paper is to advance research in hierarchical planning and establish a solid conceptual and empirical foundation. We identify four key challenges whose presence highly favors the use of hierarchical search methods: \emph{hard-to-learn value functions, complex action spaces, presence of dead ends in the environment, or data collected from diverse sources}. Through comprehensive empirical analysis, we demonstrate that hierarchical methods consistently outperform standard search techniques in overcoming these critical obstacles. Furthermore, we propose a consistent evaluation methodology to facilitate meaningful comparisons between methods and reassess current state-of-the-art algorithms. Our findings offer a clearer understanding of when hierarchical approaches should be preferred over low-level methods.

In summary, our contributions are as follows:%

\begin{itemize}
    \item We present a comprehensive empirical analysis comparing the performance of hierarchical search methods against low-level search methods across diverse problem settings.
    \item We identify problem characteristics that influence performance, providing insights into when hierarchical methods should be favored over low-level methods.
    \item We propose a standardized evaluation guidelines that facilitate meaningful and consistent comparisons across different types of search methods.
\end{itemize}

%% file: content/related-work.tex
\section{Related Work}
\paragraph{Solving Decision-Making Problems} Decision-making problems are often framed as Markov Decision Processes (MDPs) \citep{DBLP:journals/ai/SuttonPS99}, which can be solved using Reinforcement Learning (RL) algorithms like PPO \citep{schulman2017proximal} or DQN \citep{Mnih2015}. These methods learn policies through interaction with the environment. An alternative to learning from trial and error is Imitation Learning (IL), training models directly from offline demonstrations. The availability of large-scale datasets \citep{walke2023bridgedata, open_x_embodiment_rt_x_2023, grauman2022ego4d, Dosovitskiy17}, make it applicable to the most complex domains like robotics \citep{DBLP:conf/corl/MandlekarZGBSTG18, 8206196, kim24openvla}, autonomous driving \citep{8793698, li2022efficient, 10.5555/3298483.3298654}, and physics-based control \citep{pmlr-v119-kim20c, fickinger2022crossdomain}. Key foundational methods such as Behavioral Cloning (BC) \citep{DBLP:books/lib/SuttonB98}, Inverse Reinforcement Learning (IRL) \citep{baker2009action}, or DAgger \citep{dagger} have been instrumental in advancing IL for complex environments where direct exploration is less practical. In this work, we use IL to train components for the search methods, such as the policy and value function.

\paragraph{Subgoal Methods}
Hierarchical Reinforcement Learning methods tackle complex decision-making tasks by breaking them into subgoals. HIRO \citep{DBLP:conf/nips/NachumGLL18} reuses past data by goal relabeling. HAC \citep{DBLP:conf/iclr/LevyKPS19} builds a multi-layer hierarchy of policies trained with hindsight. Hierarchical Diffuser \citep{DBLP:journals/corr/abs-2401-02644} learns to predict future states with diffusion models. Graph-based methods, such as SoRB \citep{sorb} or DHRL \citep{DBLP:conf/nips/LeeKJK22} build a high-level graph of states, which then allow for efficient shortest path finding. GCP \citep{DBLP:conf/nips/PertschREZJFL20} learns to predict middle states between two given observations. Algorithms such as HPG \citep{DBLP:conf/icml/GhavamzadehM03} or H-DDPG \citep{DBLP:journals/tnn/YangMJA18} extend the classical RL algorithms to the hierarchical setting.

In the area of combinatorial reasoning, there has been growing interest in applying HRL techniques. kSubS \citep{ksubs} introduces a hierarchical search algorithm that iteratively generates subgoals to construct a search tree. Building on this, AdaSubS \citep{ada} incorporates multiple subgoal generators, each trained to predict subgoals at different distances from the target, allowing for dynamic adaptation of the planning horizon based on problem complexity. HIPS \citep{hips} and HIPS-$\varepsilon$ \citep{hipseps} perform search using subgoals generated by VQ-VAE models \citep{DBLP:conf/nips/OordVK17}.

\paragraph{Low-level Search Algorithms} Traditional search algorithms like Best-First Search (BestFS), A* \citep{10.5555/1614191, 10.5555/1671238}, and Monte Carlo Tree Search (MCTS) \citep{NIPS2009_389bc7bb, James_Konidaris_Rosman_2017} have long been the foundation for solving complex decision-making problems. Recent advancements have improved these methods by integrating neural network-based heuristics, improving their efficiency in large search spaces \citep{alphazero, DBLP:conf/icml/YonetaniTBNK21}. A variant of \BestFS{} used in \citep{ksubs, ada}, leverage heuristics learned through behavioral cloning to guide search. More recent algorithms, like PHS \citep{DBLP:conf/aaai/OrseauL21} or LevinTS \citep{ijcai2023p0624}, combine policy-driven and value-based approaches, offering both theoretical guarantees and strong empirical performance. Additionally, PDDL planners \citep{DBLP:series/synthesis/2019Haslum} solve decision-making problems by using predefined action models and goals, with domain-independent planners offering broad applicability, while domain-specific ones achieve higher performance in specialized tasks.

\paragraph{Empirical Studies on Algorithmic Performance}

Our work aligns with recent empirical studies that investigate the conditions under which various algorithmic approaches excel. For instance, \citep{what_matters} investigates how specific design choices influence the performance of PPO, while other research compares offline reinforcement learning with behavioral cloning \citep{ORL_vs_BC} or explores design choices for language-conditioned robotic imitation learning \citep{DBLP:journals/ral/MeesHB22}. In this paper, we focus on hierarchical search in combinatorial reasoning problems, specifically studying the conditions where hierarchical methods outperform low-level planners. To the best of our knowledge, this is the first systematic study of the relationship between hierarchical and low-level search in this context.

%% file: content/method.tex
\section{Combinatorial Environments}\label{sec:preliminaries_environments}

Our study targets solving combinatorial environments -- domains in which the number of possible configurations or decisions grows exponentially with the problem size, making them highly challenging to solve. This class includes several NP-hard problems, such as the Traveling Salesman Problem \citep{applegate2006traveling}, the Rubik's Cube \citep{singmaster1981notes}, Sokoban \citep{pspace_soko}, or solving non-linear inequalities \citep{DBLP:journals/siamcomp/Sahni74}. To efficiently solve combinatorial problems an algorithm should have the following key properties:

\begin{enumerate}
    \item \textbf{Learning from offline data.} Since combinatorial reasoning environments are characterized by a large space of possible configurations, learning without priors or handcrafted dense rewards is infeasible\footnote{For instance, we tested PPO \citep{schulman2017proximal}  on the Rubik's Cube, but, unsurprisingly, it failed to make any progress due to never reaching the goal in the haystack of $4.3\times 10^{19}$ states, hence never observing a positive reward. }
    Thus, the algorithm has to be able to learn from additional offline data, such as demonstrations.
    \item \textbf{Combinatorial space abstraction.} The space complexity significantly restricts the fraction of observable states. As a result, it is unrealistic to expect repeated visits to nearby states, an assumption that some approaches implicitly rely on.
    \item \textbf{Planning.} Methods that don’t use search and follow a single action trajectory are inherently limited by computational complexity, since they can perform only a constant number of operations before choosing an action. Solving NP-hard problems within a fixed computation budget is computationally infeasible \citep{DBLP:conf/nips/BruckG87}.
\end{enumerate}

Many hierarchical methods have not been designed for combinatorial problems, so they fail to meet the listed conditions and cannot be expected to be efficient in these applications. For instance, \citep{DBLP:journals/corr/abs-2401-02644, DBLP:journals/tnn/YangMJA18} require continuous state or action space, \citep{DBLP:conf/icml/GhavamzadehM03} learns only from online interactions, \citep{sorb, DBLP:conf/nips/HuangLS19, DBLP:conf/nips/LeeKJK22} assume a good coverage of the whole state space, and \citep{DBLP:conf/nips/NachumGLL18, DBLP:conf/iclr/LevyKPS19} do not use planning to determine actions.

\section{Subgoal Methods} \label{sec:preliminaries_subgoal_methods}

Subgoal methods, or hierarchical methods, are a family of algorithms designed to solve complex decision-making tasks by breaking down the overall objective into smaller, more manageable subgoals \citep{DBLP:journals/ai/SuttonPS99}. Instead of searching for a sequence of low-level actions that directly lead from the initial state to the goal, the agent first identifies high-level intermediate targets -- subgoals -- that guide the trajectory toward the final goal. The use of subgoals is widely considered as a method that scales better to longer horizons \citep{DBLP:journals/corr/abs-2401-02644, DBLP:conf/nips/LeeKJK22}, mitigates errors in value approximations \citep{ksubs}, and reduces overall complexity by decomposing the problem into smaller subproblems \citep{DBLP:journals/ai/SuttonPS99, ada}. The process of searching involves the following components:

\begin{itemize}
    \item \textbf{Subgoal generator} that, given a state within the search tree, outputs subgoals to be achieved. For instance, a subgoal may be a future state \citep{ksubs, ada} or a class of desired outcomes \citep{DBLP:conf/nips/JiangGMF19, DBLP:journals/corr/abs-1806-05292}. The generator is used by the planner to construct a search tree of subgoals.
    \item \textbf{Low-level policy} that  determines a path of low-level actions between subgoals. For instance, it may be a trained goal-reaching policy \citep{ksubs, ada}, a local search \citep{ksubs, hips}, or a stored path from previous episodes \citep{sorb, DBLP:conf/nips/LeeKJK22}.
    \item \textbf{Planner} that determines the order in which subgoals are generated. Standard planning algorithms like BestFS \citep{ksubs}, PHS \citep{hips}, or their modified forms \citep{ada}, are typically used.
    \item \textbf{Value function} that estimates the distance between the given state and the goal state. The planner uses this information to select the next node to expand with the subgoal generator. In some works it is also called \textit{heuristic value}.
\end{itemize}

In our experiments, we use kSubS \cite{ksubs} and AdaSubS \cite{ada} as subgoal methods well-suited for combinatorial problems, as they satisfy the conditions formulated in Section \ref{sec:preliminaries_environments}. We also experimented with HIPS and HIPS-$\varepsilon$ \citep{hips, hipseps}, but these methods generally fail to solve the problems within a reasonable computational budget. Therefore, their results are omitted from the main text and discussed in see Appendix \ref{appendix:hips}.

We compare the performance of the selected subgoal approaches against three popular low-level methods: BestFS, A*, and MCTS. To ensure a fair comparison and improve efficiency, we augment these algorithms by using a trained policy to select the top actions before each node expansion. We refer to them as \BestFS, \Astar, and \MCTS. A detailed description, analysis, and pseudocode for each of these algorithms can be found in Appendix~\ref{appendix:algorithms}. See also Appendix \ref{appendix:hierarchical_search} for diagrams explaining different search methods.

\subsection{Training Components}\label{sec:subgoal_setup}

In our experiments, the models for both subgoal methods and low-level searches were trained using imitation learning, following standard practice \citep{DBLP:conf/icra/NairMAZA18, ksubs}. Specifically, we collected a dataset of approximately $500\,000$ trajectories for each environment. Trajectories are sequences of consecutive states and actions leading to the goal state. We used various methods of dataset collection, like hand-crafted algorithms, trained policies, reversed random shuffles, and others, which let us to study the influence of training data characteristics on the performance of search methods.

To ensure a fair comparison, all methods shared common components whenever applicable (e.g., each method uses the same value function). This allows us to focus on the differences between the search algorithms, rather than heuristic biases. No additional heuristics were used, ensuring that performance differences arise solely from the algorithmic approaches.

More details on training the components, including specific objectives, are provided in Appendix \ref{appendix:pretraining}.

\subsection{Performance Metric}\label{sec:subgoal_metric}
Our performance metric is the \textit{success rate}, defined as the percentage of problem instances solved within a given \textit{complete search budget}. The complete search budget is the total number of visited states in the search tree. In particular, for subgoal methods, the budget includes both the generated subgoals and the states visited by the low-level policy used to connect these subgoals. 

By accounting for the total number of visited states, this metric provides a unified and fair comparison of search efficiency across different methods. We argue that reporting only the number of visited subgoal nodes would unfairly favor subgoal methods (see Appendix~\ref{appendix:hips} for details).

%% file: content/experiments.tex
\section{Analysis} \label{sec:experiments}

We investigate how environmental properties and training data influence the performance of hierarchical methods compared to low-level search approaches in combinatorial reasoning tasks. While previous works \citep{ksubs, ada, hips, hipseps} show a considerable advantage of hierarchical methods, our experiments reveal that this advantage is not consistent across all scenarios (see Figures \ref{fig:rubik_random20} or \ref{fig:rubik_beginner} for specific examples). Specifically, we answer the following research questions:

\newcommand{\rqperformance}{Q1}
\newcommand{\rqproperty}{Q2}
\newcommand{\rqpitfall}{Q3}

\begin{enumerate}
\item[Q1.] Is hierarchical search more effective than low-level search for solving combinatorial reasoning problems?
\item[Q2.] What environmental properties and characteristics of the training data amplify performance differences? When hierarchical search should be preferred over low-level search?
\item[Q3.] What pitfalls should be avoided when interpreting experimental results?
\end{enumerate}

To address these questions, we conducted a wide range of experiments comparing subgoal and low-level search algorithms across a variety of combinatorial reasoning tasks. Below, in each subsection we summarize the key findings that reveal the most significant factors affecting performance, followed by a brief discussion. For each finding, we link it to the relevant research questions. The extended analysis of these factors can be found in Appendix \ref{appendix:properties}.

We present our findings using the \textit{Rubik's Cube, Sokoban, N-Puzzle}, and \textit{Inequality Theorem Proving} (INT) \citep{int} environments. These classical benchmarks are widely used in planning research \citep{agnostelli, ksubs} and are known to be NP-hard \citep{Demain, pspace_soko, npuzzle}. Detailed descriptions of these environments can be found in Appendix~\ref{appendix:environments}.

All methods in our study were trained using imitation learning, with each approach sharing the same value function, as outlined in Section \ref{sec:subgoal_setup}.
To ensure fair comparisons, we measured complete search budgets, in contrast to counting only high-level search nodes, to avoid giving any unfair advantage to subgoal methods, as discussed in Section \ref{sec:subgoal_metric} (which contributes to the research question \rqpitfall).

\subsection{Subgoal Methods Benefit from Diverse Sources of Data}\label{sec:analysis_moe}

Achieving superhuman performance in complex tasks often involves large-scale datasets of demonstrations obtained from agents with varying skill levels and strategies \citep{alphago}. By training models on data collected from a variety of solvers and testing them in the Rubik’s Cube and N-Puzzle environments, we show that the variability in training data has a significant impact on the performance of search algorithms.

\begin{figure}[h]
    \centering
    \begin{minipage}[t]{0.45\linewidth}
        \includegraphics[width=\linewidth]{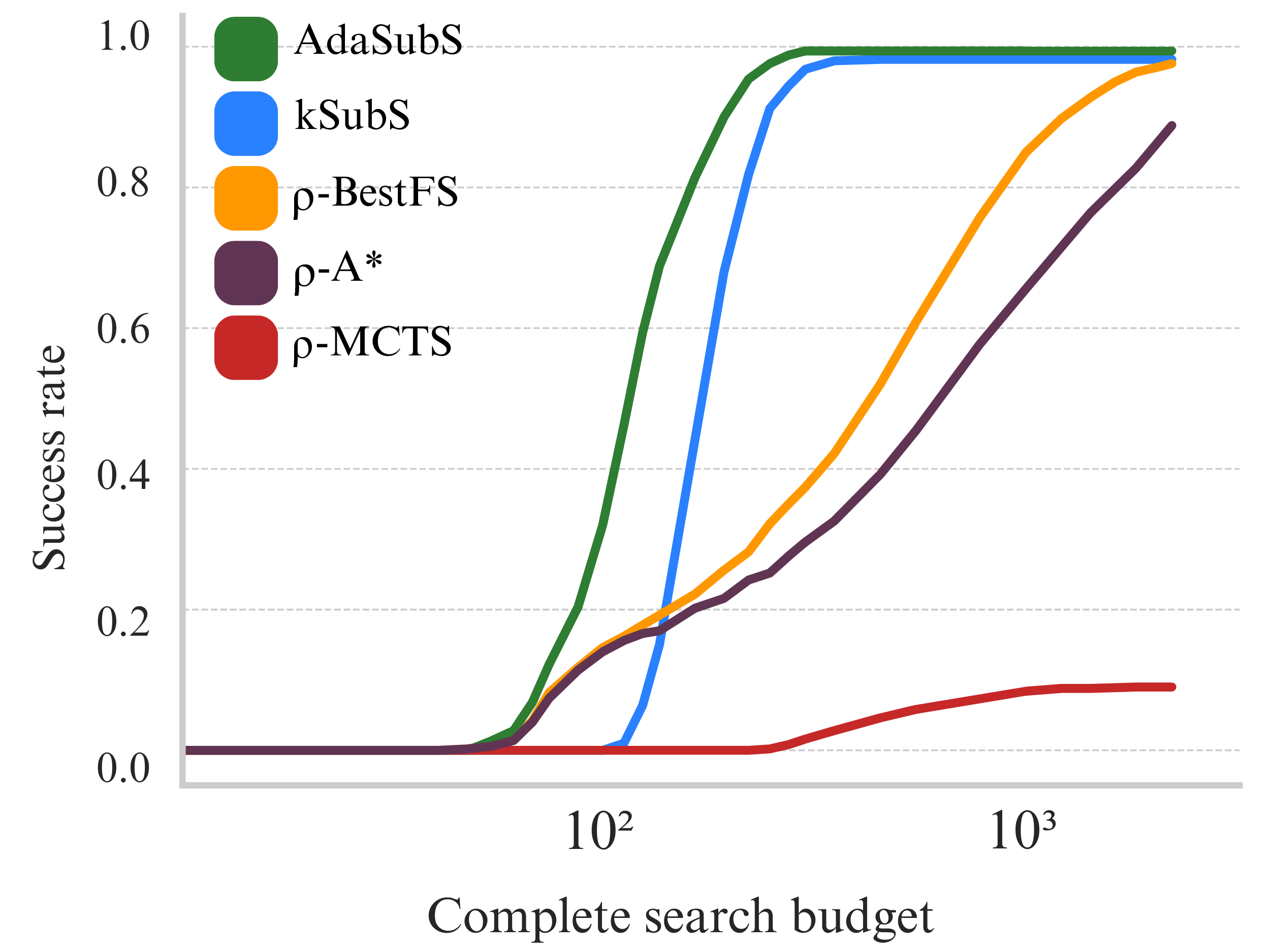}
        \caption{\small Solving the Rubik's Cube. Components are trained on data from 4 different solvers.}
        \label{fig:rubik_moe}
    \end{minipage}
    \hfill 
    \begin{minipage}[t]{0.45\linewidth}
        \includegraphics[width=\linewidth]{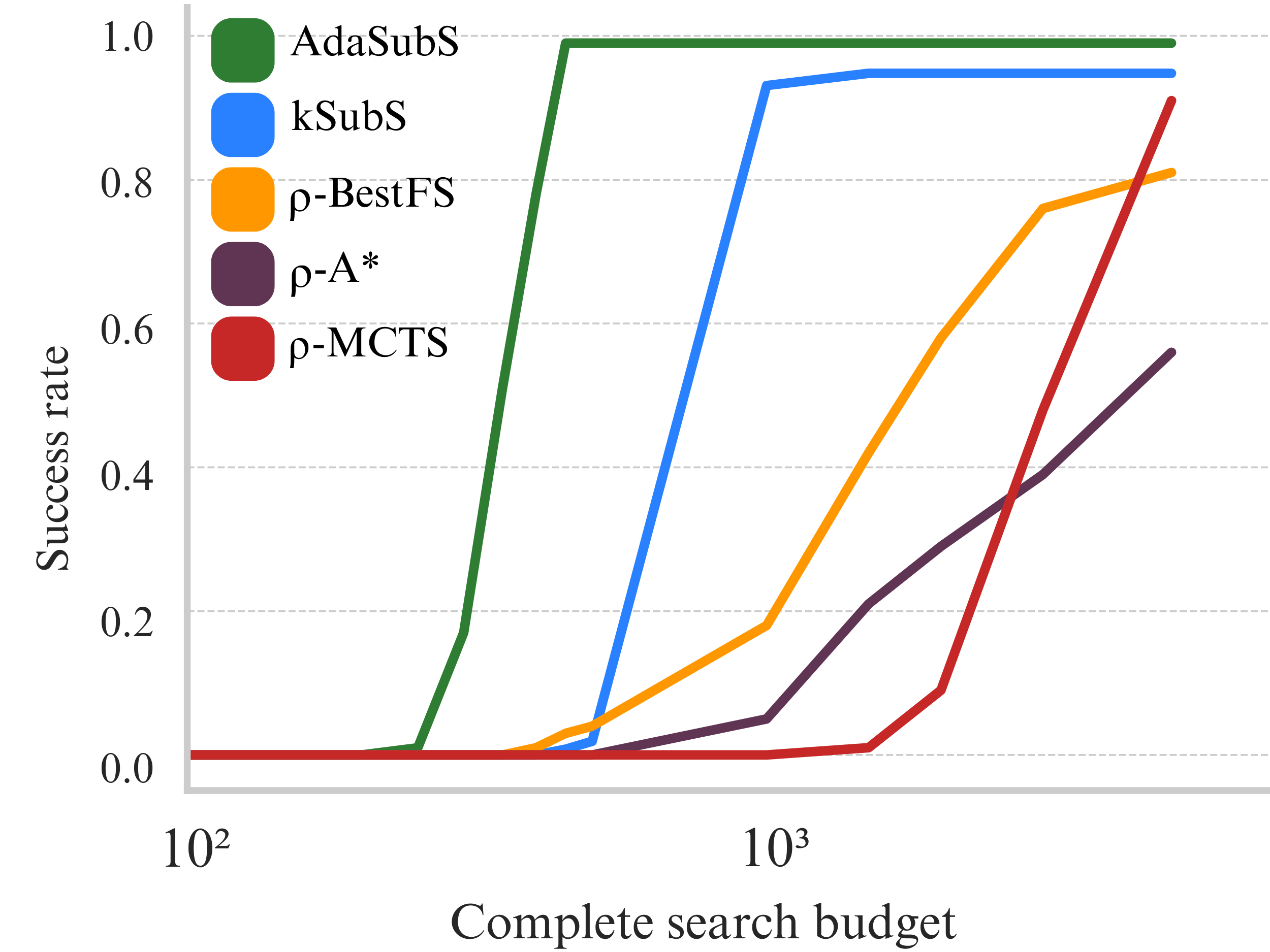}
        \caption{\small Solving the N-Puzzle. Components are trained on data from 2 different solvers.}
        \label{fig:npuzzle_moe}
    \end{minipage}
\end{figure}

\begin{figure}[h]
    \centering
    \begin{minipage}[t]{0.32\linewidth}
        \includegraphics[width=\linewidth]{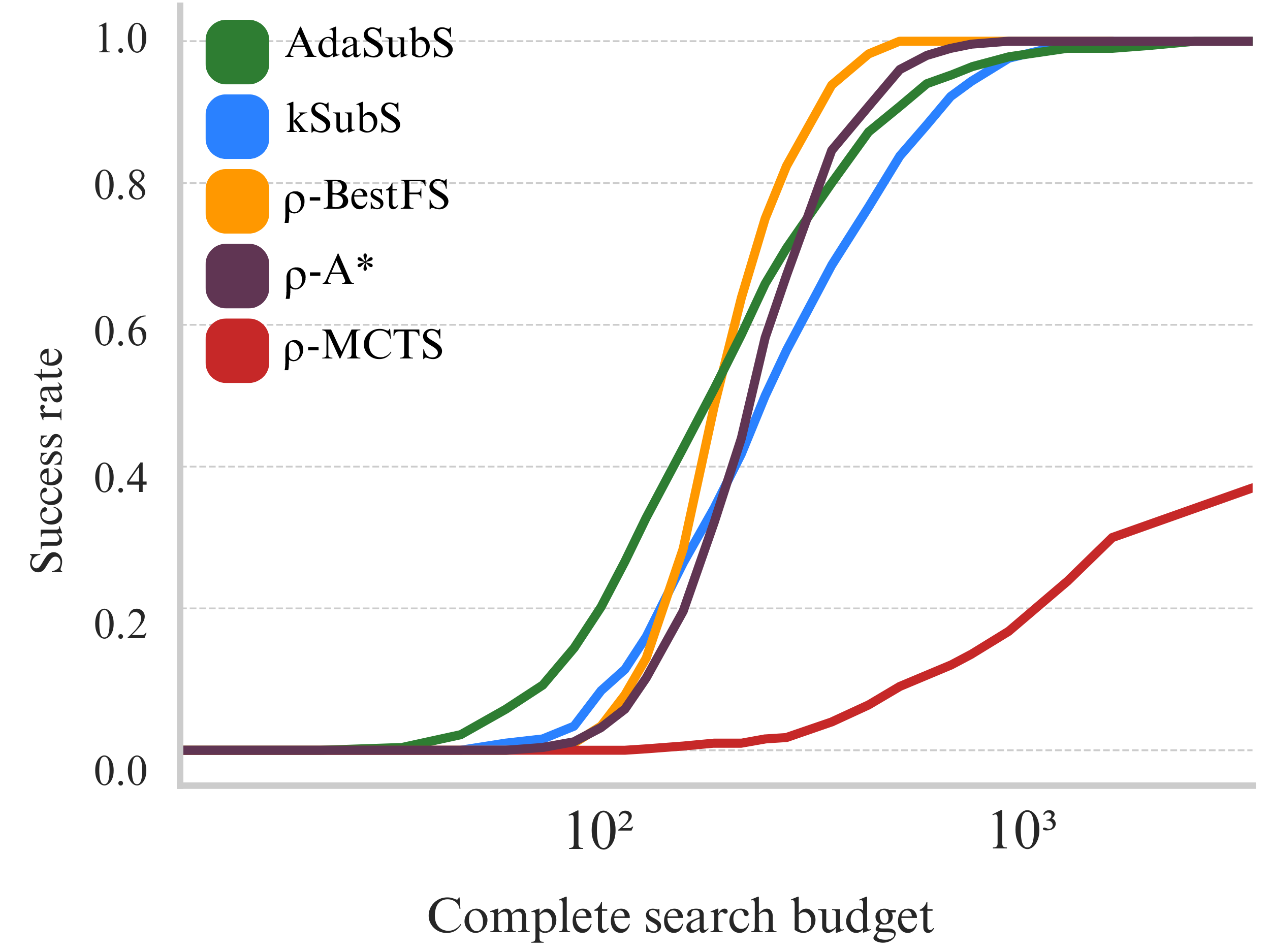}
        \caption{Solving the Rubik's Cube. Components are trained on reversed random shuffles.}
        \label{fig:rubik_random20}
    \end{minipage}
    \hfill
    \begin{minipage}[t]{0.32\linewidth}
        \includegraphics[width=\linewidth]{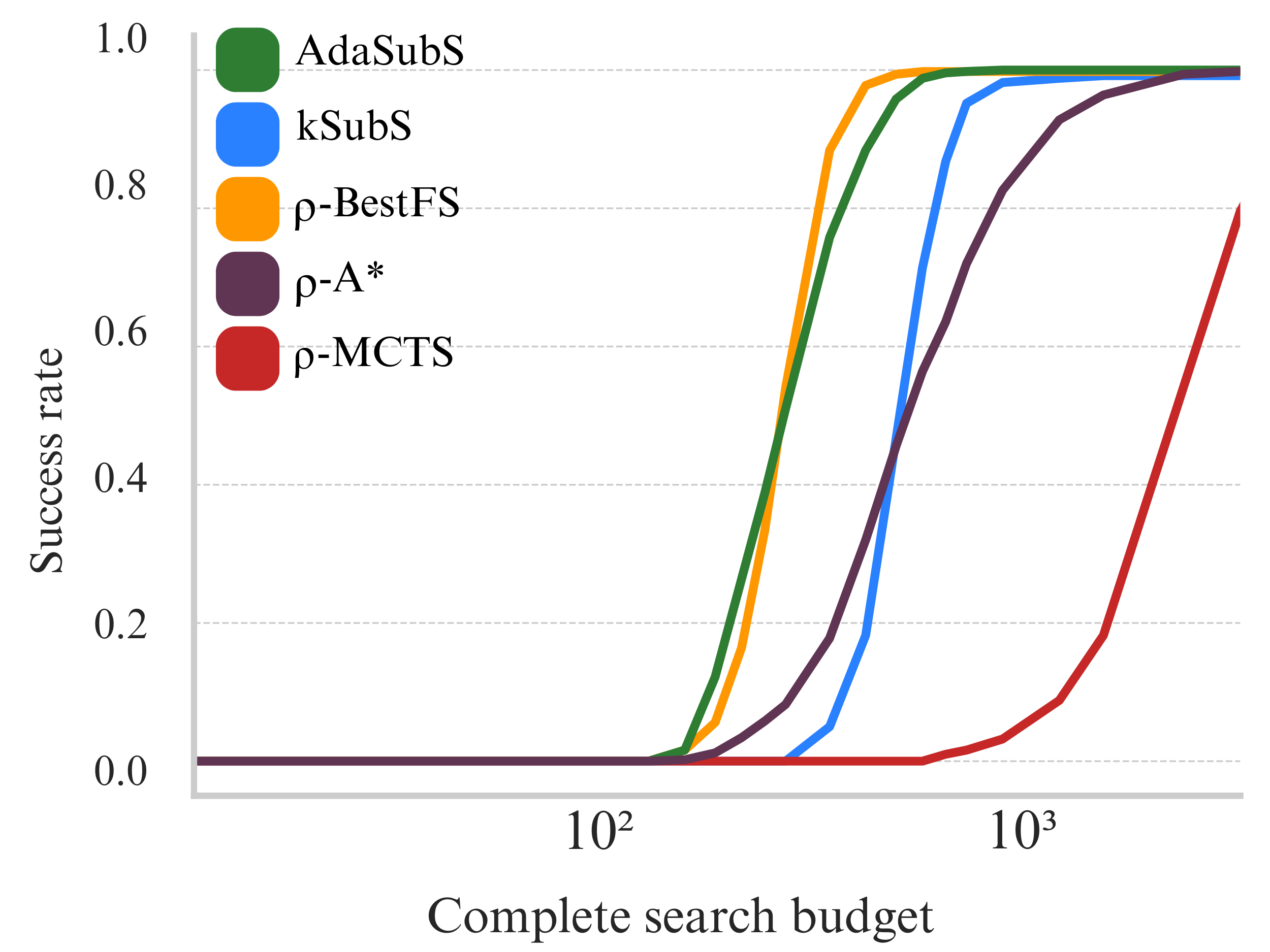}
        \caption{Solving the Rubik's Cube. Components are trained on the \textit{Beginner} algorithmic solver.}
        \label{fig:rubik_beginner}
    \end{minipage}
    \hfill
    \begin{minipage}[t]{0.32\linewidth}
        \includegraphics[width=\linewidth]{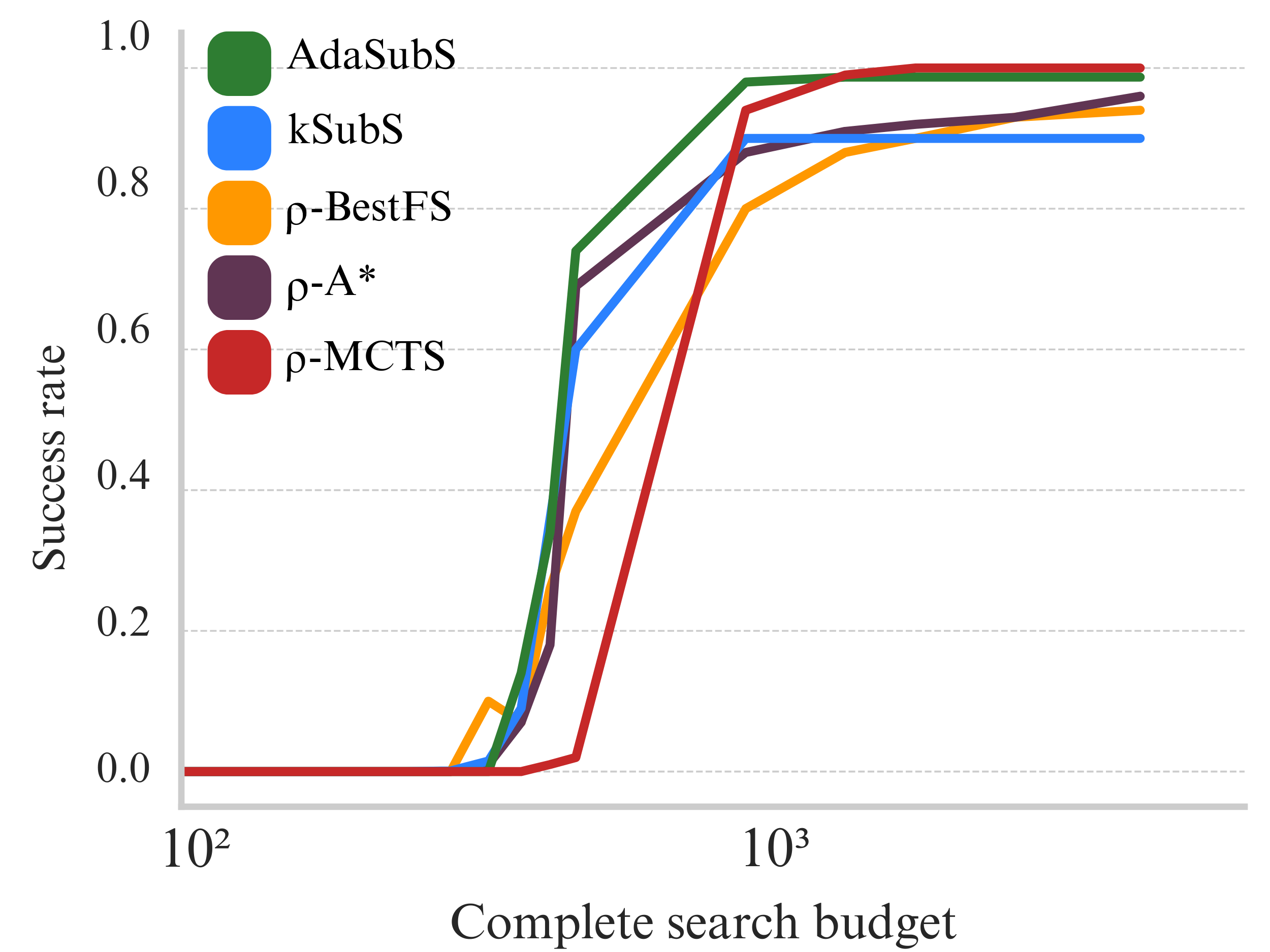}
        \caption{Solving N-Puzzle. Components are trained on an algorithmic solver.}
        \label{fig:npuzzle_structured}
    \end{minipage}
\end{figure}

As shown in Figures \ref{fig:rubik_moe}-\ref{fig:npuzzle_moe}, subgoal methods consistently outperform low-level methods by a wide margin (\rqperformance). However, when the training dataset is limited to a single source of demonstrations -- whether the demonstrations are long and structured or short and direct -- this performance gap disappears (see Figures \ref{fig:rubik_random20}-\ref{fig:npuzzle_structured}). Notably, subgoal methods, particularly AdaSubS, maintain stable performance across all training setups, while low-level methods are highly sensitive to the characteristics of the training data.


To explain those results, we found that value functions trained on diverse data often fail to assign consistently low values to the initial states of tasks. When demonstrations differ significantly in their length or execution style, the value function learns this variation, leading to inconsistent value predictions. Hierarchical methods can overcome this issue by relying on subgoals. Subgoals enable the agent to make long steps toward the solution, effectively bypassing regions of the state space where the value function is inconsistent or noisy, as it does not need to assess every small step along the way (this property is further studied in Section \ref{sec:analysis_noise}). In contrast, low-level methods operate on a finer, step-by-step level, executing small, atomic actions. This makes them more sensitive to the variability in the value function because they must evaluate each intermediate state on the way.

More detailed analysis of the experiments involving diverse data sources is provided in Appendix~\ref{appendix:properties_moe}.

\takeaways{Subgoal methods successfully leverage diverse demonstrations (\rqproperty), while low-level search performs better when trained on homogeneous trajectories (\rqproperty).}{}

\subsection{Subgoal Methods are Value Noise Filters}\label{sec:analysis_noise}

We found that the classical search algorithms are highly sensitive to the quality of the value function. To show this in a controlled setting, we added Gaussian noise to the value estimates and observed how different noise levels impacted the success rate of solving tasks.

\begin{figure}[htb]
    \centering
    \includegraphics[width=0.9\linewidth]{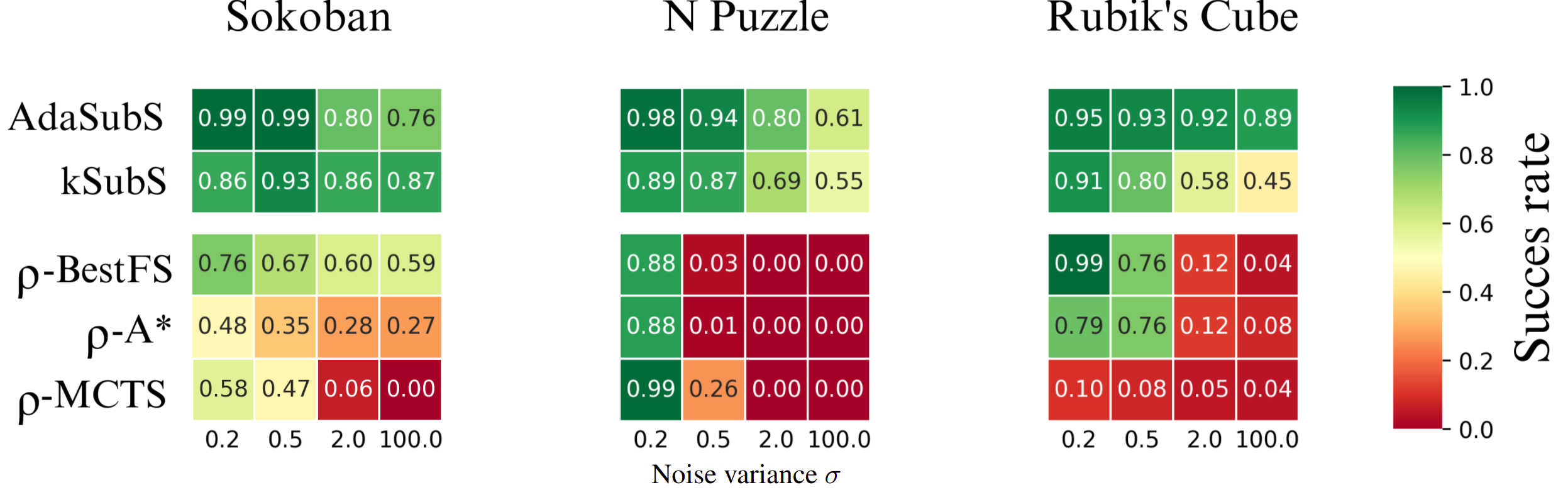}
    \caption{Success rate of low-level and subgoal methods as the approximation errors of the value function increase. $\sigma=100$ results in completely random value estimates.}
    \label{fig:noise_all}
\end{figure}

While \BestFS{} is able to solve nearly all instances under ideal conditions, its performance significantly declines as value function errors increase, even to $0\%$ (see Figure \ref{fig:noise_all}). \Astar{} and \MCTS{} behave similarly. In contrast, the subgoal methods show remarkable resilience. Particularly AdaSubS, which maintains nearly unchanged success rate, despite high value errors (\rqproperty).




These results align with our findings in Section \ref{sec:analysis_moe}, where using diverse training data naturally introduced value estimation errors. As observed by \citet{ada}, the search process of subgoal methods is guided by subgoal generators, which reduces reliance on the value function. Subgoal generators and the conditional policies connecting subgoals are not directly influenced by the value approximation errors. The value function is used only in high-level nodes, which represent only a fraction of the search tree.

\begin{figure}[h]
    \centering
    \begin{minipage}[t]{0.49\linewidth}
        \centering
        \includegraphics[width=0.9\linewidth]{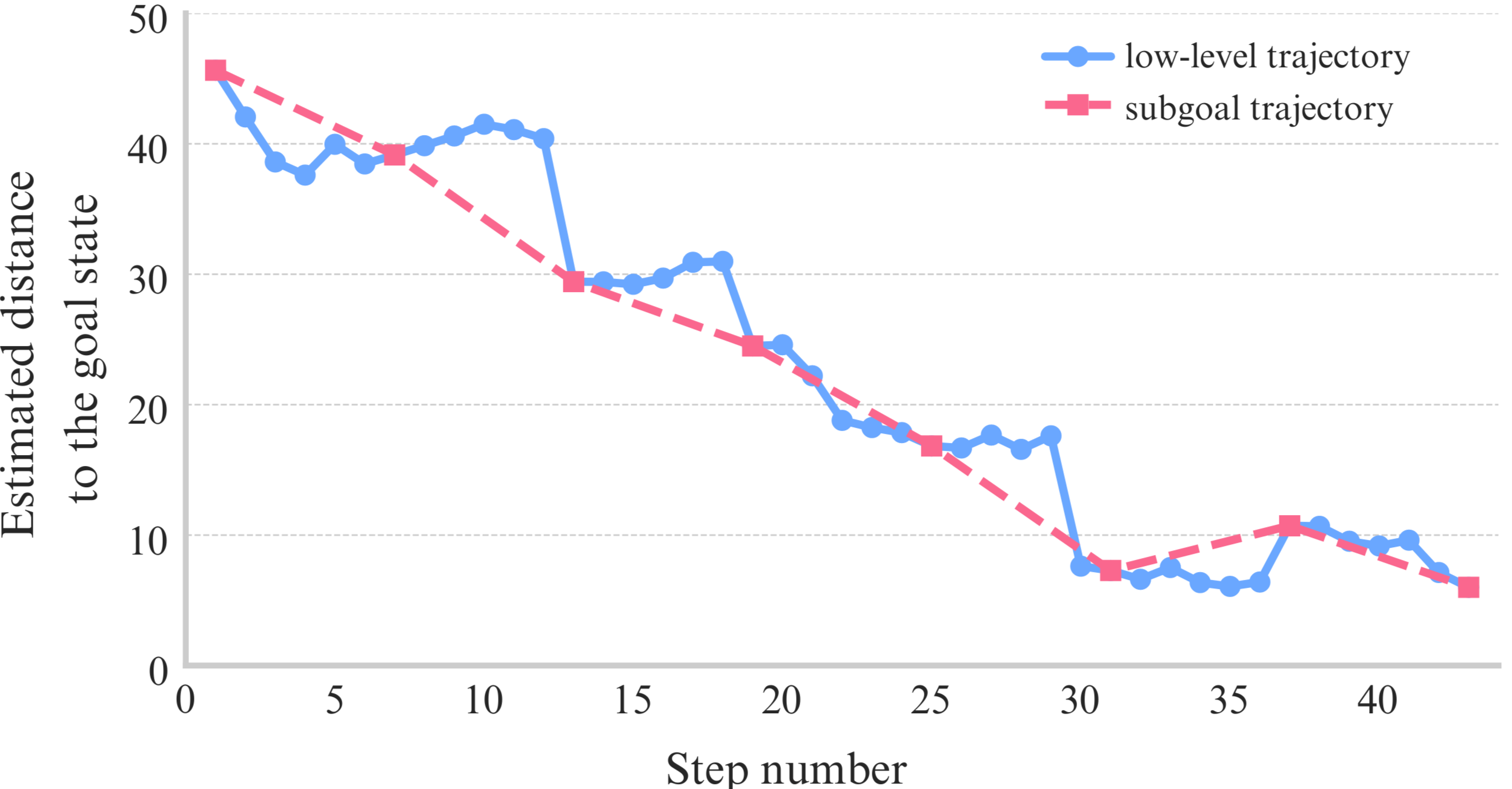}
        \caption{Value estimates along a solving trajectory generated by \BestFS. Even small approximation errors cause non-decreasing values, slowing down the search. In contrast, the subgoal path mitigates these errors, leading to mostly monotonic values along the trajectory.}
        \label{fig:noisy_trajectory}
    \end{minipage}
    \hfill 
    \begin{minipage}[t]{0.49\linewidth}
        \centering
        \includegraphics[width=0.9\linewidth]{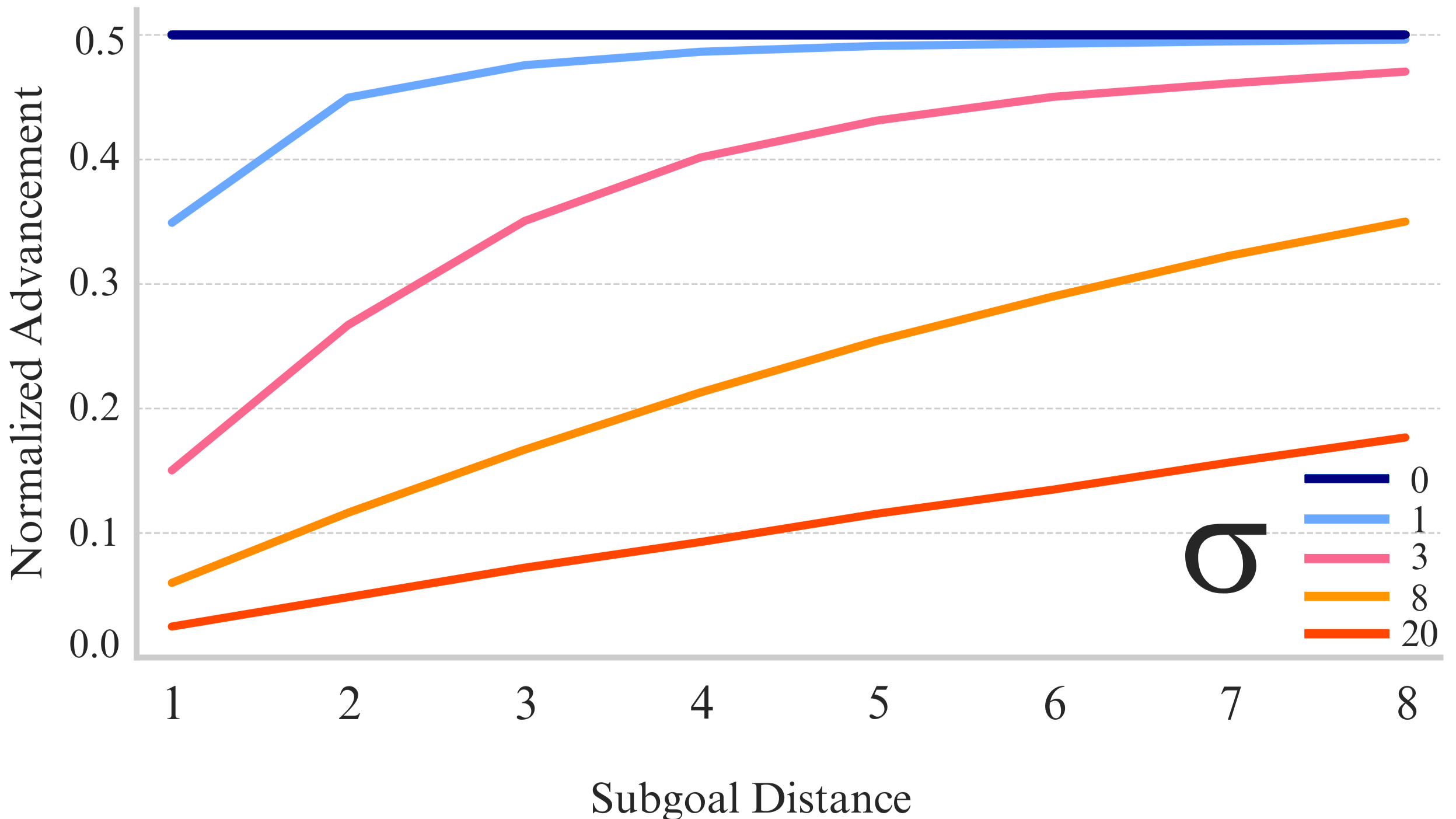}
        \caption{Normalized advancement ${\mathbb E_{Adv}}/{k}$ for a single search iteration, according to Theorem \ref{thm:noisy_search}. The value for each subgoal is divided by its length to represent the advancement per atomic action for easier comparison.}
        \label{fig:noise_advancement}
    \end{minipage}
\end{figure}

In hierarchical methods, the distance between high-level nodes spans multiple steps, increasing the likelihood that value estimates for subsequent high-level nodes along the solution path will be monotonic (see Figure \ref{fig:noisy_trajectory} for an illustrative example), which makes planning more efficient. This supports the claim by \citet{ksubs} that subgoals effectively mitigate the impact of value noise. To further ground that result, we prove the following theorem:

\begin{theorem}[Search advancement formula]\label{thm:noisy_search}
Let $g_k: S\to \mathcal P(S)$ be a stochastic \mbox{$k$-subgoal} generator that, given a state $s\in S$ samples a set of $b$ subgoals $\{s_i\}$ such that the distances $d(s_i, s)$ are independent, uniformly distributed in the interval $[-k;k]$. Let \mbox{$V:S\to\mathbb R$} be a value function with approximation error uniformly distributed in the interval $[-\sigma;\sigma]$.

Then, after $n$ iterations of search, the expected total progress toward the goal is:
\begin{equation}\label{eq:noise_main}
    \mathbb E_{Adv} = \frac{nb}{4\sigma k} \int_{-k}^k x\left( \int_{-\sigma}^{\sigma} \tilde u(x+h)^{b-1}\mathrm dh \right) \mathrm dx,
\end{equation}
where $\tilde u(x)$ is CDF of the sum of two uniform variables $U(-k,k)+U(-\sigma,\sigma)$.
Additionally, if we approximate that sum as $U(-k-\sigma,k+\sigma)$, we get
\begin{equation}\label{eq:noise_approximated}
    \mathbb E_{Adv} \approx \frac{n\left((k+\sigma)^b(bk^2+bk\sigma-2k\sigma-2\sigma^2) + \sigma^{b}(2k\sigma+bk\sigma+2\sigma^2)-k^b(bk^2)\right)}{(b+1) (b+2) k \sigma (k+\sigma)^{b-1}}
\end{equation}
\end{theorem}

\begin{proof}
See Appendix \ref{appendix:noise_theorem_proof} for the proof.
\end{proof}

Theorem \ref{thm:noisy_search} quantifies the expected progress of the search at each step, with Equation \ref{eq:noise_main} giving an exact formula and Equation \ref{eq:noise_approximated} providing a useful approximation. To compare subgoal methods with low-level methods in theory, under different levels of value approximation error, we model low-level search by setting $k=1$, which represents a single action. Figure \ref{fig:noise_advancement} shows the expected search progress with a branching factor of $b=3$, normalized by the number of actions leading to a subgoal. 

When value estimates are perfect (i.e., $\sigma=0$), both subgoal and low-level searches perform similarly. However, as value approximation errors increase, subgoal methods become significantly more resilient. At high noise levels ($\sigma=20$), single-step searches make very little progress, advancing only $0.025$ per action. In contrast, subgoals of length 8 achieve much greater progress -- $1.4$ for the entire subgoal, which is $0.175$ per action. This 7-fold increase in theoretical efficiency explains why subgoal methods outperform low-level methods in our experiments.

Further analysis of these experiments can be found in Appendix \ref{appendix:properties_noise}.

\takeaways{Subgoal methods successfully handle value approximation errors. Thus, they should be used when estimating the value is hard, for instance, when learning from diverse and suboptimal demonstrations (\rqproperty).}{}

\subsection{Subgoal Methods Handle Complex Action Spaces}\label{sec:analysis_complex_actions}

\begin{figure}[h]
    \centering
    \begin{minipage}[t]{0.45\linewidth}
        \includegraphics[width=0.8\linewidth]{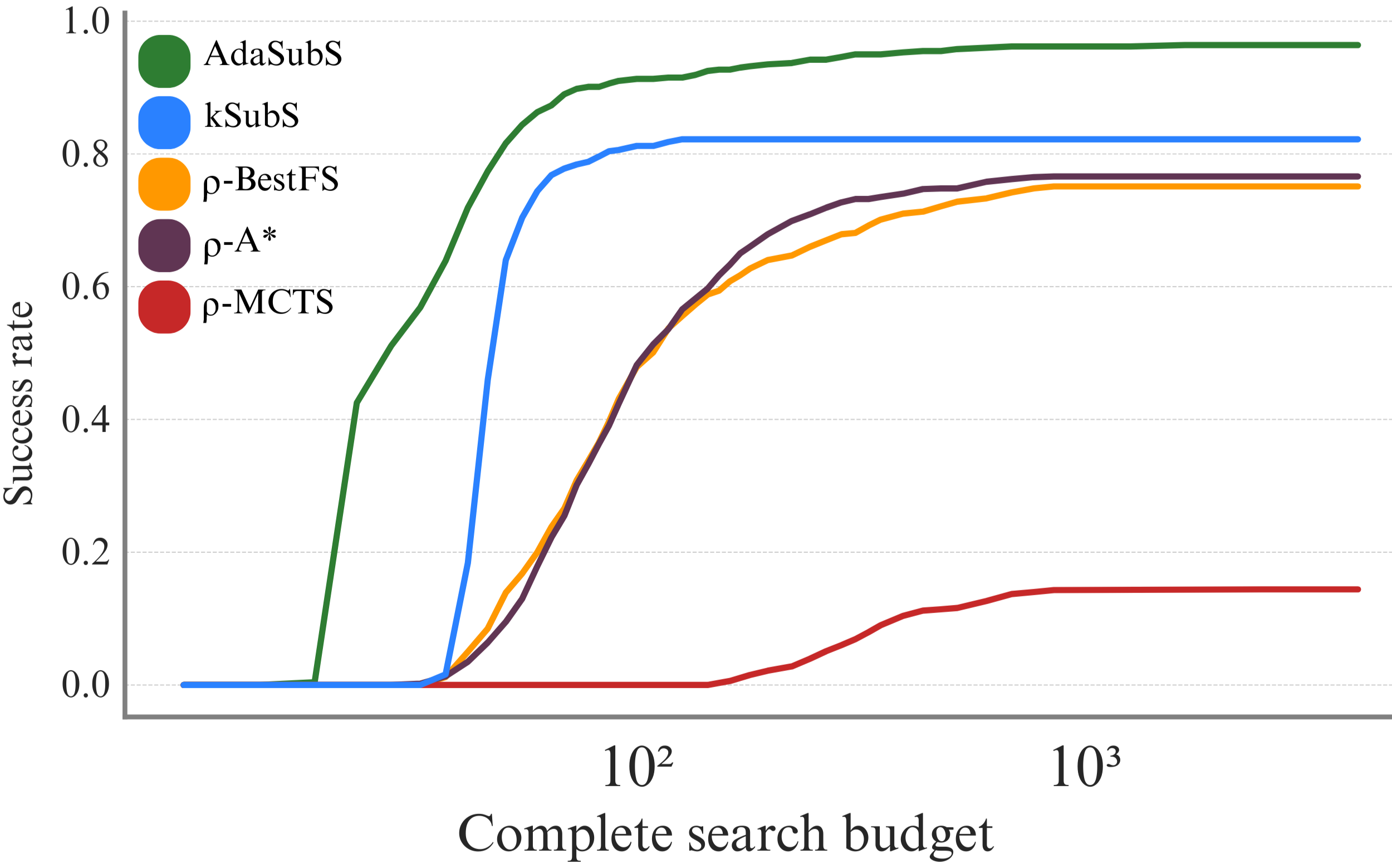}
        \caption{Solving INT. Components are trained on randomly generated proofs.}
        \label{fig:int}
    \end{minipage}
    \hfill
    \begin{minipage}[t]{0.45\linewidth}
        \includegraphics[width=0.8\linewidth]{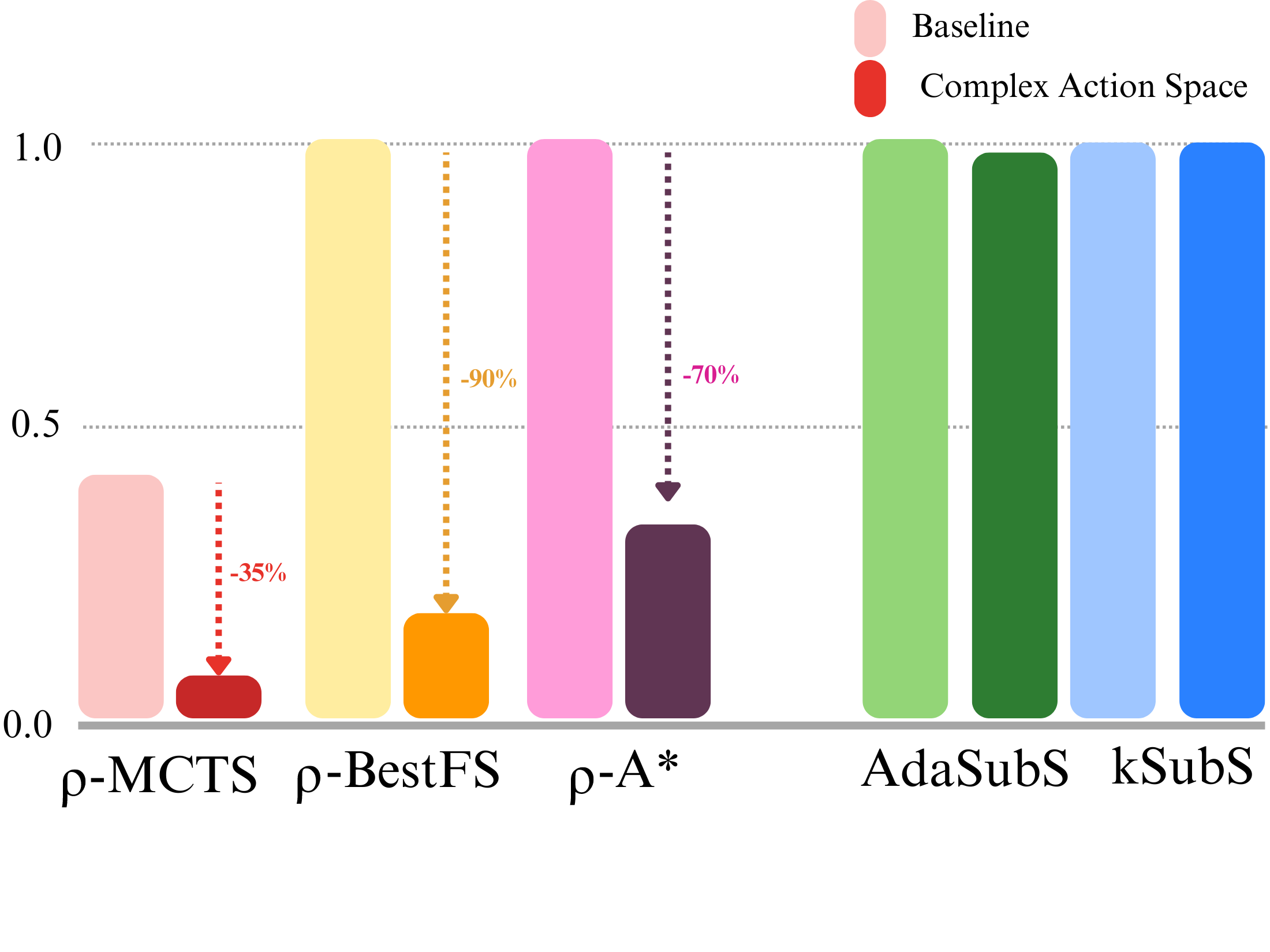}
        \caption{Solving the Rubik's cube with expanded action space, compared with the standard setup. Components are trained on reverse random shuffles.}
        \label{fig:rubik_actions_x100}
    \end{minipage}
\end{figure}

In environments with large action spaces, search methods often struggle due to the exponential increase in the number of choices \citep{DBLP:books/lib/SuttonB98}. As shown in Figure \ref{fig:int}, subgoal methods demonstrate a clear advantage over low-level search methods in the INT environment \citep{int}, a benchmark on proving mathematical inequalities (\rqperformance). The INT environment is particularly challenging because of its highly complex observation and action spaces, making it the most difficult benchmark among those used in \citep{ksubs, ada, hips, hipseps}.


Given a complex action space, each node expansion in low-level methods involves executing many similar actions, limiting their ability to efficiently search through the space. In contrast, subgoal methods compute actions only to connect subgoals, which is a much simpler task. This targeted approach reduces the negative impact of a large action space, allowing subgoal methods to maintain their efficiency even as the action space grows (\rqproperty).



To confirm this explanation, we conducted experiments on a modified version of the Rubik's Cube, where the action space was artificially inflated by giving the agent access to 100 copies of each action. As shown in Figure \ref{fig:rubik_actions_x100}, this simple modification drastically reduces the success rates of all low-level methods, even below 35\%. In contrast, subgoal methods remain largely unaffected, performing similarly to the standard setup. We can explain that result with the following theorem:

\begin{theorem}[Densification of the action space]\label{thm:complex_actions}
Fix any state $s$ from the state space $S$. Let $f : A \to [0, 1]$ be the action distribution induced by the data-collecting policy for the state $s$. Assume that $f$ is continuous and has a unique maximum.

For clarity, assume $A=[0,1]$. Consider a sequence of increasingly dense discrete action spaces $A_n := \{i / n\}_{i=0}^n\subset A$.
Let $\rho_n : S \times A_n \to [0, 1]$ be a family of policies that learn the distribution $f|_{A_n}$ over actions, with uniform approximation error $U(-E, E)$, where $E\in\mathbb R_+$.
Let $r_n$ be the range of the top $K$ actions according to the probabilities estimated by $\rho_n$. Then

$$\lim_{n \to \infty} \mathbb{E}[r_n] = 0.$$
\end{theorem}

\begin{proof}
See Appendix \ref{appendix:actions_theorem_proof}
\end{proof}

Intuitively, this theorem states that as the action space become more dense and complex, the actions sampled for search become increasingly less diverse, which strongly impedes successful planning. Note that this analysis is strictly more general than the last experiment, where we simply copied the available actions.
Further analysis of the experiments involving large action spaces is provided in Appendix~\ref{appendix:properties_action_space}.

\takeaways{When facing a problem with a complex action space, subgoal methods should outperform low-level search (\rqproperty).}{}

\subsection{Subgoal Methods Avoid Dead Ends}\label{sec:analysis_deadends}

\begin{wrapfigure}[12]{r}{0.45\textwidth}
\label{tab:dead_ends}
\vspace{-11pt}
    \centering
    \begin{tabular}{cc}
        \toprule
        \textbf{Search algorithm} & \textbf{Dead ends rate} \\ 
        \midrule
        \MCTS            & 22.0\%    \\
        \BestFS          & 18.5\%    \\
        \Astar              & 13.7\%    \\
        \midrule
        kSubS ($4$ steps)       & 12.7\%    \\
        kSubS ($8$ steps)       & 10.0\%    \\
        AdaSubS   & 8.86\%    \\
        \bottomrule
    \end{tabular}
\caption{Fraction of dead ends encountered during search between hierarchical and low-level methods in Sokoban.}
\end{wrapfigure}

Once an agent encounters a dead end, reaching the goal becomes impossible, leading to wasted computational effort. Our results, presented in Figure \ref{tab:dead_ends}, show that subgoal methods tend to enter dead ends less often than low-level methods. Using longer subgoals improves the ability to bypass those areas.

Among low-level methods, \Astar{} performs the best at minimizing dead ends rate, as its node selection regularizes values by depth in the search tree, preventing it from over-committing to dead ends. However, even \Astar{} is outperformed by subgoal methods, which rely on greedy value estimates and subgoals.

Deciding whether a state is a dead end can be NP-hard. Hence, it is much harder for the value function to penalize dead ends compared to the policy, which only ranks the available actions and does not have to identify dead ends \citep{feng2022left}. Furthermore, demonstrations used for imitation learning lead to the goal state, hence they contain no dead ends. Therefore the value function trained this way is never directly instructed to penalize dead ends. At the same time, during training of the policy the actions leading to dead ends are never reinforced. Our experiments show that hierarchical search relies much less on the value guidance compared to low-level search (Section \ref{sec:analysis_noise}), which further supports our conclusions. For a more detailed analysis, see Appendix \ref{appendix:properties_dead_ends}.

\takeaways{Subgoal Methods Are More Effective at Avoiding Dead Ends Compared to Low-Level Search (\rqproperty).}{}

\subsection{Subgoal Methods Generalize Out-Of-Distribution}\label{sec:analysis_ood}

\begin{figure}[ht]  
  \centering           
  \includegraphics[width=0.75\textwidth]{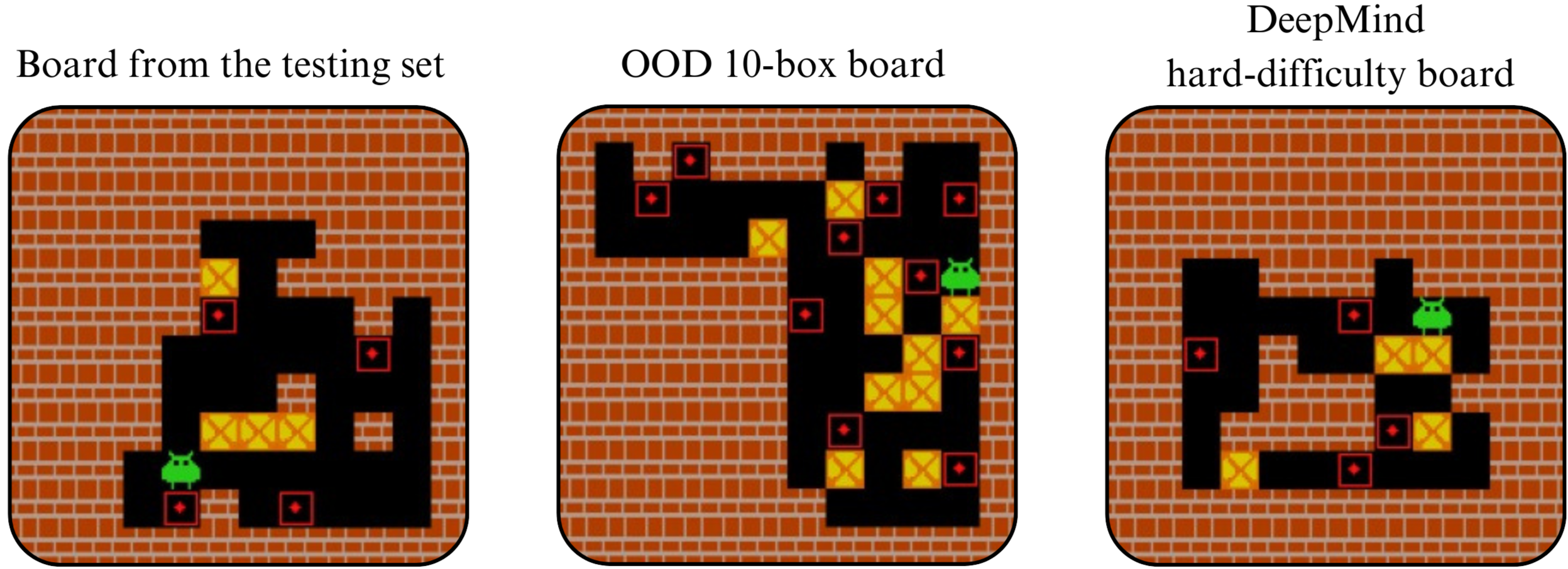} 
  \caption{Examples of Sokoban boards used in OOD experiments}
  \label{fig:sokoban_boards} 
\end{figure}

Planners that can generalize to out-of-distribution (OOD) instances are essential for robust decision-making \citep{DBLP:journals/jair/KirkZGR23, DBLP:journals/corr/abs-2108-13624}. We tested two types of generalization in the Sokoban environment: by significantly changing the layout of the board and by using extremely difficult boards from the DeepMind dataset \citep{boxobanlevels} (see Figure \ref{fig:sokoban_boards} for examples).

\begin{figure}[ht]
    \centering
    \begin{minipage}[t]{0.45\linewidth}
        \includegraphics[width=\linewidth]{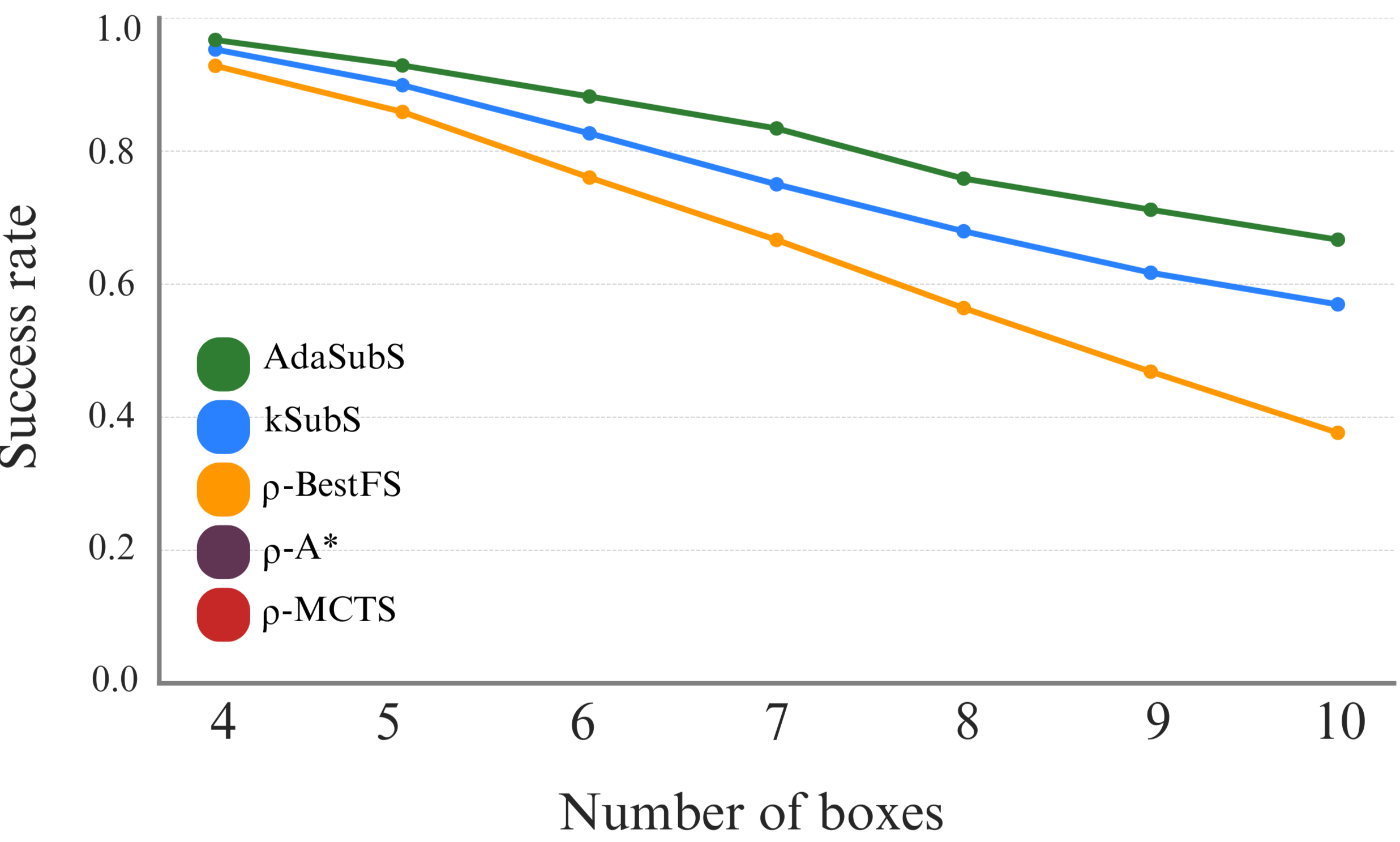}
        \caption{Averaged OOD results on Sokoban boards with OOD layouts. These instances were generated by systematically varying all parameters of the instance generator.}
        \label{fig:sokoban_ood}
    \end{minipage}
    \hfill
    \begin{minipage}[t]{0.45\linewidth}
        \includegraphics[width=\linewidth]{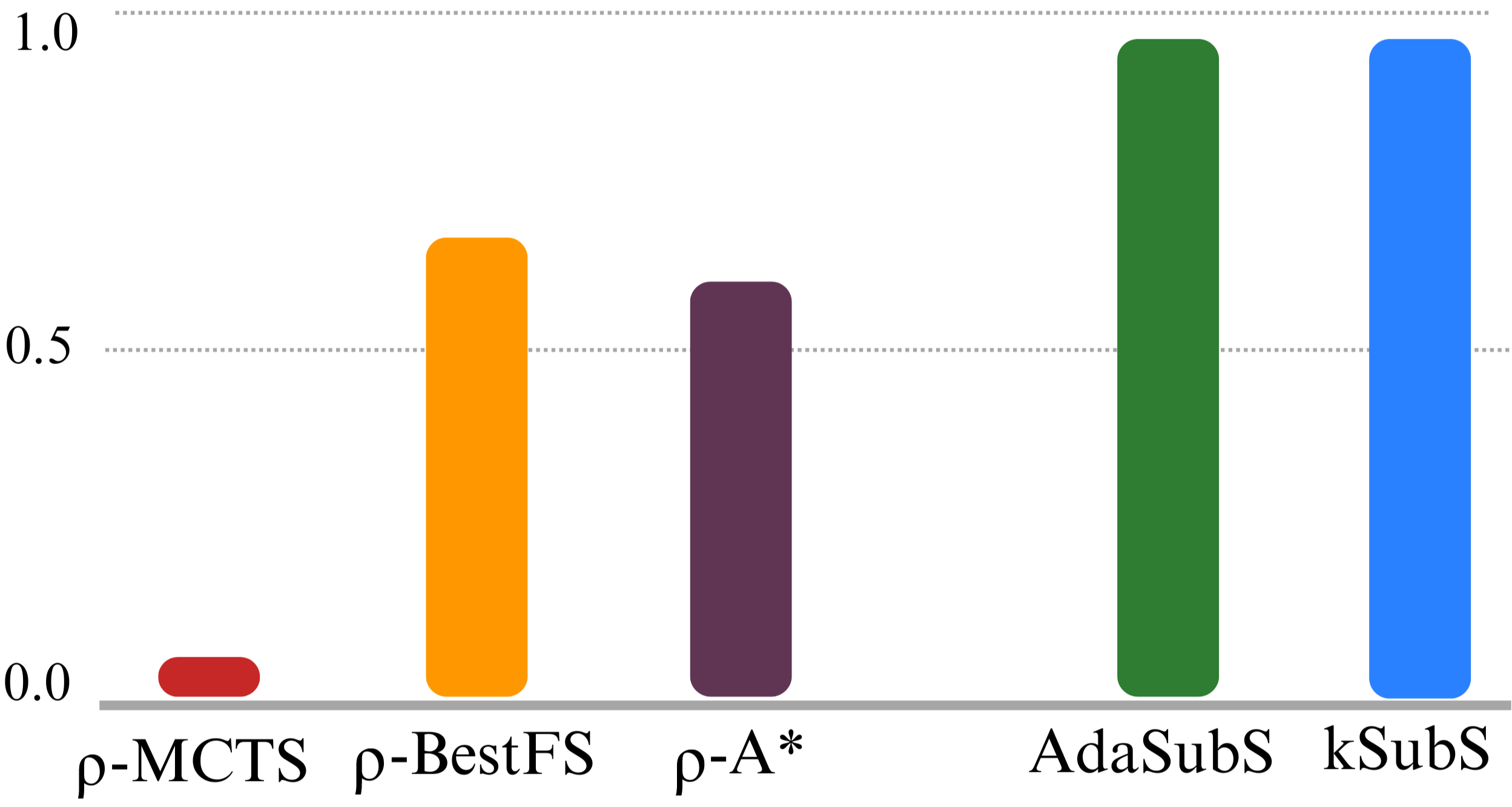}
        \caption{Performance on DeepMind extra hard boards.}
        \label{fig:sokoban_deepmind}
    \end{minipage}
\end{figure}

In both cases, subgoal methods show better performance than low-level methods, with the gap increasing as the distribution shift become more visible (see Figures \ref{fig:sokoban_ood}-\ref{fig:sokoban_deepmind}). However, we found that kSubS, when using twice longer subgoals, collapses in OOD evaluations, despite outperforming \BestFS{} and other low-level methods on in-distribution tasks. As the subgoal distance increases, predicting the distant future becomes more challenging, making it less likely for the generated subgoals to be valid and reachable, especially in OOD tasks. In contrast, low-level methods avoid this issue, as selecting an action from a limited set always results in a valid move. Thus, while subgoal methods can be effective in OOD scenarios, excessively long subgoals can degrade performance (\rqproperty).

When evaluated on extremely challenging instances (see Figure 1) introduced by \citep{boxobanlevels}, all methods required a significantly higher search budget but maintained the same performance order as in the previous experiment (\rqperformance). Solving these instances requires more advanced strategies than those learned during training. Subgoal methods are better equipped to handle this increased complexity because selecting subgoals is closely related to choosing a broader strategy because of their longer horizon. In contrast, low-level methods must assess each individual action, which limits their ability to foresee the long-term consequences of their choices.

\takeaways{Subgoal methods can scale better than low-level methods on OOD instances, provided the subgoals are not too long (\rqproperty).}{}

%% file: content/conclusions.tex
\section{Open Questions and Future Directions}\label{sec:open-questions}

While we identified several features that facilitate the performance of subgoal methods, that list is not exhaustive. Thus, it is essential to study this topic further, expand the analysis to more subgoal-based and low-level algorithms, and include even more types of environments. While most of our takeaways were confirmed in multiple environments, extending the evaluation to more domains would strengthen our conclusions. Additionally, our work provides mostly experimental validation of the claims. Finding theoretical foundations for the observed properties, such as Theorem \ref{thm:noisy_search}, would also be a valuable direction.

\section{Conclusions} \label{sec:conclusions}

We conducted a thorough comparison of hierarchical and low-level search methods for combinatorial reasoning tasks. Our experiments provides empirical and some theoretical evidence that hierarchical approaches should be preferred in environments where value estimation is challenging and learned estimates face significant uncertainty, particularly when learning from diverse suboptimal data. Furthermore, subgoal methods demonstrate better scalability in complex action spaces and are more effective at avoiding dead ends than low-level methods. Thus, in environments characterized by those properties, it is advisable to consider subgoal methods as an alternative to low-level search.

Based on our results, we propose guidelines for future research in this area. According to our experiments, the best-performing low-level search was usually \BestFS{} with a confidence threshold (see Appendix \ref{appendix:algorithms}). Although it is rather sensitive to the threshold value, which has to be optimized for each domain separately, we advocate using this simple method as a standard baseline for further research in hierarchical search. Our guidelines are comprehensively discussed in Appendix \ref{appendix:pitfalls}.

Additionally, we identified easy-to-overlook mistakes in reporting the results that may lead to misleading conclusions. Most importantly, the reported \emph{complete search budget} of hierarchical methods must include all the visited states and not only the high-level nodes as used in some prior works.

\section{Broader Impact}

Our study has broader implications for other complex domains. For example, advancements in robotics often face significant challenges due to limited data, leading many methods to rely on collective datasets like Open X-Embodiment \citep{open_x_embodiment_rt_x_2023}. As shown in our experiments, hierarchical search methods benefit substantially from training on diverse expert data (Section \ref{sec:analysis_moe}). Furthermore, the data bottleneck increases the need for the models to generalize to out-of-distribution scenes and tasks, which is also an advantage of hierarchical methods (Section \ref{sec:analysis_ood}). Finally, an essential aspect of robotics involves preventing the robot from becoming stuck or losing the manipulated object, events that can be seen as dead-end scenarios (Section \ref{sec:analysis_deadends}). Successful applications of hierarchical methods in robotics include models such as SuSIE \citep{DBLP:conf/iclr/BlackNAWFKL24} and HIQL \citep{DBLP:conf/nips/ParkGEL23}.

Additionally, our experiments indicate that hierarchical methods scale well in long-horizon tasks, as evidenced by their performance in the N-Puzzle and the Rubik’s Cube (using Beginner demonstrations), where the average sequence of steps often exceeds 200. Interestingly, while low-level methods can still perform well in these scenarios, we observed that they tend to be much more sensitive to hyperparameter tuning.

It is important to note that we do not claim hierarchical methods are universally superior to low-level approaches in all complex domains. Instead, the properties highlighted in our analysis suggest cases where they should be considered.

\section{Reproducibility Statement}

The code used to run all our experiments is available at \url{https://github.com/subgoalsearchmatters/what-matters-in-hierarchical-search}. We also link there datasets used for training our models. Hence, all our results are fully reproducible.

%% file: content/appendix.tex
\newpage
\section{Environments}\label{appendix:environments}

\begin{figure}[ht]
\centering
\includegraphics[width=\textwidth]{ 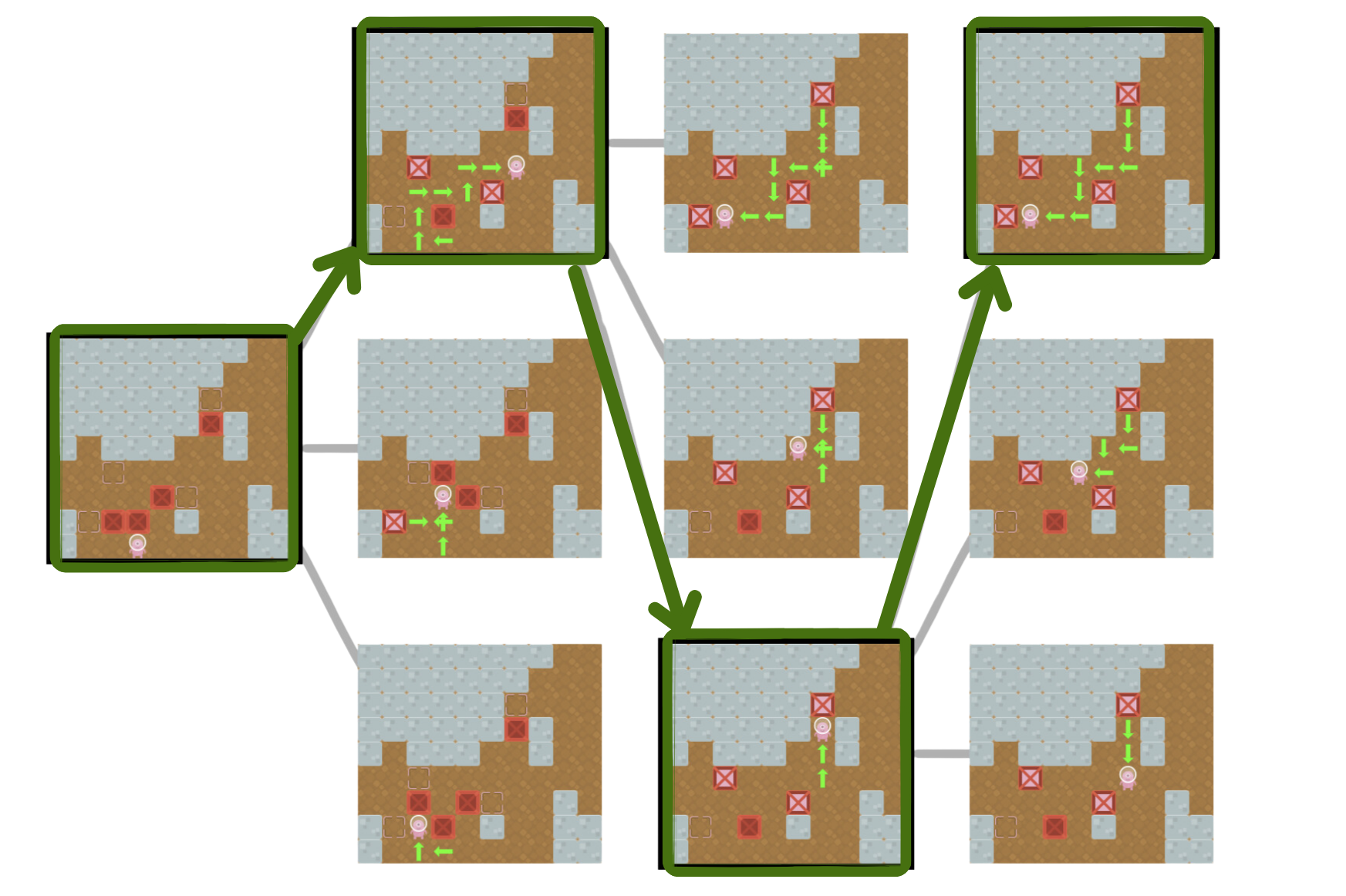}
\caption{Hierarchical Search applied to solving Sokoban. This tree, depicted in figures, employs bolded green arrows to highlight selected subgoals within a hierarchical search framework earmarked for subsequent exploration. The illustration demonstrates that these intermediate goals exhibit variability in terms of both their spatial distance and the methodology by which a planning algorithm may leverage them.
}
\label{fig:sokoban_search_tree}
\end{figure}

\textbf{Sokoban} Sokoban is a classic puzzle game where the objective is to push boxes onto target locations within a confined space. It is a popular testing ground for classical planning methods and deep-learning approaches due to its combinatorial complexity and difficulty in finding solutions. Recognized as a PSPACE-hard problem, Sokoban is used to evaluate different computational strategies. Our experiments use 12 × 12 Sokoban boards with four boxes to assess the performance of our proposed models. An illustrative example of a simple Sokoban search tree with a solving path is shown in Figure \ref{fig:sokoban_search_tree}.

\textbf{Rubik's Cube} The Rubik’s Cube, a renowned 3D puzzle, has over $4.3 \times 10^{19}$ possible configurations, highlighting the huge search space and the computational challenge it poses. Recent advancements in solving the Rubik's Cube with neural networks underscore the potential of deep learning methods in navigating complex, high-dimensional puzzles. For the exact representation of the Rubik's Cube state, see Figure \ref{fig:rubik_env}.

\begin{figure}[ht]
\begin{tcolorbox}[colback=white, colframe=black, width=\textwidth, height=3.9cm]
\scriptsize
\begin{align*}
\text{wbrwyggwwoboybygbryrorroboygrbggbggbwybrooogrywrowywwy} && s_0 && \text{Initial State} \\
\text{wbrwyggggobwybwgbgooyrroyrbrrbwgbygbwybrooogroyrowywwy} && s_1 && \text{One Action (= single rotation)} \\
\text{wbywyoggbobwybwgbgoorrryyryywrggrbbbgybgoorgroyoowrwww} && s_2 \\
\text{gyowyoggbwbwwbwwbgoorrryyryywwggbbbyboryogggroyoowrbrr} && s_3 \\
\dots \\
\text{yyyyyyyyybbbbbbrrrrrrrrrgggggggggooooooooobbbwwwwwwwww} && s_{n-1} \\
\text{yyyyyyyyybbbbbbbbbrrrrrrrrrgggggggggooooooooowwwwwwwww} && s_n && \text{Solving State}
\end{align*}
\end{tcolorbox}
\caption{Example trajectory of Rubik starting from initial state $s_0$ leading to the final solution $s_n$.}
\label{fig:rubik_env}
\end{figure}

\begin{figure}[ht]
\centering
\includegraphics[width=0.8\textwidth]{ 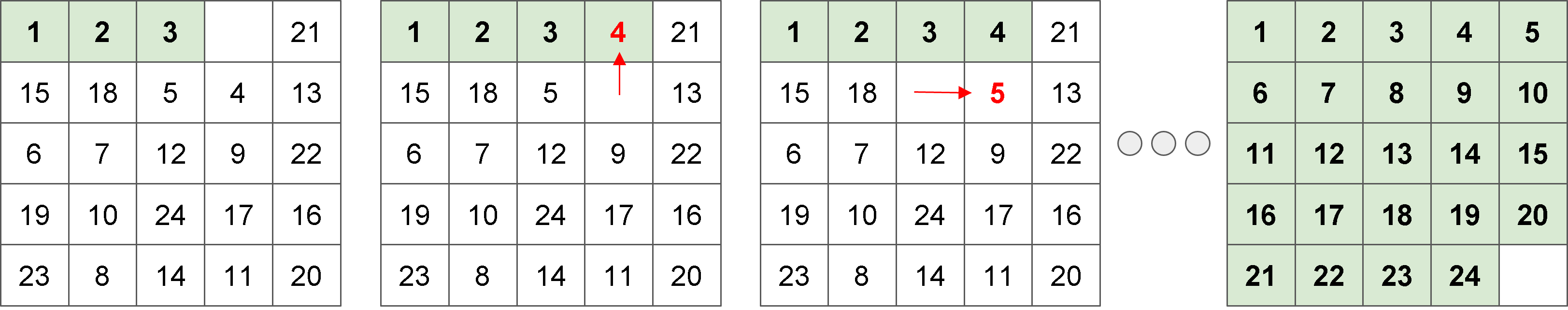}
\caption{Example trajectory of n-puzzle starting from initial state $s_0$ leading to the final solution $s_n$. Red arrows indicate low-level actions.}
\label{fig:n_puzzle_env}
\end{figure}

\textbf{N-Puzzle} The N-Puzzle, a classic sliding puzzle game, comes in various sizes, including the 3x3 (8-puzzle), 4x4 (15-puzzle), and 5x5 (24-puzzle). The goal is to rearrange a frame of numbered square tiles into a specific pattern, a task that tests the algorithm's ability to plan and execute a sequence of moves efficiently. Figure \ref{fig:n_puzzle_env} shows a visualization of a trajectory in 24-puzzle.

\textbf{INT} INT (INequality Theorem proving) is an automated theorem-proving benchmark for high school algebraic inequality proofs. \citep{int} provides a generator of mathematical inequalities and a proof verification tool. Each action in INT maps to a proof step, which specifies a chosen axiom and its input entities - which makes action space very high-dimensional, enabling up to a million valid actions at a step. This large action space makes INT a desirable but challenging environment for expanding HRL paradigms to vast action spaces. 

We used 25-step proofs for this paper, representing an uplift from 15 considered in \citep{ksubs, ada} (the latter used longer proofs, but only for evaluating 15-trained models). Each step is an application of an axiom to an axiom-specific number of entities (entities are bracketed or bracketable parts of the theorem's goal).

\begin{figure}[h]
\begin{tcolorbox}[colback=white, colframe=black, title={Example Theorems for INT environment}, height=7.6cm]
\scriptsize
\begin{align*}
\textbf{Theorem 1 Premises:} & \quad \begin{aligned}[t]
& ((c + c) + d) \geq a; \\
& (d + e) \geq 0; \\
& ((c + c) + f) \geq (0 + a); \\
& (b + g) \geq 0;
\end{aligned} \\
\textbf{Goal:} & \quad \begin{aligned}[t]
& (((((((c+c)+(c+c))\cdot 4c) +((c+c)+d))+(d+e))+((c+c)+f))+(b+g)) \\
& \quad \geq ((((0+a)+0)+(0+a))+0)
\end{aligned} \\
\textbf{Theorem 2 Goal:} & \quad (((0 + b) + c) + a) \geq (0 + (0 + (b + (c + a)))) \\
\textbf{Theorem 3 Premises:} & \quad \begin{aligned}[t]
& (a + d) \geq 0; \\
& (a + e) \geq (c \cdot c); \\
& (e + f) \geq 0; \\
& (c + g) \geq 0; \\
& (c + h) \geq (c + g); \\
& (c + i) \geq 0;
\end{aligned} \\
\textbf{Goal:} & \quad \begin{aligned}[t]
& (((((((c \cdot c)\cdot(a + d)) + (a + e))\cdot(e + f))\cdot(c + g)) + (c + h))\cdot(c + i)) \\
& \quad \geq ((((((0 \cdot(a + d)) + (c \cdot c))\cdot(e + f))\cdot(c + g)) + (c + g)) \cdot (c + i))
\end{aligned}
\end{align*}
\end{tcolorbox}
\caption{A comprehensive representation of theorems pertaining to goal achievement in mathematical expressions, showcasing the logical structure and underlying premises leading to the formulated goals.}
\label{fig:int_env}
\end{figure}

\newpage
\section{Key Factors For Hierarchical Search}\label{appendix:properties}

According to our experiments, the attributes pivotal for leveraging the advantages of high-level search include:
\begin{itemize}
    \item learning from diverse data sources,
    \item hard-to-learn value function,
    \item complex action space,
    \item presence of dead ends
\end{itemize}

In Section \ref{sec:experiments}, we show our main experiments that support our findings. In this appendix, we present an extended analysis of each property.



\subsection{Learning from diverse data sources}\label{appendix:properties_moe}

Achieving superhuman performance in complex tasks, as demonstrated by AlphaGo \cite{alphago}, often involves large-scale datasets of demonstrations obtained from agents with varying skill levels and strategies. However, this diversity introduces challenges such as inconsistencies in demonstrations and variations in quality \citep{DBLP:journals/corr/abs-2004-07219, DBLP:conf/nips/ChenLRLGLASM21, DBLP:journals/corr/abs-2005-01643}. Widely used datasets like D4RL \citep{DBLP:journals/corr/abs-2004-07219}, Open X-Embodiment \citep{open_x_embodiment_rt_x_2023}, or Waymo Open Dataset \citep{DBLP:conf/cvpr/SunKDCPTGZCCVHN20} reflect this diversity, highlighting the need to address these challenges effectively. We want to answer the question whether such setting is handled better by high-level or low-level search algorithms.

\textbf{Experiment setup} 
For this analysis, we focus on the Rubik's cube environment. We collected a dataset of $500\,000$ trajectories, computed with four different solvers for the Rubik's cube:

\begin{itemize}
\item Beginner -- the simplest human-oriented solving algorithm. It aims to order the cube layer by layer with a few primitive tactics. Because of that the solutions are structured, but also very long (typically between 150 and 200 moves).
\item CFOP -- an algorithm designed for speedcubers. It is based on the same principle as Beginner, but employs many advanced tactics that make the solutions faster (typically about 100 moves).
\item Kociemba -- a computational solver that finds near-optimal solutions (usually between 20 and 40 moves) in short time. It is heavily optimized based on the algebraic properties of the Rubik's cube.
\item Random -- solutions obtained by scrambling an ordered cube with random moves and reversing the trajectory.
\end{itemize}

Figure \ref{fig:example_solutions} shows example solutions generated with each solver. Clearly, the algorithmic solvers (Beginner and CFOP) generate much longer solutions that the other methods. They are also more structured, as they are based on building patterns. The computational solver Kociemba on the other hand go directly towards the solution because its moves are carefully optimized to ensure maximal advantage. Because of that, this dataset represent a truly diverse set of demonstrations.

\textbf{Results} As shown in Figure \ref{fig:rubik_moe}, the subgoal methods outperform the low-level methods by a wide margin. While \BestFS{} is comparable on small budgets, it struggles with solving most of the instances. Also, it should be noted that the performance of the subgoal methods changes only slightly compared to training on a single Random solver (Figure \ref{fig:rubik_random20}) while the low-level searches are heavily affected.

\textbf{Learned values} To find the sources of that outcome, we checked the values learned by the heuristic function. Because of the diversity introduced by combining the experts, we should expect that the estimates are subject to high uncertainty and possibly high variance.

\begin{figure}[ht]
    \centering
    \includegraphics[width=0.5\textwidth]{ 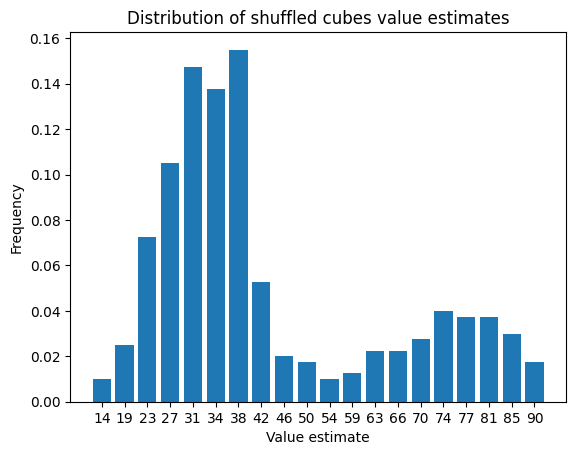}
    \caption{Value distribution for fully scrambled cubes, learned on data coming from diverse experts. The values are rescaled so that the x-axis represent the estimated number of steps to the solution. The values represent the mean of each interval.}
    \label{fig:initial_values}
\end{figure}

Figure \ref{fig:initial_values} shows the distribution of the learned heuristic for random fully shuffled cubes. Although most instances can be solved optimally within 20-26 moves, the estimates range from 14 to 90 steps. Furthermore, the distribution is clearly bimodal -- one mode correspond to a typical length of Kociemba solution, the other to CFOP. 

\begin{figure}[ht!]
    \def \plotw {0.75}
    \centering
    \begin{minipage}[t]{\plotw\textwidth}
        \includegraphics[width=\textwidth]{ 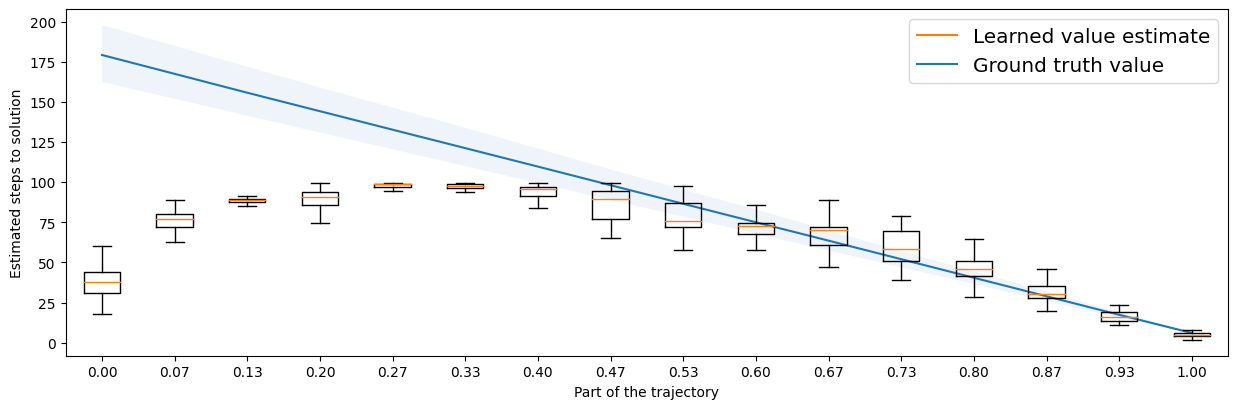}
        \vspace{-2em}
        \caption{Beginner solver}\label{fig:values_beginner}
    \end{minipage}
    
    \begin{minipage}[t]{\plotw\textwidth}
        \includegraphics[width=\textwidth]{ 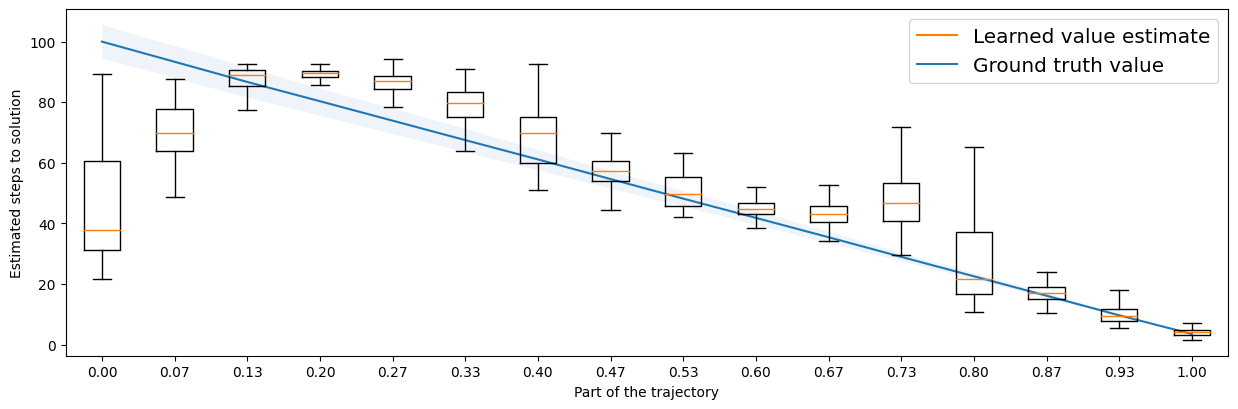}
        \vspace{-2em}
        \caption{CFOP solver}\label{fig:values_cfop}
    \end{minipage}
    
    \begin{minipage}[t]{\plotw\textwidth}
        \includegraphics[width=\textwidth]{ 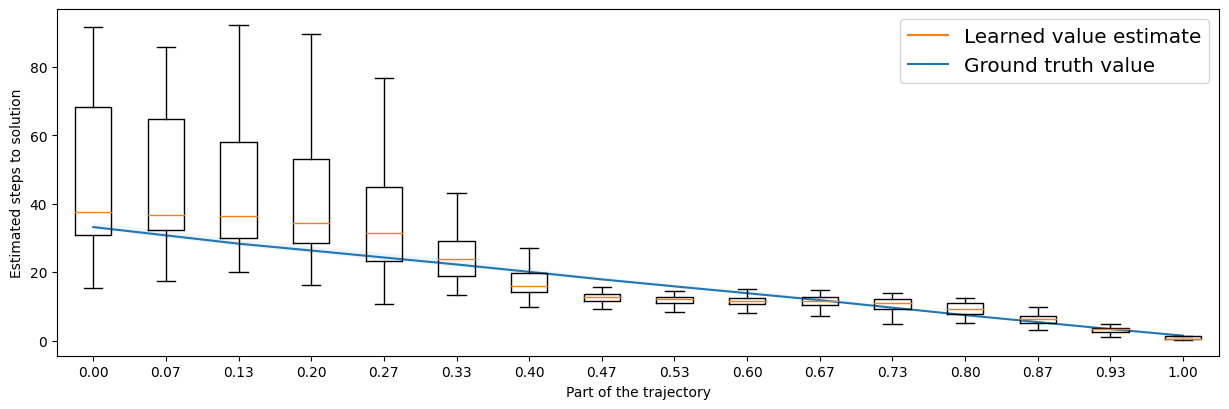}
        \vspace{-2em}
        \caption{Kociemba solver}\label{fig:values_kociemba}
    \end{minipage}
    
    \begin{minipage}[t]{\plotw\textwidth}
        \includegraphics[width=\textwidth]{ 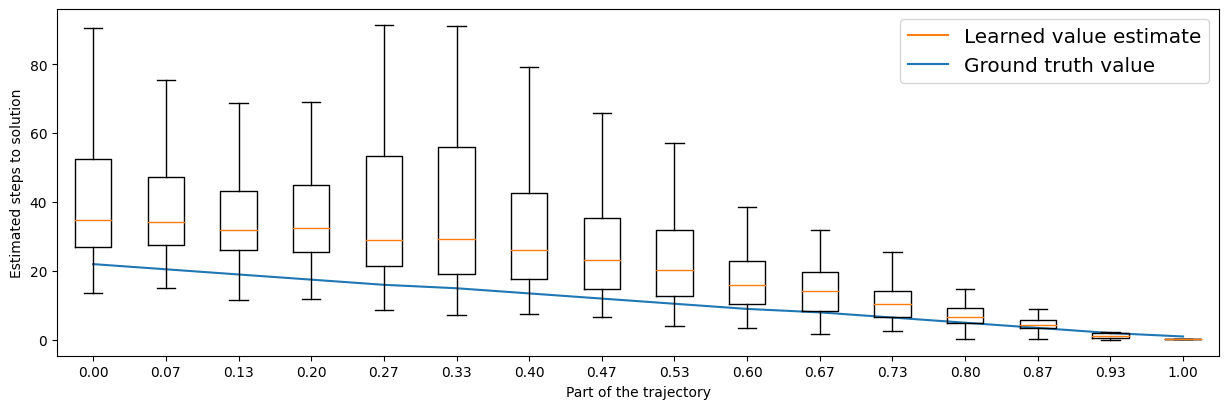}
        \vspace{-2em}
        \caption{Reversed random 20-move trajectories}\label{fig:values_random}
    \end{minipage}
    
    \caption{The learned value estimates distribution for various solvers. For each plot 100 episodes were solved using the respective solver. The boxes represent the distribution of value estimates for the consecutive points of the solution. The x-axis denotes the relative part of the trajectory (i.e., $0.5$ denotes the middle point in each trajectory, regardless of its length). The blue line indicates the true number of steps to the solution.}\label{fig:values_moe}
\end{figure}

Furthermore, Figure \ref{fig:values_moe} shows the distribution of value estimates throughout the solutions for each solver. We observe that for the algorithmic solvers the initial distance is considerably underestimated. After about $20\%$ moves the value network recognizes the pattern of layers built by the solvers and expect a long solution by assigning values close to $100$. On the other hand, the values learned for the states visited by the computational solvers start as overestimated, but steadily decrease towards 0.

While it is a reasonable strategy for the value to fit to the provided dataset, it creates a challenge for the search. If a search algorithm aims to imitate Beginner or CFOP, it has to reach the layer pattern, characteristic of those solvers. However, the random states tend to have very low distance estimate, compared to the initial layer patterns. Because of that, for tens of steps the heuristic estimates would be actually increasing, making the reached states less and less probable to expand.

In practice, the low-level searches usually fail to cross this gap. On the other hand, the high-level methods are partially guided by the subgoal generators that ignore the values. The value gap that spans across about 30 steps can be crossed with as few as 5 subgoals of length 6. Because of that both kSubS and AdaSubS can successfully leverage the schematic algorithmic solutions.

To finally confirm that conclusion, we must answer the question whether the performance of low-level searches would increase if they could leverage the algorithmic solutions as well. For that purpose, we trained the components for each method using data only from the Beginner solver. This way we remove the challenge of noisy initial values. As shown in Figure \ref{fig:rubik_beginner}, the low-level searches indeed perform much better. BestFS even matches the performance of AdaSubS. That confirms our observation that low-level searchas fail to utilize multimodal data because they rely too much on the value function and seek monotonic slopes.

At the same time we observe that since BestFS and AdaSubS show nearly identical performance on Beginner solutions, it is questionable that hierarchical methods handle long-horizon tasks better, which is a common belief \citep{DBLP:conf/nips/NachumGLL18, sorb, DBLP:journals/corr/abs-2401-02644}.

    
    
    
    

\begin{figure}[h!]
    \centering
    \begin{minipage}[t]{0.2\textwidth}
        \centering
        \fontsize{2}{0}
        \selectfont
        \texttt{\input{ resources/example_Beginner.txt}}
        \caption{Beginner}\label{fig:example_beginner}
    \end{minipage}
    \hfill
    \centering
    \begin{minipage}[t]{0.23\textwidth}
        \centering
        \fontsize{1mm}{3}
        \selectfont
        \texttt{\input{ resources/example_CFOP.txt}}
        \caption{CFOP}\label{fig:example_CFOP}
    \end{minipage}
    \hfill
    \centering
    \begin{minipage}[t]{0.27\textwidth}
        \centering
        \fontsize{3}{4}
        \selectfont
        \texttt{\input{ resources/example_Kociemba.txt}}
        \caption{Kociemba}\label{fig:example_kociemba}
    \end{minipage}
    \hfill
    \centering
    \begin{minipage}[t]{0.27\textwidth}
        \centering
        \fontsize{3}{4}
        \selectfont
        \texttt{\input{resources/example_Random.txt}}
        \caption{Random}\label{fig:example_random}
    \end{minipage}
    \caption{Example solutions computed by each solver. Because the algorithmic solvers typically require over $100$ steps, we use a tiny font to display it.}\label{fig:example_solutions}
\end{figure}

\subsection{Value Approximation Errors}\label{appendix:properties_noise}

In many practical scenarios, value function estimates are based on either limited data samples or handcrafted heuristics \citep{deepblue, Mnih2015, walke2023bridgedata}. This often leads to high approximation errors. If search algorithms rely too heavily on these imperfect estimates, they can make poor decisions, especially in large and complex environments where accurate value estimates are even harder to obtain \citep{open_x_embodiment_rt_x_2023, starcraft}.

Section \ref{appendix:properties_moe} hints that when value estimates are subject to high uncertainty, subgoal methods should outperform low-level searches.
To confirm that intuition, we run an experiment in a Rubik's cube, N-Puzzle, and Sokoban environments (Section \ref{sec:analysis_noise}). During inference, we add additional noise to the value estimates. That is, whenever a node is added to the search tree and its value estimate equals $\hat v$, we add it with the value of $\hat v + \mathcal N(0, \sigma)$ instead.

Figure \ref{fig:noise_all} shows that as the amount of noise increases, each low-level method gets less and less efficient. On the extreme, when using fully random values ($\sigma=100$), they struggle to solve any instance.

On the other hand, subgoal methods are much more resilient to noise in the value. Adaptive Subgoal Search is nearly not affected by the presence of noise. kSubS is able to retain as much as $40\%-90\%$ success rate, even with completely random values.

Observe that the search performed by low-level methods is guided mainly by the value function. Hence, if the computed estimates are subject to high variance, low-level search struggles to make any progress. On the other hand, the subgoal search is guided both by the value function and the subgoal generator. Both the subgoal generator and the conditional policy that connects subgoals do not depend on the values. Hence, the value function is used only in the high-level nodes, which is only a fraction of the search tree.

An extreme case of that behavior is demonstrated by Adaptive Subgoal Search. Because in our configuration each generator outputs a single subgoal, the value is nearly not used at all for search. Only when the search is stuck, the secondary generators select the highest-ranked node to expand, which in this case is simply a random node of the tree. To summarize, given random value estimates, AdaSubS reduces to the following strategy:
\begin{enumerate}
    \item Start from the root node,
    \item Move from the current node to the subgoal until possible,
    \item If the search is stuck, expand a random node in the search tree with a secondary generator and return to (2).
\end{enumerate}
The experiments show that this simple strategy is surprisingly competitive to the greedy best-first approach, even without noise. Interestingly, it could be implemented in low-level search as well. We leave that promising experiment for future work.


\subsection{Complex Action Spaces}\label{appendix:properties_action_space}

In environments with large action spaces, search methods often struggle due to the exponential increase in the number of choices at each decision point \citep{DBLP:books/lib/SuttonB98}. This complexity makes it difficult to efficiently identify optimal actions, slowing down decision-making and exploration \citep{DBLP:journals/corr/Dulac-ArnoldESC15, alphago}.

The primary difference between low-level methods and subgoal methods is that the former predicts the next action, and the latter -- the next state. In many environments, the action space is as simple as a few bits, allowing for iterating over all possible actions, and sampling them. At the same time, states may be considerably larger, up to the extreme of image observations. However, in some environments, the action space is comparable to the state space, or even more complex. A classic example is the AntMaze environment, in which actions are 8-dimensional, while the goal space is only 2-dimensional \citep{DBLP:journals/corr/abs-2004-07219}.

Among the combinatorial reasoning environments we consider, INT has the most complex action space. In INT, actions correspond to proof steps and are represented as the chosen axiom, specification of its input entities, and the required premises \citep{int}. Thus, the complexity of the action is at least comparable to the states. Moreover, solving the INT inequalities is based on constant simplification of the given expression, so the state is getting even smaller with each step.

Our experiments, shown in Figure \ref{fig:int}, clearly confirm the advantage of using subgoal methods in the INT environment. To further verify the source of that advantage, we conducted another experiment, in a modified Rubik's cube environment. Recall that the experiment presented in Section \ref{sec:analysis_moe} shows that subgoals offer no significant advantage in the \textit{original} Rubik's cube (with a single data source). Now, we want to check whether the outcome would be different if the action space were more complex. For that purpose, we extended the action space 100 times. That is, the new action space consists of 1200 possible moves to choose from -- 100 copies of each original action.

As shown in Figure \ref{fig:rubik_actions_x100}, the subgoal methods are barely affected by the change, while the low-level searches are unable to exceed $20\%$ success rate. That result confirms our proposition that when facing a complex action space, hierarchical methods offer considerably better performance.

According to our analysis, the primary issue with low-level searches in the augmented Rubik's cube is the lack of diversity of visited states. When for each state there are hundreds of actions that lead to a similar outcome, they are rated similarly by the policy. Hence, all the top actions essentially lead to the same outcome, which strongly limits the branching factor and trivializes the search trees. On the other hand, subgoal methods are not affected because subgoal generation does not depend on the action space. The conditional policy that connects the generated subgoals does not build a search tree, but always follows the single best action. Because of that, subgoal methods maintain their performance, even though the action space is much more complex.

It is also important to note that even though some state spaces may seem complex, the underlying manifold of possible configurations is in fact low-dimensional. For instance, we use 12x12 Sokoban boards, where each square is encoded as one-hot of 7 possible items, so technically the state space is 1008-dimensional, while there are only 4 actions. However, in practice the subgoal is defined by the positions of agent and boxes, which is at most 10-dimensional, hence rather simple to generate.

\subsection{Dead Ends}\label{appendix:properties_dead_ends}


\begin{wrapfigure}{r}{0.45\textwidth}
\vspace{-10pt}
\centering
\includegraphics[width=0.22\textwidth]{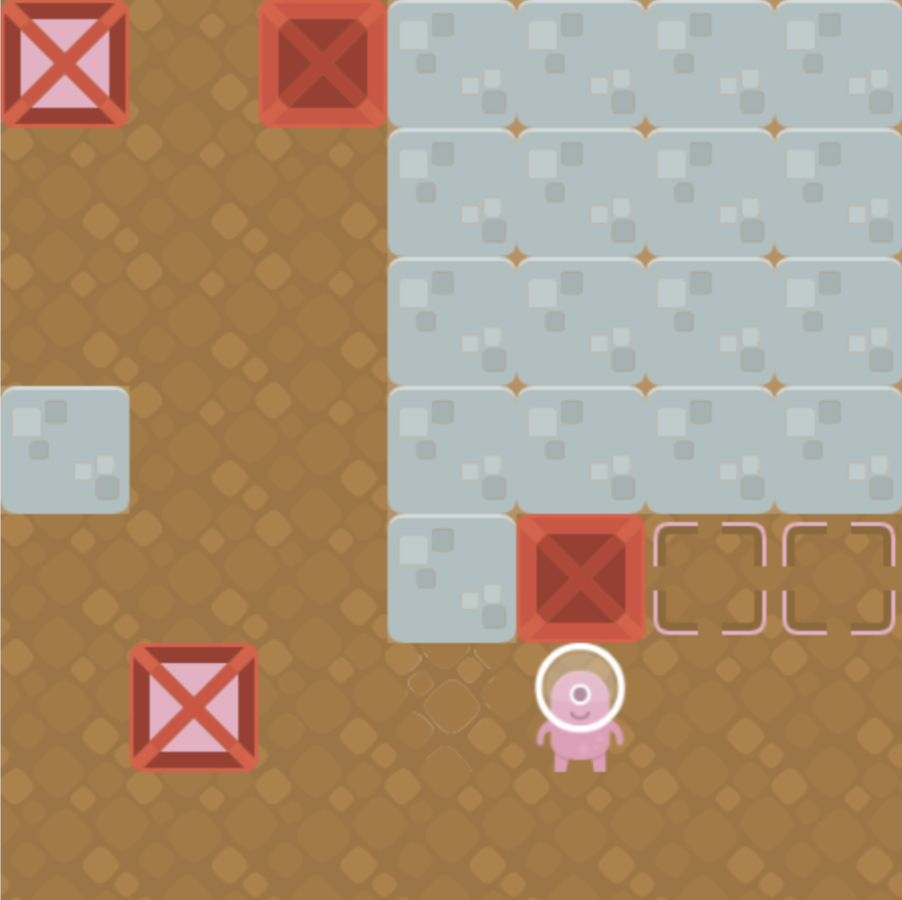}
\caption{An example dead-end in Sokoban -- a box that is pushed to the corner cannot be moved anymore, so the objective is not possible to achieve.
}
\label{fig:deadend}
\end{wrapfigure}

Dead-end states present a major challenge in decision-making and planning tasks. Once an agent encounters a dead end, reaching the goal becomes impossible, leading to wasted computational effort as the algorithm may continue exploring parts of the search space that do not contribute to solving the problem \citep{russel_norvig}. Failing to identify dead-ends may even lead to unsafe behavior \citep{DBLP:conf/nips/FatemiKSG21, DBLP:books/lib/SuttonB98}. At the same time, identifying dead-ends is NP-complete in many environments.

Specifically, a dead-end state $s$ is one from which there exists no feasible sequence of actions that leads to the goal state. Figure \ref{fig:deadend} shows an illustrative example of a dead-end state.



\subsubsection{Examples Of Dead-Ends In kSubS vs. BestFS}

\begin{figure}[h]
\centering
\includegraphics[width=0.6\textwidth]{ 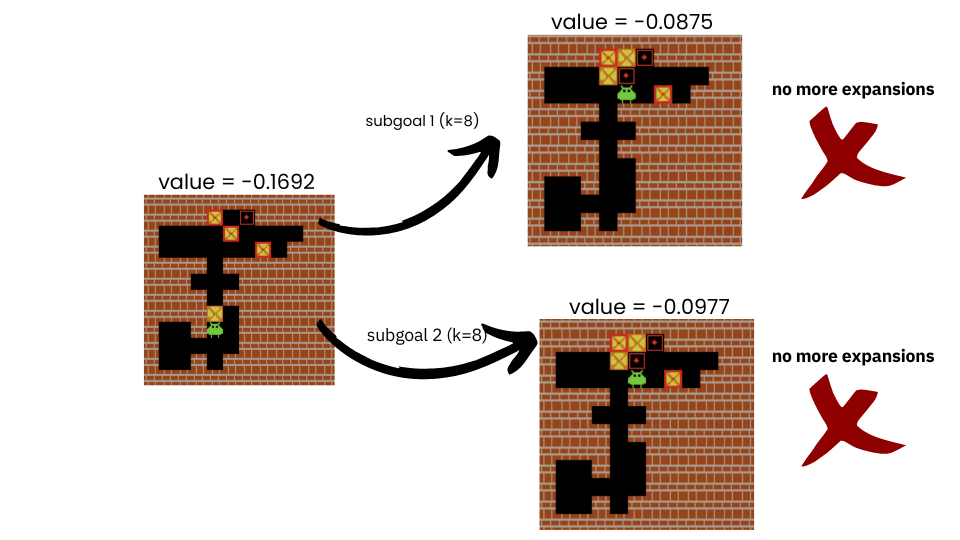}
\caption{We illustrate a scenario where the kSubS algorithm encounters dead-ends, hindering the search process. The figure shows a case where the algorithm generates two subgoals at an expected distance (k=8), but both lead to dead-ends, wasting a portion of the search budget (18 nodes). As a result, the kSubS algorithm backtracks from this subtree and continues searching elsewhere within the tree.}
\label{fig:ksubs_deadend}
\end{figure}

\begin{figure}[h]
\centering
\includegraphics[width=0.8\textwidth]{ 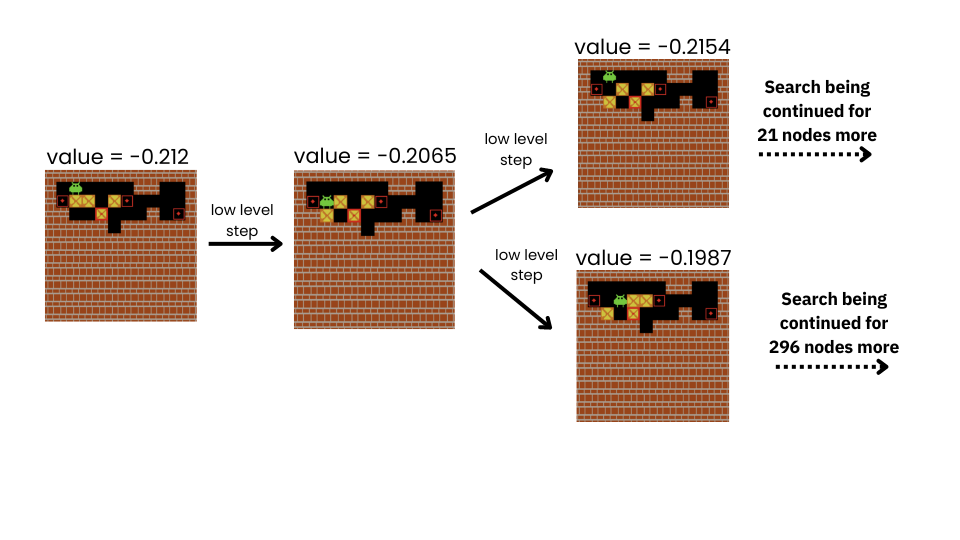}
\caption{The figure shows BestFS expanding two nodes from a dead-end. This resulted in the exploration of over 300 additional nodes from that state, ultimately failing to find a solution within the given search budget.}
\label{fig:bestfs_deadend}
\end{figure}

In this subsection, we present examples of how each method handles dead-end situations during the search process.

For this presentation, we analyzed 128 search trees initiated from identical starting boards for both algorithms. The kSubS algorithm encountered dead-ends in 3 instances. To resolve these, it navigated through 13 high-level nodes and 105 low-level nodes within the corresponding subtrees. In contrast, the BestFS algorithm encountered dead-ends in 18 instances, requiring the traversal of 4431 nodes.  Note that BestFS does not distinguish between high-level and low-level nodes in its search.

Examples of dead-end handling are shown in Figure \ref{fig:ksubs_deadend} for kSubS and Figure \ref{fig:bestfs_deadend} for BestFS. Observe that in the case showed in Figure \ref{fig:ksubs_deadend} expanding the parent node resulted in adding two more dead-ends to the search tree. Because they have higher values, they were immediately expanded. However, the subgoal generator understood that the only way to reach solution is to make an invalid transition of releasing the blocked box. Such subgoals cannot be achieved by the conditional policy, hence no more subgoal was created in that branch. On the other hand, low-level search is unable to propose invalid transitions, so it stays in dead-end until the value estimates are higher than for other branches.

\newpage

\section{Network Architectures \& Training Details}\label{appendix:hyperparameters}

\begin{table}[h]
    
    \centering
    \small
    \begin{tabular}{llllll}
    \hline
    Environment               & Hyperparameter & Generator & CLLP    & Value   & Policy \\ \hline
        \multirow{5}{*}{INT} & learning rate & 0.0001 & 0.0001 & 0.0003 & 0.0001 \\ 
& learning rate scheduling & linear & linear & linear & linear \\
 & warmup steps & 4000 & 4000 & 2000 & 4000 \\ 
 & batch size & 32 & 32 & 128 & 32 \\ 
 & weight decay & 1e-05 & 1e-05 & 1e-05 & 1e-05 \\ 
 & dropout & 0.1 & 0.1 & 0 & 0.1 \\ \hline
\multirow{5}{*}{Rubik's Cube} & learning rate & 0.0001 & 0.0005 & 3e-7 & 0.0001 \\ 
& learning rate scheduling & linear & linear & linear & linear \\
 & warmup steps & 5000 & 50000 & 50000 & 1000 \\ 
 & batch size & 512 & 5000 & 5000 & 2048 \\ 
 & weight decay & 0.0001 & 0.001 & 0.00001 & 0.0001 \\ 
 & dropout & 0.1 & 0 & 0 & 0 \\ \hline
\multirow{5}{*}{Sokoban} & learning rate & 0.00001 & 0.0001 &  0.0001  & 0.0001 \\ 
& learning rate scheduling & linear & linear & linear & linear\\
 & warmup steps & 2500 & 1000 & 1000 & 1000 \\ 
 & batch size & 512 & 2048 & 2048 & 2048 \\ 
 & weight decay & 0.0001 & 0.0001 & 0.0001 & 0.000001 \\ 
 & dropout & 0 & 0.1 & 0 & 0 \\ \hline
\multirow{5}{*}{N-Puzzle} & learning rate & 0.0001 & 0.0001 & 0.0001 & 0.0001 \\ 
& learning rate scheduling & linear & linear & linear & linear\\
 & warmup steps & 5000 & 2000 & 2000 & 2000 \\ 
 & batch size & 4096 & 4096 & 512 & 4096 \\ 
 & weight decay & 0.00001 & 0.00001 & 0.00001 & 0.0001 \\ 
 & dropout & 0.1 & 0 & 0 & 0 \\ \hline
        
    \end{tabular}
    \caption{Training-related hyperparameter values}
    \label{tab:training_hyperparameters}
\end{table}

We used BART \citep{bart} and BERT \citep{bert} architectures from HuggingFace Transformers for all components. Subgoal generators and INT's policies (CLLP and baseline policy) use BART. The remaining policies and value functions use BERT. Following the practice in \citep{ada}, we've reduced model size parameters, as detailed in Table \ref{tab:model_hyperparameters}.

\noindent
\textbf{INT}
As states in INT are complex objects, we prefer to use their string representations and avoid mapping arbitrarily generated strings into complex states. Requisite modifications to the component definition are best illustrated analogously to \ref{appendix:components}. A generator is redefined as follows:

$$ \mathcal{G}_{\text{int}}: \underbrace{\mathcal{S}}_{\text{state to expand}} \rightarrow \underbrace{P(\mathcal{T})}_{\text{set of \textit{proposed} subgoals (in string format)}} $$

and conditional level policy: 

$$ \mathcal{P}_{\text{int}}: \underbrace{\mathcal{S}}_{\text{current state}} \times \underbrace{\mathcal{T}}_{\text{subgoal \textit{representation}}} \rightarrow \underbrace{\mathcal{A}}_{\text{action}} $$

\noindent
\textbf{Sokoban } 
Unlike prior work \citep{ada, ksubs}, which used convolutional networks for all components, we work on tokenized representations of Sokoban boards and use BERT/BART architectures instead. This modification did not adversely impact our ability to replicate AdaSubS and kSubS results.

\noindent
\textbf{Training pipeline } We trained our models from scratch using the HuggingFace Transformer pipeline. Detailed training parameters, which varied across environments, can be found in \ref{tab:training_hyperparameters}.

\noindent
\textbf{Infrastructure } For training, we used a single NVIDIA A100 40GB GPU node, and each component's training took up to 48 hours. Because we used pre-trained trajectories, we did not need to use more than one core during training. We ran an evaluation using 24-core CPU jobs on Xeon Platinum 8268 nodes with 192GB of memory.

\begin{table}[H]
    
    \centering
    \small
    \begin{tabular}{llllll}
    \hline
    Environment               & Hyperparameter & Generator & CLLP    & Value   & Policy \\ \hline
        \multirow{5}{*}{INT} & d model & 512 & 512 & - & 512 \\ 
         & decoder layers & 6 & 6 & - & 6 \\ 
         & intermediate size & - & - & 256 & - \\ 
         & encoder attention heads & 8 & 8 & - & 8 \\ 
         & hidden size & - & - & 128 & - \\ 
         & num hidden layers & - & - & 2 & - \\ 
         & decoder ffn dim & 2048 & 2048 & - & 2048 \\ 
         & encoder ffn dim & 2048 & 2048 & - & 2048 \\ 
         & encoder layers & 6 & 6 & - & 6 \\ 
         & decoder attention heads & 8 & 8 & - & 8 \\ \hline
        \multirow{5}{*}{Sokoban} & d model & 256 & - & - & - \\ 
         & decoder layers & 3 & - & - & - \\ 
         & intermediate size & - & 512 & 128 & 512 \\ 
         & encoder attention heads & 4 & - & - & - \\ 
         & hidden size & - & 512 & 128 & 512 \\ 
         & num hidden layers & - & 6 & 1 & 6 \\ 
         & encoder ffn dim & 2048 & - & - & - \\ 
         & decoder ffn dim & 1024 & - & - & - \\ 
         & encoder layers & 3 & - & - & - \\ 
         & decoder attention heads & 4 & - & - & - \\ \hline
        \multirow{5}{*}{N-Puzzle} & d model & 64 & - & - & - \\ 
         & decoder layers & 3 & - & - & - \\ 
         & intermediate size & - & 128 & 128 & 256 \\ 
         & encoder attention heads & 4 & - & - & - \\ 
         & hidden size & - & 128 & 128 & 256 \\ 
         & num hidden layers & - & 2 & 1 & 3 \\ 
         & encoder ffn dim & 64 & - & - & - \\ 
         & decoder ffn dim & 64 & - & - & - \\ 
         & encoder layers & 3 & - & - & - \\ 
         & decoder attention heads & 4 & - & - & - \\ \hline
        \multirow{5}{*}{Rubik's Cube} & d model & 256 & - & - & - \\ 
         & decoder layers & 3 & - & - & - \\ 
         & intermediate size & - & 512 & 128 & 512 \\ 
         & encoder attention heads & 4 & - & - & - \\ 
         & hidden size & - & 512 & 128 & 512 \\ 
         & num hidden layers & - & 2 & 1 & 6 \\ 
         & encoder ffn dim & 2048 & - & - & - \\ 
         & decoder ffn dim & 1024 & - & - & - \\ 
         & encoder layers & 3 & - & - & - \\ 
         & decoder attention heads & 4 & - & - & - \\ \hline

    \end{tabular}
    \caption{Model-related hyperparameter values}
    \label{tab:model_hyperparameters}
\end{table}

\newpage

\section{Offline Pretraining}\label{appendix:pretraining}

Models are pretrained using an offline imitation learning approach. Specifically, given a set of solution trajectories $\{(s_0, s_1, \dots, s_{n_i})\}_{i=1}^{N}$ produced by an expert $\mathcal{M}$, or multiple experts $\{\mathcal{M}_j\}_{j=1}^{M}$ in cases where offline trajectories are collected from multiple experts, the objective is to learn from these trajectories. It is important to note that these trajectories are not required to be optimal; they may include loops or numerous redundant actions. Description of all components can be found in section $\ref{appendix:components}$ and supervised training objectives in section $\ref{appendix:pretrainig_objectives}$.

\subsection{Components} \label{appendix:components}

During the pretraining phase, models undergo an offline imitation learning process. Specifically, they are trained on a set of solution trajectories $\{(s_0, s_1, \dots, s_{n_i})\}_{i=1}^{N}$, which are collected to facilitate the learning of decision-making strategies.

\textbf{Generator} The generator component is responsible for generating subgoal propositions upon receiving a state. These propositions are designed to facilitate progress toward the solution by suggesting intermediate steps that direct the search process more efficiently.
$$ \mathcal{G}: \underbrace{\mathcal{S}}_{\text{state to expand}} \rightarrow \underbrace{P(\mathcal{S})}_{\text{set of subgoal propositions}} $$

\noindent
\textbf{Conditional Low-Level Policy} The Conditional Low-Level Policy (CLLP) plays a crucial role in node expansion by evaluating each subgoal proposition. For a given current state and a subgoal, the CLLP recommends actions that lead toward achieving the subgoal. A path from the current node to the subgoal is constructed through the iterative execution of these actions. Subgoals reached within a predefined number of steps, $k$, are incorporated into the graph, while those that are not are discarded.
$$ \mathcal{P}: \underbrace{\mathcal{S}}_{\text{current state}} \times \underbrace{\mathcal{S}}_{\text{subgoal state}} \rightarrow \underbrace{\mathcal{A}}_{\text{action}} $$

\noindent
\textbf{Value} The value function estimates the distance from a current state to the final solution. This estimation is used to guide the selection and expansion of nodes, influencing the overall search strategy.
$$ \mathcal{V}: \underbrace{\mathcal{S}}_{\text{state to evaluate}} \rightarrow \underbrace{\mathbb{R}}_{\text{value of the state}} $$

\noindent
\textbf{Behavioral Cloning Policy} The policy $\Pi_{\text{BC}}$ is a decision-making function that maps the current state to an action. It encapsulates the strategy derived from the learning process, guiding the agent's actions towards achieving the final goal.
$$ \Pi_{\text{BC}}: \underbrace{\mathcal{S}}_{\text{current state}}  \rightarrow \underbrace{\mathcal{A}}_{\text{action}} $$

\subsection{Supervised Objectives} \label{appendix:pretrainig_objectives}

Each expert trajectory is defined as a sequence of states and corresponding actions $(s_0, a_0), \ldots, (s_{n-1}, a_{n-1}), s_n$ that delineate a path to a solution. The training methodology leverages this data through several key self-supervised imitation mappings:

\begin{itemize}
    \item A $k$-subgoal generator that maps a state $s_i$ to a future state $s_{i+k}$, simulating the achievement of intermediate goals.
    \item A value function that estimates the remaining steps to the solution by mapping state $s_i$ to a numerical value $(i-n)$, representing the estimated distance from the goal.
    \item A policy that maps each state-action pair $(s_i, s_{i+d})$, with $d \leq k$, to the corresponding action $a_i$, thereby guiding the decision-making process towards the solution.
\end{itemize}


\newpage
\section{Offline Pretraining: Trajectories}

\subsection{Rubik's Cube}

\subsubsection{Random}

To construct a random successful trajectory, we performed 20 random permutations on an initially solved Rubik’s Cube and took the reverse of this sequence, replacing each move with its reverse. Such solutions are usually sub-optimal since random moves are not guaranteed to increase the distance from the solution. They can even make loops in the trajectories. However, a cube scrambled with 20 moves is usually close to a random state, so such trajectories give a decent space coverage.

\subsubsection{Beginner, CFOP}

\textit{Beginner} and \textit{CFOP} are algorithms commonly used by humans. They solve the cube by ordering the stickers layer by layer. Because of that, the solutions are highly structured and long -- usually between 100 and 200 moves. Both algorithms are composed of several subroutines that help building the consecutive layers. Thus, the structure of such solutions highly resembles the subgoal search.

\subsubsection{Kociemba}

The \textit{Kociemba two-stage solver} leverages the algebraic structure of the Rubik's Cube. In the first stage, its goal is to enter a specific subgroup. Since that subgroup is much smaller than the whole space, completing the solution may be done efficiently. \textit{Kociemba} finds reasonably short solutions (usually between 20 and 40 moves) and works reasonably fast.

\subsubsection{Size Of Datasets}

For training the components on a dataset collected by a single solver, we generate $100\,000$ trajectories. For the experiment with diverse experts, each solver generates $25\,000$ trajectories for a total of $100\,000$.

\subsection{INT}

Trajectories are constructed from sequences of axiom applications, similarly to \citep{ada}, who followed \citep{int}. A set of up to 15 (out of 18) axioms is first selected, and then a random axiom order is set and validated. Finally, a proof is converted to a relevant trajectory. Approximately 500,000 trajectories were generated for model pre-training.

We capped the number of axioms at 15 because some pairs of axioms (eg. terminal axioms) cannot be in one trajectory.

\subsection{N-Puzzle}
To collect data for N-puzzles, we utilized an algorithm that initially arranges block number $1$, followed by block number $2$, and so forth, as depicted in Figure \ref{fig:n_puzzle_env}. The training set comprises approximately $10,000$ trajectories.

\subsection{Sokoban}
To collect trajectories for Sokoban, we used a trained MCTS agent that gathered approximately $100,000$ trajectories.

\newpage
\section{Algorithms}\label{appendix:algorithms}

\subsection{Best-First Search}\label{appendix:algorithms_bestfs}


\paragraph{Overview} Best-First Search greedily prioritizes node expansions with the highest heuristic estimates, aiming for paths that likely lead to the goal. While not ensuring optimality, BestFS provides a simple yet efficient strategy for navigating complex search spaces. The high-level pseudocode for BestFS is outlined in Algorithm \ref{alg:high-level-bestfs}, and the detailed pseudocode is presented in Algorithm \ref{alg:BestFS_plain}.

\begin{minipage}{\linewidth} 
\centering
\begin{minipage}{0.5\linewidth} 
\centering
\begin{algorithm}[H]
\caption{Pseudocode for Best-First Search}
\label{alg:high-level-bestfs}
\begin{algorithmic}
\WHILE{has nodes to expand}
    \STATE Take node $N$ with the highest value
    \STATE Select children $n_i$ of $N$
    \STATE Compute values $v_i$ for the children
    \STATE Add $(n_i, v_i)$ to the search tree
\ENDWHILE
\end{algorithmic}
\end{algorithm}
\end{minipage}
\end{minipage}

\paragraph{Heuristic}\label{par:bestfs_heuristic} In our implementation, we adhere to the Best-First Search principle by utilizing the learned value function, a common practice in the planning domain \citep{heuristics_rubic, ksubs, ada, hips}. It should be noted that in each of our experiments, all the compared algorithms use the same value function network. This way we ensure that the differences come solely from the algorithmic part.

\paragraph{Selecting children}\label{par:bestfs_children} When expanding a node during search, the standard BestFS algorithm adds all its children. However, in our implementation, we aimed to reduce the search tree size by selecting only the most promising children. We achieve this by sorting the children according to their probability distribution predicted by the policy network. For choosing the final subset of children, we employ two approaches. In the simpler variant, we always select the top $k$ actions. In the second variant, we add top children until their cumulative probability exceeds a fixed threshold $t_{conf}$.

\begin{figure}[!ht]
    \centering
    \begin{minipage}[b]{0.45\textwidth}
        \centering
        \includegraphics[width=\textwidth]{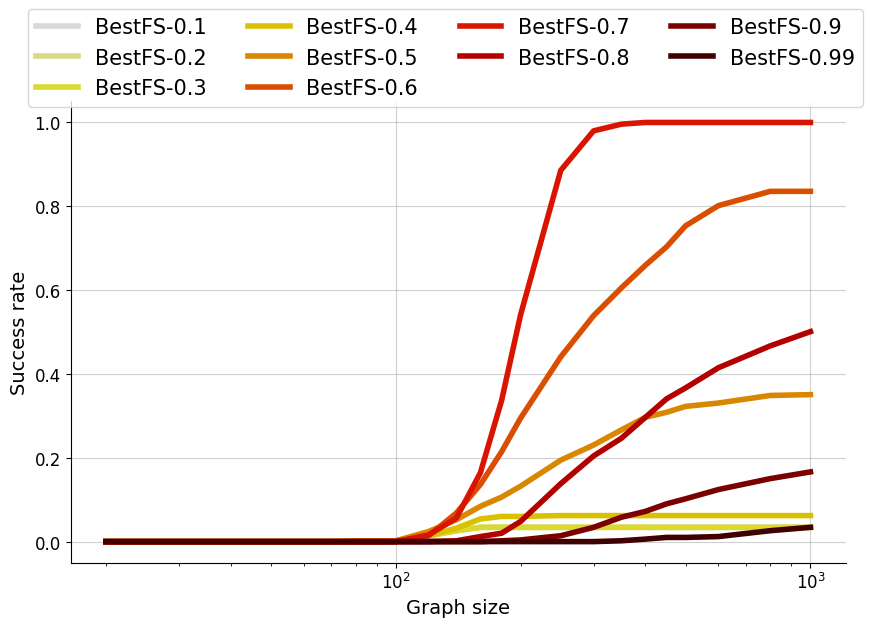}
    \end{minipage}
    \hspace{1cm} 
    \begin{minipage}[b]{0.45\textwidth}
    \centering
    \includegraphics[width=\textwidth]{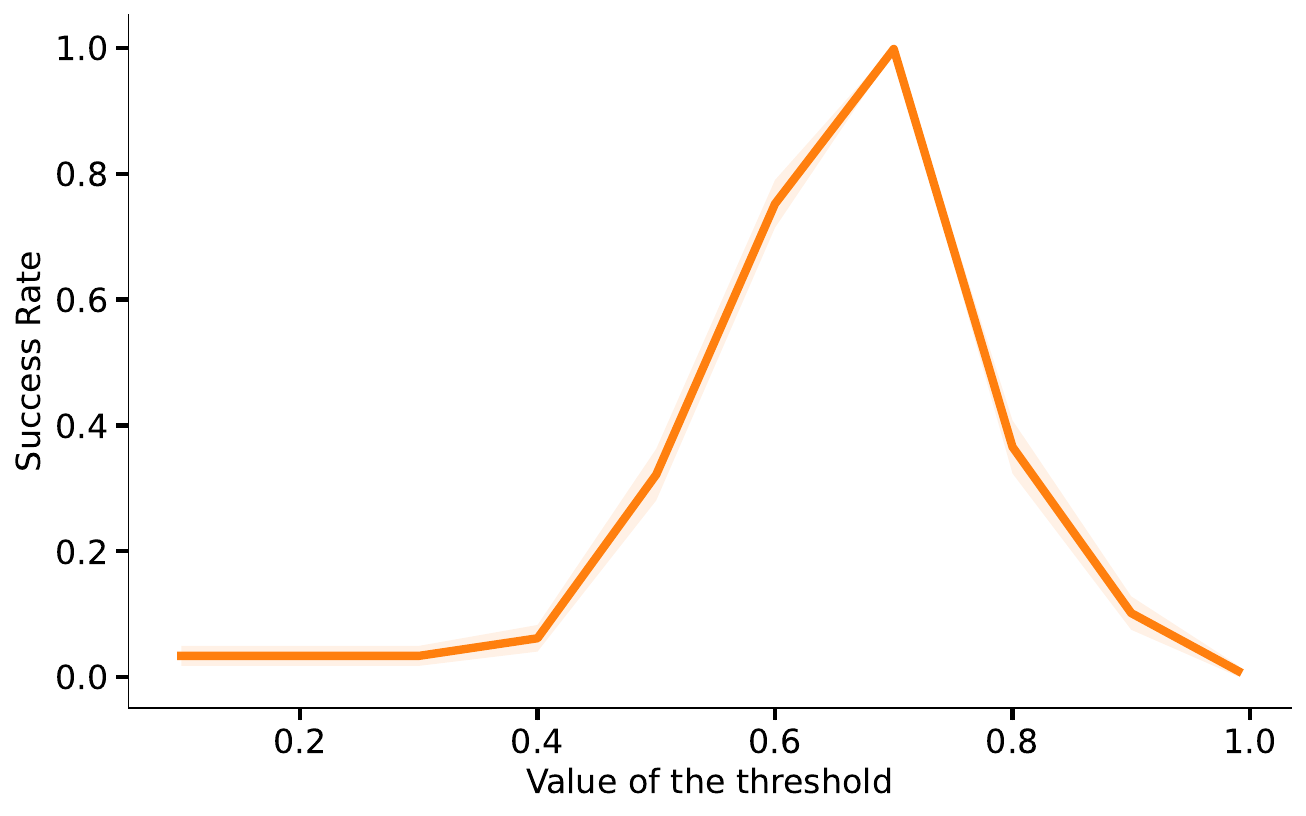}
    \end{minipage}
    \caption{Comparison of success rates for the BestFS algorithm on the Rubik's Cube with various confidence threshold values. BestFS-X represents the BestFS algorithm with the confidence threshold set to X. \textit{Left:} The plot displays the achieved success rate relative to the graph size. \textit{Right:} The plot illustrates the success rate for a budget of 500 nodes.}
    \label{fig:bestfs_confidence}
\end{figure}

This pruning does not adversely affect the standard algorithm, as nodes are still chosen based on their heuristic values, while the threshold sets a practical limit on the search space. Our results demonstrate that BestFS tends to perform much better with a confidence threshold (Figure \ref{fig:bestfs_confidence}). However, its performance is highly sensitive to this threshold as it balances exploration and exploitation, 
illustrating the impact of different confidence thresholds on success rates.

\paragraph{Completeness} In the Rubik's Cube environment with random trajectories, the subgoal methods solve more instances than BestFS given a low search budget, but with more resources, BestFS takes the lead (see Figure \ref{fig:rubik_random20}). Also, in other experiments, we may observe that BestFS typically requires higher computational budget to solve the simplest instances, but its performance increases considerably with more resources.

That behavior is related to the fact that the search trees built by hierarchical methods are much sparser because the branching occurs only in the high-level nodes. On the other hand, the low-level algorithms can cover a higher fraction of the space. On the extreme, if we used all the available actions for every expansion, the low-level search would be \textit{guaranteed} to find a solution if one exists. Our mechanism of selecting the actions removes that guarantee. However, at the same time, it drastically improves performance (compare BestFS-0.7 with BestFS-0.99 which is complete), which makes it a much better choice for our study.

We note that the high-level algorithms could be made complete, as proposed in \citep{hipseps, ada}. However, to maximize the efficiency we choose to keep the tested algorithms in their original form. The ability to search with sparse trees not only lets the methods advance fast, but also withdraw quickly if the branch does not lead to the solution (is a dead end).



\textbf{Hyperparameters} To identify the most suitable solving parameters, we used grid search. Initially we grid over coarse values (namely $0.1$, $0.2$, $0.3$, $0.4$,$0.5$, $0.6$, $0.7$, $0.8$, $0.9$, and $0.99$). Then we check finer values (with precision of $0.05$) around the best-performing threshold. The best-performing thresholds range from $0.6$ to $0.85$, depending on the environment and the components that are used.

For determining the best number of top actions $k$ for the simpler variant, we simply check every possible number of actions. Usually selecting $2$ actions is by far the best choice.

Details regarding hyperparameters of the networks are listed in Appendix \ref{appendix:components}.


\begin{algorithm}[H]
    \footnotesize
    \caption{Complete pseudocode for Best-First Search}
    \label{alg:BestFS_plain}
    \begin{algorithmic}
        \REQUIRE
        \STATE value function network $V$,
        \STATE policy $\rho_{BFS}$
        \STATE predicate of solution \textsc{solved}
        \STATE
        \FUNCTION{\textsc{search}}{$s_0$}
            \STATE $T \gets \emptyset$ \COMMENT{priority queue}
            \STATE $T$.\textsc{push}$((V(s_0), s_0))$
            \STATE \textit{parents} $\gets \{\}$
            \STATE \textit{seen}.\textsc{add}$(s_0)$ \COMMENT{\textit{seen} is a set}
            \vspace{5pt}
            \WHILE{$0 < \textsc{len}(T)$ \textbf{and} $\textsc{len}(\textit{seen}) < \textit{max\_budget}$}
                \STATE $\_, s \gets T.\textsc{extractMax}()$ \COMMENT{select node with the highest value}
                \STATE $\textit{actions} \gets \rho_{BFS}(s)$
                \vspace{5pt}
                \FOR{$a \textbf{ in } \textit{actions}$}
                    \STATE $s' \gets \textsc{envStep}(s, a)$
                    \vspace{5pt}
                    \IF{$s' \textbf{ in } \textit{seen}$}
                        \STATE \textbf{continue}
                    \ENDIF
                    \vspace{5pt}
                    \STATE $\textit{seen}.\textsc{add}(s')$
                    \STATE $\textit{parents}[s'] \gets s$
                    \STATE $T.\textsc{push}((V(s'), s'))$
                    \vspace{5pt}
                    \IF{\textsc{solved}$(s')$}
                        \STATE \COMMENT{solution found}
                        \STATE \textbf{return} \textsc{extractLowLevelTrajectory}($s'$, \textit{parents})
                    \ENDIF
                \ENDFOR
            \ENDWHILE
            \vspace{5pt}
            \STATE \textbf{return} $\mathtt{False}$ \COMMENT{solution not found}
    \end{algorithmic}
\end{algorithm}

\newpage
\subsection{Monte Carlo Tree Search}
\begin{figure}[H]
    \centering
    \includegraphics[width=0.85\linewidth]{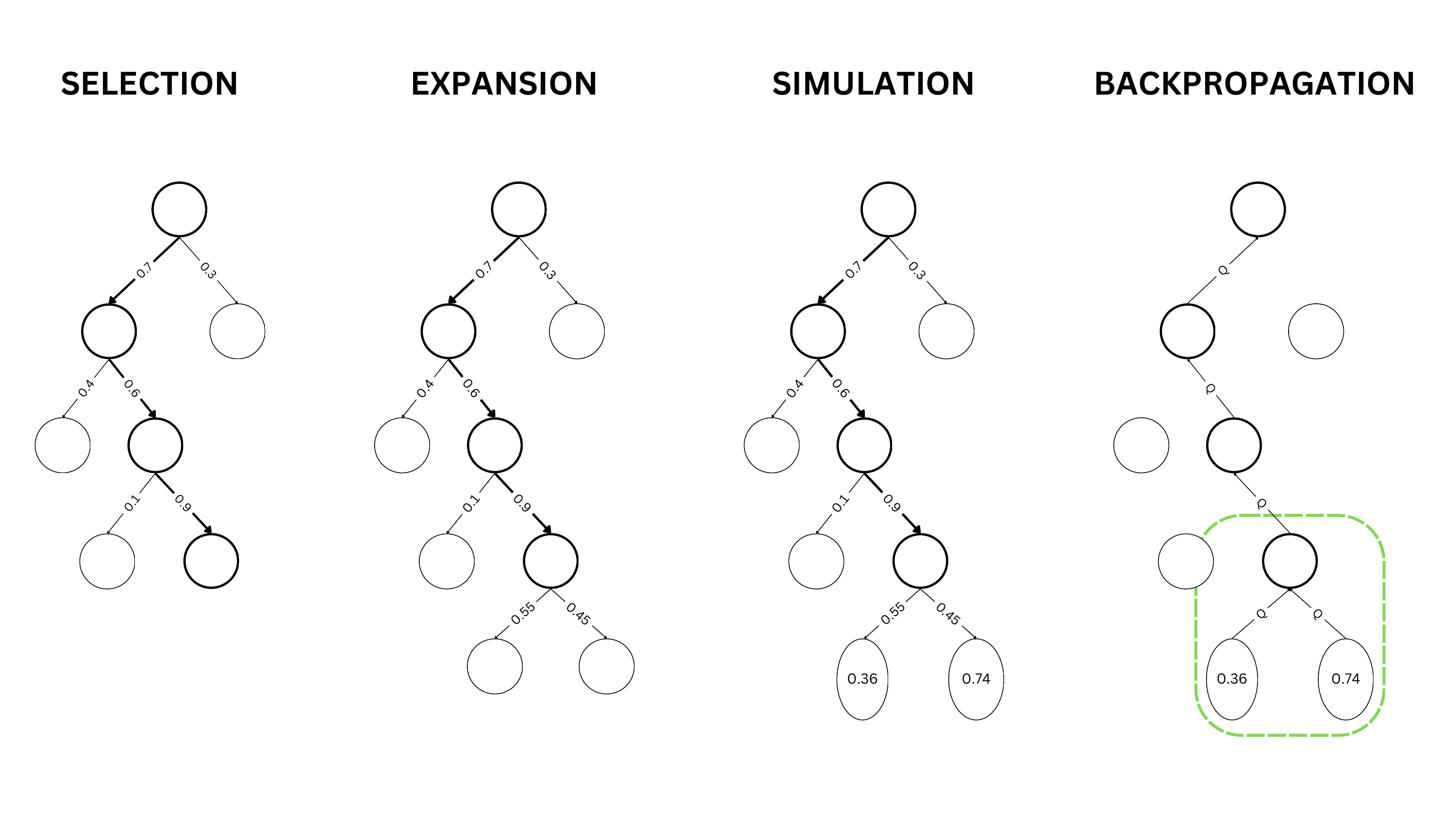}
    \caption{Schematic diagram of the MCTS algorithm in our implementation. Arrows show policy network probabilities and node values are valued network predictions. Q values, calculated via PUCT, integrate these with exploration-exploitation balance.}
    \label{fig:mcts-schema}
\end{figure}

\paragraph{Overview} Our Monte Carlo Tree Search (MCTS) solver, designed for a single-player setting, is based on the AlphaZero framework \citep{alphazero}. The high-level workflow of MCTS is illustrated in Figure \ref{fig:mcts-schema}, and detailed pseudocode is provided in Algorithm \ref{alg:SMCTS}.

The algorithm's operation consists of four primary stages:
\begin{itemize}
    \item \textbf{Selection}: The most promising node is selected using Polynomial Upper Confidence Trees (PUCT), augmented with an exploration weight to strike a balance between exploiting known strategies and investigating new pathways.
    \item \textbf{Expansion}: The selected node is expanded, generating new child nodes that correspond to prospective future actions. This expansion widens the search tree and enables the exploration of various outcomes.
    \item \textbf{Simulation}: Following the AlphaZero approach \citep{alphazero}, policy and value networks replace traditional simulations. The policy network suggests favorable moves, while the value network predicts their probability of success, directing the algorithm towards beneficial trajectories.
    \item \textbf{Backpropagation}: The insights derived from the networks are used to update node values, improving future decision-making.
\end{itemize}

\paragraph{Hyperparameters} In the MCTS algorithm, the parameters were set as follows: sampling temperatures were chosen from [0, 0.5, 1]. The number of steps varied between 200 and 1000, and the number of simulations ranged from 5 to 300. The discount factor and exploration weight were consistently set at 1.

\newpage

\begin{algorithm}[H]
\footnotesize
\caption{MCTS Solver}
\label{alg:SMCTS}
\begin{algorithmic}
    \REQUIRE
    \STATE Number of simulations: $N_s$
    \STATE Discount factor: $\gamma$
    \STATE Exploration weight: $c_{\text{puct}}$
    \STATE Sampling temperature: $\tau$
    \STATE Value function: $V$
    \STATE Environment model: $M$
    \STATE Initial state: $initial\_state$ from $\text{env}$
    \STATE
    \FUNCTION{\textsc{Search}}{($initial\_state$)}
        \STATE $root \gets initial\_state$
        \STATE $iteration \gets 0$
        \WHILE{$iteration < N_s$}
            \STATE $node \gets root$
            \WHILE{$node$ is not a leaf}
                \STATE $node \gets \textsc{selectChild}(node)$, according to PUCT formula
            \ENDWHILE
            \STATE $leaf \gets node$
            \STATE Expand the leaf using the environment model $M$, policy $\pi$, value function $V$, and discount factor $\gamma$
            \STATE Backpropagate results through the path to update $N, W, Q$
            \STATE $iteration \gets iteration + 1$
        \ENDWHILE
        \STATE $best\_child \gets \text{Sample child of the } root \text{ according to $\tau$ and $N$} $ 
        \STATE \textbf{return} action leading to $best\_child$
\end{algorithmic}
\end{algorithm}

\newpage
\subsection{A$^*$ Search} 
\paragraph{Overview} Like Best-First Search, A* prioritizes the exploration of promising nodes. However, A* strategically guides its search by incorporating both the actual cost to reach a node and a heuristic estimate of the remaining distance to the goal. This way it balances the greedy exploitation and conservative exploration. The high-level pseudocode for A* is outlined in Algorithm \ref{alg:high-level-astar}, and the detailed pseudocode is presented in Algorithm \ref{alg:AStar}.

\begin{minipage}{\linewidth} 
\centering
\begin{minipage}{0.5\linewidth} 
\begin{algorithm}[H]
\caption{Pseudocode for A* }
\label{alg:high-level-astar}
\begin{algorithmic}
\WHILE{has nodes to expand}
    \STATE Take node $N$ with the highest value
    \STATE Select children $n_i$ of $N$
    \STATE Compute values $v_i$ for the children
    \STATE Compute depth $d_i$ for the children
    \STATE Add $(n_i, \lambda d_i + v_i)$ to the search tree
\ENDWHILE
\end{algorithmic}
\end{algorithm}
\end{minipage}
\end{minipage}

\paragraph{Heuristic}
A* guidance is achieved through the following cost function:
$$f(node)=\lambda g(node)+h(node)$$
where:
\begin{itemize}
    \item $g(node)$: The cost to reach $node$ from the start state, in our case its depth in the search tree.
    \item $h(node)$: A heuristic estimate of the cost from $node$ to the goal state.
    \item $\lambda$: A scaling factor balancing the influence of actual cost and heuristic estimate. 
\end{itemize}

For heuristic $h$, we used a value network, like for BestFS (see Appendix \ref{par:bestfs_heuristic}). If the heuristic used for A* is \textit{admissible}, i.e. it never overestimates the cost of reaching the goal, A* is guaranteed to find an optimal solution. For instance, if we used $h(node)\equiv 0$, A* would reduce to the Dijkstra algorithm. The heuristic that we learn is not guaranteed to be admissible. Firstly, it estimates the distance according to the demonstrations, which is always an upper bound for the optimal distance. Secondly, the approximation errors introduce additional uncertainty. However, our main focus is on finding any solution, not necessarily an optimal one.

\paragraph{Selecting children}
During the search, A* maintains a priority queue of nodes to be explored. Similarly to BetsFS (Appendix \ref{par:bestfs_children}) for reducing the search tree size, we select the most promising children. At each iteration, the node with the lowest $f(node)$ value is selected for expansion. The algorithm proceeds until the goal state is reached or the computational budget is exceeded.

\begin{figure}[ht]
    \centering
    \begin{minipage}[b]{0.45\textwidth}
        \centering
        \includegraphics[width=\textwidth]{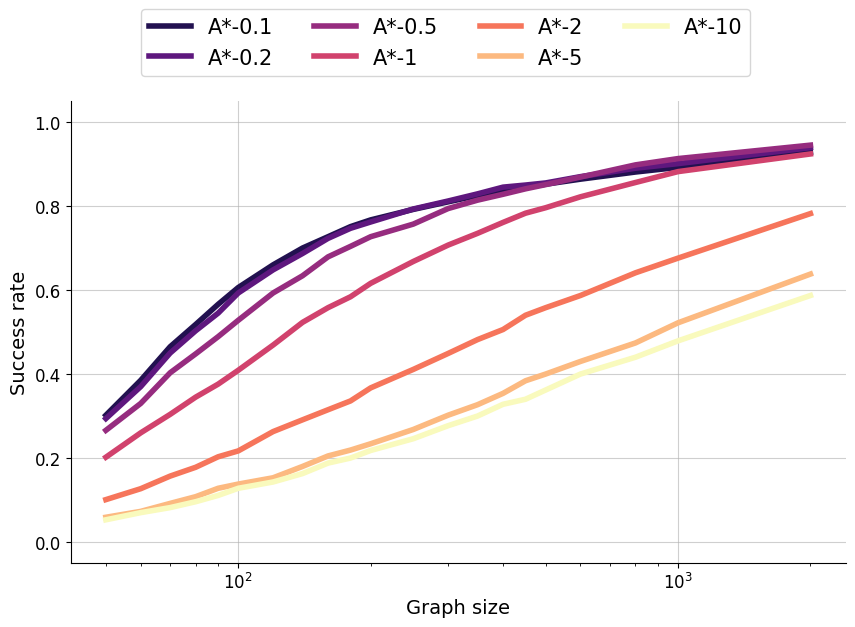}
    \end{minipage}
    \hspace{1cm} 
    \begin{minipage}[b]{0.45\textwidth}
    \centering
    \includegraphics[width=\textwidth]{ 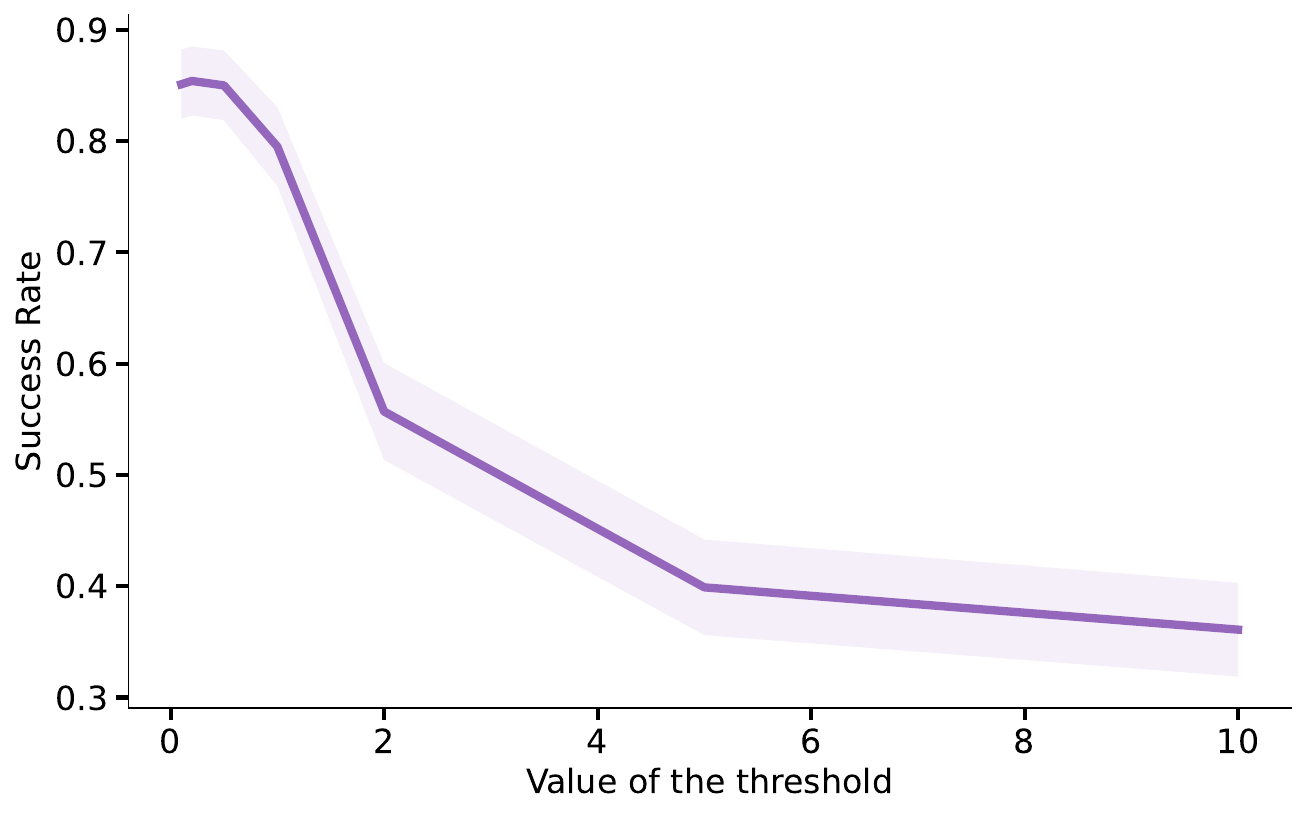}
    \end{minipage}   
    \caption{Figures presented above illustrate the impact of depth cost scaling on the overall success rate of the A* algorithm on Sokoban, employing a confidence threshold of 0.85. In most experiments, the smaller the depth scaling factor is, the better is the final success rate. The left figure shows the success rate curves for different choices of cost weight $\lambda$, while the right plot compares those variants for a fixed budget of 500 computation nodes.}
    \label{fig:astar_depth}
\end{figure}

\paragraph{Hyperparameters} The key parameter for A* is the cost weight $\lambda$. On the extreme, setting $\lambda=0$ reduces A* to greedy BestFS, while setting $\lambda=\infty$ makes it equivalent to Breadth-First Search. By tuning that parameter, we control the trade-off between exploration and exploitation of the search.

To tune the depth parameter for our experiments, we grided over values $[0.1, 0.2, 0.5, 1, 2, 5, 10]$. However, usually the best choice was to keep the cost weight low ($0.1$ or $0.2$, see Figure \ref{fig:astar_depth}). While conservative search allows A* avoid more dead-ends than BestFS (see Figure \ref{tab:dead_ends}), usually greedy steps lead to finding the solution much faster.

\begin{algorithm}[H]
    \footnotesize
    \caption{Complete pseudocode for $A^*$ Search}
    \label{alg:AStar}
    \begin{algorithmic}
        \REQUIRE
        \STATE value function network $V$
        \STATE policy $\rho_{BFS}$
        \STATE predicate of solution \textsc{solved}
        \STATE depth scaling factor $\lambda$
        \STATE
        \FUNCTION{\textsc{search}}{$s_0$}
            \STATE $T \gets \emptyset$ \COMMENT{priority queue}
            \STATE $T$.\textsc{push}$((V(s_0), s_0))$
            \STATE \textit{parents} $\gets \{\}$
            \STATE \textit{seen}.\textsc{add}$(s_0)$ \COMMENT{\textit{seen} is a set}
            \vspace{5pt}
            \WHILE{$0 < \textsc{len}(T)$ \textbf{and} $\textsc{len}(\textit{seen}) < \textit{max\_budget}$}
                \STATE $\_, s \gets T.\textsc{extractMax}()$ \COMMENT{select node with the highest value}
                \STATE $\textit{actions} \gets \rho_{BFS}(s)$
                \vspace{5pt}
                \FOR{$a \textbf{ in } \textit{actions}$}
                    \STATE $s' \gets \textsc{envStep}(s, a)$
                    \vspace{5pt}
                    \IF{$s' \textbf{ in } \textit{seen}$}
                        \STATE \textbf{continue}
                    \ENDIF
                    \vspace{5pt}
                    \STATE $\textit{seen}.\textsc{add}(s')$
                    \STATE $\textit{parents}[s'] \gets s$
                    \STATE $T.\textsc{push}((V(s') - \lambda\cdot \textit{depth$(s')$}, s'))$
                    \vspace{5pt}
                    \IF{\textsc{solved}$(s')$}
                        \STATE \COMMENT{solution found}
                        \STATE \textbf{return} \textsc{extractLowLevelTrajectory}($s'$, \textit{parents})
                    \ENDIF
                \ENDFOR
            \ENDWHILE
            \vspace{5pt}
            \STATE \textbf{return} $\mathtt{False}$ \COMMENT{solution not found}
    \end{algorithmic}
\end{algorithm}

\newpage

\subsection{kSubS And AdaSubS}
\paragraph{Overview}
AdaSubS is a hierarchical search algorithm designed to solve combinatorial problems by operating on high-level nodes, which represent multiple steps rather than single actions. It employs multiple generators $\mathcal{G}_{k_1}, \mathcal{G}_{k_2}, \dots, \mathcal{G}_{k_m}$ to generate subsequent subgoals, a value function $\mathcal{V}$ to estimate the distance from a given state to the solution, and a conditional low-level policy $\mathcal{P}$ to execute a series of actions leading from one subgoal to the next. kSubS is a special case of AdaSubS, where only a single generator is used. These methods are introduced and studied in \citep{ksubs, ada}.

\paragraph{Stages}
The method begins by adding $m$ initial nodes (one per each generator) to a priority queue, where each initial node $i$ is assigned a priority $(k_i, \mathcal{V}(s_0))$. Here, $k_i$ is the length of the generator used during the node's expansion, and $\mathcal{V}(s_0)$ estimates the distance (in low-level actions) between $s_0$ and the solution. The following steps are repeated until a solution is found or the budget is exhausted:
\begin{itemize}
    \item \textbf{Selection for expansion}: The node $\left((k, \mathcal{V}(s), s\right)$ with the highest priority is extracted from the queue. This priority structure ensures that the algorithm prioritizes expanding the longest subgoals whenever possible.
    \item \textbf{Generating subgoals}: The current state $s$ is passed to the selected generator $\mathcal{G}_{k}$, which produces multiple subgoal propositions represented as states $s^*_1, s^*_2, \dots, s^*_p$.
    \item \textbf{Verifying reachability}: Since $\mathcal{G}_{k}$ can produce invalid or unreachable subgoals, each proposed subgoal must be verified. The conditional low-level policy $\mathcal{P}$ begins an iterative process, taking single steps from $s$ towards the proposed subgoal $s^*_j$. If $s^*_j$ is reached within $k$ steps, the subgoal is accepted, and new high-level nodes $\{\left( (k_i, \mathcal{V}(s^*_j)), s^*_j \right)\}_{i \in \{1 \dots m\}}$ are added to the priority queue as potential future subgoals to expand.
\end{itemize}

For a graphical overview of how AdaSubS works, see Appendix \ref{appendix:hierarchical_search}.

\begin{algorithm}[H]
\footnotesize
    \caption{Complete pseudocode for Adaptive Subgoal Search}
    \label{alg:adasubs}
\begin{algorithmic}
    \REQUIRE
    \STATE $C_1$ max number of nodes,
    \STATE $V$ value function network,
    \STATE $\rho_{k_0}, \ldots, \rho_{k_m}$ subgoal generators,
    \STATE $\textsc{Solved}$ predicate of solution
    \STATE
    \FUNCTION{\textsc{Solve}}{($s_0$)}
        \STATE $T \gets \emptyset$ \COMMENT{priority queue with lexicographic order}
        \STATE $parents \gets \{\}$
        \FOR{$k$ in $k_0, \ldots, k_m$}
            \STATE $T.push((k, V(s_0)), s_0)$
        \ENDFOR
        \STATE $seen.add(s_0)$ \COMMENT{$seen$ is a set}
        \WHILE{$0 < \text{len}(T) \textbf{ and } \text{len}(seen) < C_1$}
            \STATE $(k, \_), s \gets T.extract\_max()$
            \STATE $subgoals \gets \rho_k(s)$
            \FOR{$s'$ \textbf{ in } $subgoals$}
                \IF{$s'$ \textbf{ not in } $seen$}
                    \IF{\textsc{Is\_Valid}(s, s')}
                        \STATE $seen.add(s')$
                        \STATE $parents[s'] \gets s$
                        \FOR{$k$ in $k_0, \ldots, k_m$}
                            \STATE $T.push((k, V(s')), s')$
                        \ENDFOR
                        \IF{\textsc{Solved}(s')}
                            \STATE \textbf{return} \textsc{ExtractLowLevelTrajectory}(s', parents)
                        \ENDIF
                    \ENDIF
                \ENDIF
            \ENDFOR
        \ENDWHILE
        \STATE \textbf{return} $\mathtt{False}$
\end{algorithmic}
\end{algorithm}

\newpage
\subsection{HIPS And HIPS-$\varepsilon$}

Here we show a pseudocode for HIPS and HIPS-$\varepsilon$ methods. For details see Alg. \ref{alg:ExtendedHIPSVaePHS}

\begin{algorithm}[H]
\footnotesize
    \caption{Complete pseudocode for HIPS with BestFS-PHS* and VQ-VAE}
    \label{alg:ExtendedHIPSVaePHS}
\begin{algorithmic}
    \REQUIRE
    \STATE $C_1$ max number of nodes,
    \STATE $VAE$ Variational Autoencoder for subgoal generation,
    \STATE $\textsc{Solved}$ predicate of solution,
    \STATE $\epsilon$ exploration parameter for balancing,
    \STATE $V$ value function for PHS* cost estimation
    \STATE
    \FUNCTION{\textsc{Extended\_HIPS\_Solve}}{($s_0$)}
        \STATE Initialize search data structures, including priority queues.
        \STATE $seen.add(s_0)$ \COMMENT{Track seen states}
        \WHILE{search conditions are met}
            \STATE Use PHS* search strategy to select a state $s$.
            \STATE Generate subgoals $subgoals \gets VAE(s)$.
            \FOR{each $s'$ in $subgoals$}
                \IF{$s'$ not seen and is valid}
                    \STATE Evaluate $s'$ using $V$ for PHS* cost.
                    \STATE Update priority queue based on PHS* cost.
                    \IF{\textsc{Solved}$(s')$}
                        \STATE \textbf{return} Construct solution path.
                    \ENDIF
                \ENDIF
            \ENDFOR
        \ENDWHILE
        \STATE \textbf{return} $\mathtt{False}$ \COMMENT{Solution not found}
\end{algorithmic}
\end{algorithm}

\newpage
\section{Statistical Analysis Of High-Level And Low-Level Algorithms}

\begin{table}[ht]
\centering
\small
\begin{tabular}{p{2.2cm} p{1.8cm} p{1.5cm} p{1.3cm} p{1.5cm} p{1.5cm} p{1.5cm}}
\hline
Environment               & Algorithm & Tree size & Number of leaves & Branching factor   & Solution length & Solution length (subgoals)   \\ \hline
    \multirow{5}{*}{N-Puzzle} & BestFS & $354.43$ & $1.34$ & $1.0$ &  $354.08$ &  - \\ 
     & A* & $354.09$ & $1.34$ & $1.0$ & $353.56$ &  - \\ 
     & MCTS & $742.04$ & $371.52$ & $2.0$ & $347.43$ &  - \\ 
     & kSubS-8 & $353.66$ & $1.0$ & $1.0$ & $353.66$ & $45.67$  \\ \hline
    \multirow{5}{*}{Sokoban} & BestFS & $185.24$ & $36.88$ & $1.22$ & $48.98$ &  - \\ 
     & A* & $85.04$ & $12.22$ & $1.43$ & $45.68$ &  - \\ 
     & MCTS & $255.0$ & $128.0$ & $2.0$ & $45.1$ & -  \\ 
     & kSubS-8 & $101.92$ & $6.6$& $1.06$& $46.88$ & $7.23$ \\ \hline
    \multirow{5}{*}{Rubik's Cube} & BestFS & $152.25$ & $58.02$ & $1.65$ & $48.92$ & -  \\ 
     & A* & $185.23$ & $69.57$ & $1.64$ & $45.46$ & -  \\ 
     & MCTS & $716.46$ & $358.73$ & $2.0$ & $33.32$ &  - \\ 
     & kSubS-4 & $303.52$ & $133.44$ & $1.12$ & $73.58$ &  $26.65$ \\  \hline

\end{tabular}
\caption{Average values of tree size, number of leaves, branching factor (average number of children), and solution length were calculated for 100 boards solved by all presented algorithms. Additionally, for the subgoal method, the average number of subgoals on the winning path was determined. }
\label{tab:statistics}
\end{table}

\begin{figure}[H]
    \centering
    \begin{minipage}[b]{0.58\textwidth}
        \centering
        \includegraphics[width=0.99\textwidth]{ 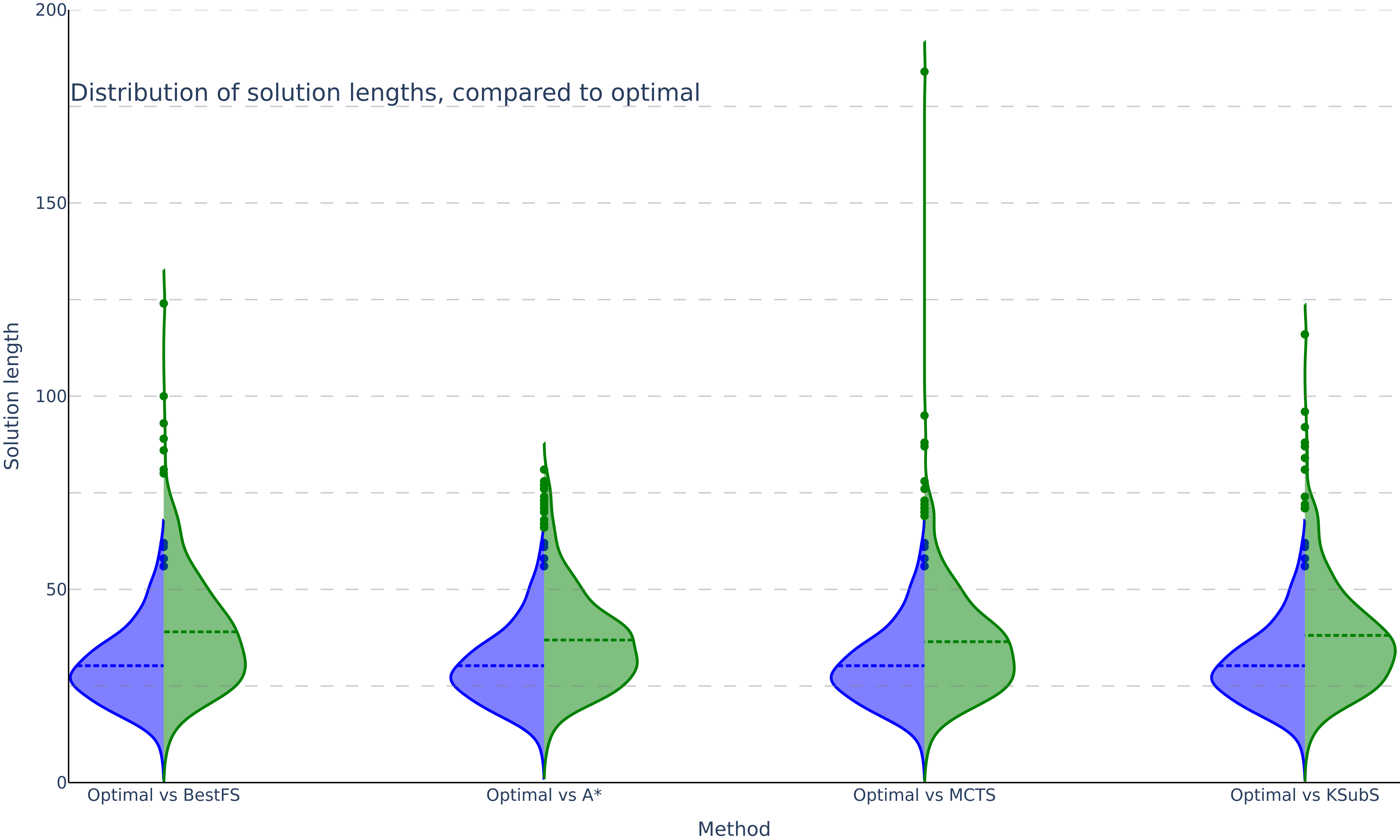}
        \caption{The distribution of solution length in Sokoban. The right part of each plot illustrates the distribution for the methods that we used. The left part corresponds to the optimal solutions for the tested instances obtained using Breadth-First Search. These algorithms were evaluated on 494 commonly solved instances.}
        \label{fig:solution_length_violin}
    \end{minipage}
    \hspace{0.5cm} 
    \begin{minipage}[b]{0.37\textwidth}
        \centering
        \includegraphics[width=0.99\textwidth]{ 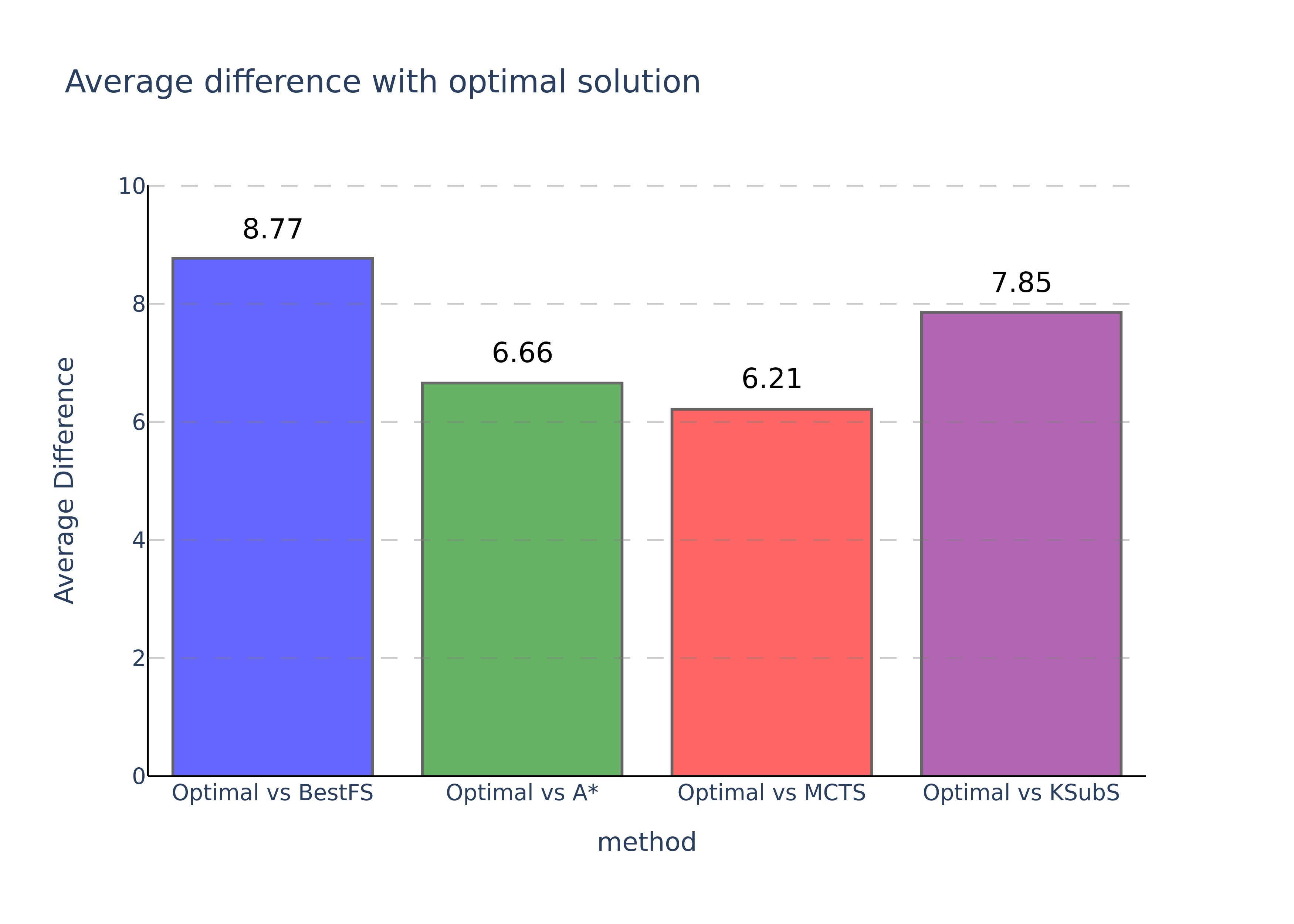}
        \caption{The average difference between the solutions found by each algorithm and the optimal solutions for the Sokoban environment. These algorithms were evaluated on 494 commonly solved instances.}
        \label{fig:solution_length_bar}
    \end{minipage}
\end{figure}

\newpage
\section{Hierarchical Search}\label{appendix:hierarchical_search}

\begin{figure}[ht]
    \centering
    \includegraphics[width=0.8\textwidth]{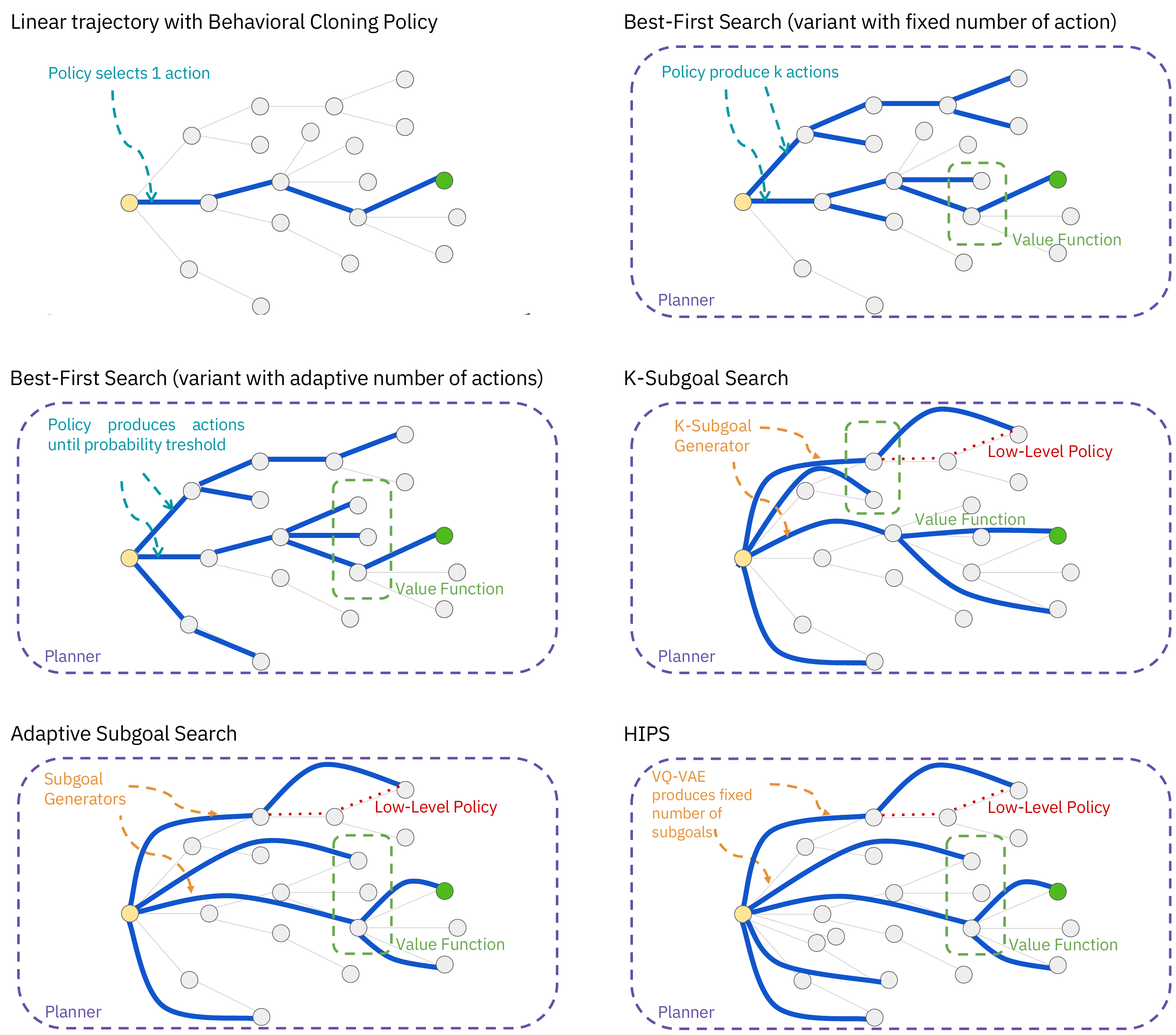}
    \caption{Overview of the search methods under consideration, accompanied by illustrative examples depicted in various plots for each method. Specifically, straight blue lines are utilized to represent low-level actions that occur within the search space. In contrast, long skip connections are used to symbolize subgoals within the search process.
}
    \label{fig:hierarchical_search}
\end{figure}

\newpage
\section{Further Discussion On HIPS Results}\label{appendix:hips}

HIPS and HIPS-$\varepsilon$ \citep{hips,hipseps} are recent hierarchical search algorithms proposing to generate subgoals with variational autoencoders. We attempted to use HIPS and HIPS-$\varepsilon$ in greedy and prior-informed variations, and for all HIPS methods, the cost of inference was prohibitively high.

To compare these methods, we used A*-generated data from HIPS papers, in contrast to all other experiments (which use data generated by us).

\begin{figure}[ht]
    \centering
    \begin{minipage}[b]{0.4\textwidth}
        \centering
        \includegraphics[width=\textwidth]{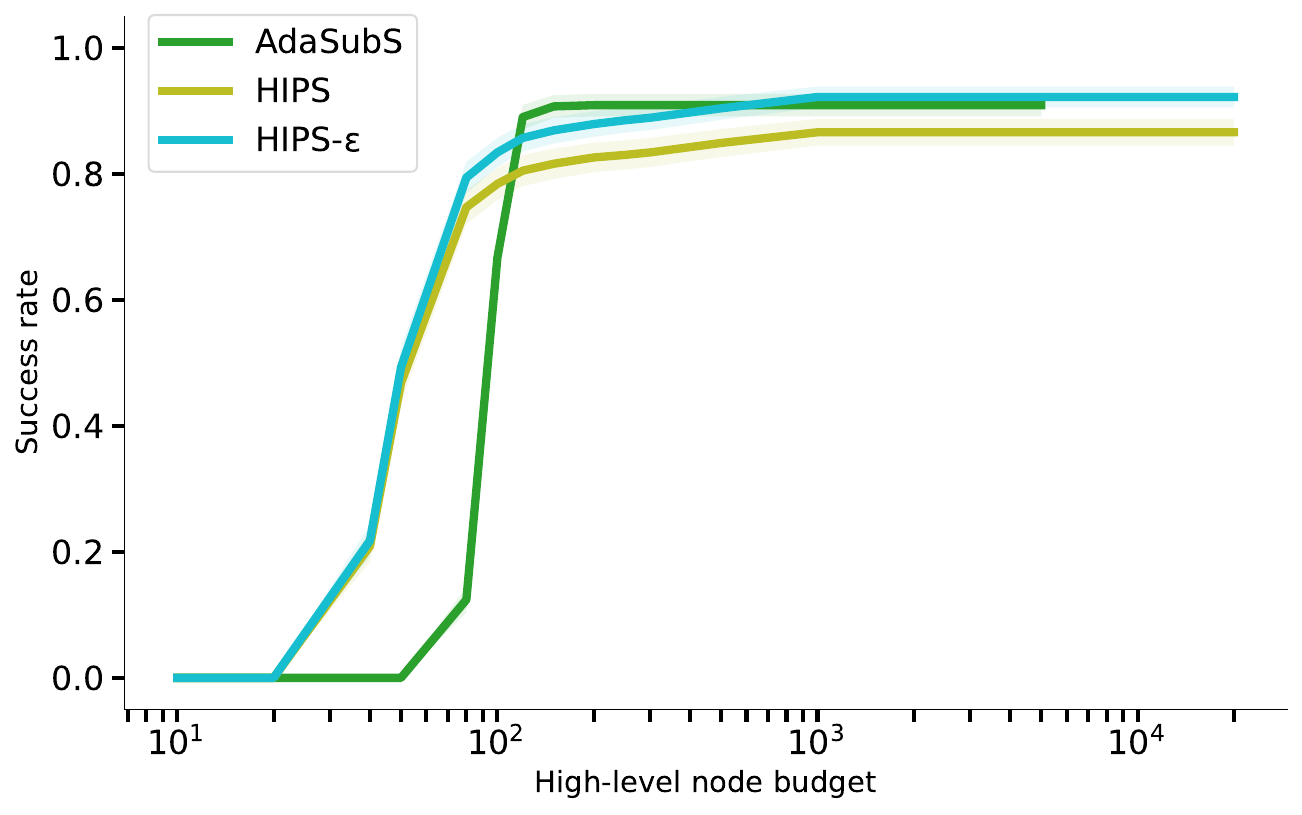}
    \end{minipage}
    \hspace{1cm} 
    \begin{minipage}[b]{0.4\textwidth}
           \centering
\includegraphics[width=\textwidth]{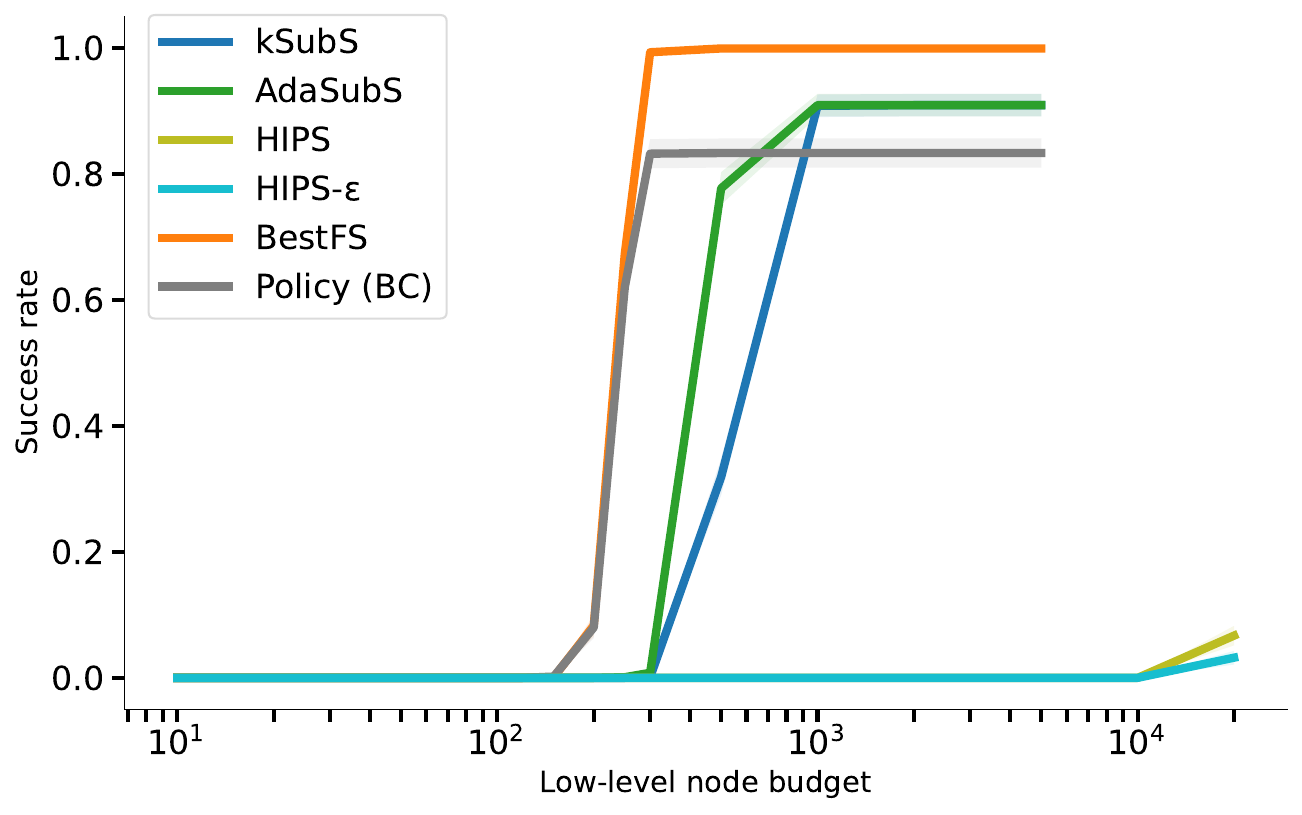}
    \end{minipage}
        \caption{A comparison of high-level and low-level node budgets for considered methods: HIPS, subgoal search methods, and baselines on N-Puzzle. The low-level node budget represents the number of all states that have ever been visited during the search. The bimodal distribution indicates that HIPS methods use disproportionately (over 100x) more low-level nodes than comparable subgoal search methods and baselines. This directly translates to prohibitively slow solving time.}
    \label{fig:hips_full}
\end{figure}

Our evaluation, illustrated in \ref{fig:hips_full}, shows that HIPS uses 100x more low-level nodes in search than comparable subgoal search methods and baselines - despite relatively similar subgoal efficiency as calculated in relevant papers. These findings informed our decision not to evaluate HIPS in the rest of the paper.

\newpage
\section{Common Pitfalls In Hierarchical Search evaluations}\label{appendix:pitfalls}

In this study, one of our primary goals is to identify common but often overlooked pitfalls in evaluating hierarchical search methods, which can lead to misleading conclusions. Based on our findings, we propose a set of guidelines that help ensure meaningful and consistent comparisons across different methods. We observed that the nature of hierarchical search makes it easy, whether intentionally or not, to present results in a way that favors certain methods, often without readers being aware. In this section, we present key insights on this issue, with an emphasis on the following evaluation guidelines:

\begin{itemize}
\item Report results using a \textit{complete search budget}.
\item Include \BestFS{} with a confidence threshold as a baseline.
\item Ensure careful tuning of the confidence threshold.
\item Use up-to-date code for running experiments.
\end{itemize}

\subsection{Complete Search Budget}
We define the performance metric in terms of \textit{success rate}, which is the percentage of problem instances solved within a specified \textit{complete search budget}. This budget refers to the total number of states visited during the search process. For hierarchical methods, this includes both the subgoals generated and the states visited by the low-level policies connecting those subgoals.

Reporting the \textit{complete search budget} is crucial, as opposed to the \textit{sparse search budget}, which counts only the high-level nodes in the search tree. As discussed in Appendix \ref{appendix:hips}, \cite{hips} rely on the sparse search budget for their evaluations. This creates a misleading impression that HIPS outperforms low-level baselines, while in reality, it requires significantly more computational effort to solve the same problems.

To illustrate this issue, consider a simple environment where an agent must navigate a 100x100 empty room to reach a goal on the opposite side. In this case, a hierarchical method may require only a single subgoal -- directly corresponding to the goal state -- while a low-level method, even if following the optimal path, would require at least 100 steps. A sparse search budget would misleadingly indicate that the hierarchical method solves the task in one step, while the low-level approach requires 100 steps, implying a 100x higher cost. However, both methods traverse the same path, making this comparison inaccurate. Using the \textit{complete search budget}, both methods would be assigned the same cost, providing a much more meaningful comparison.

\begin{figure}[h]
    \centering
    \begin{minipage}[t]{0.48\linewidth} 
        \centering
        \includegraphics[width=\linewidth]{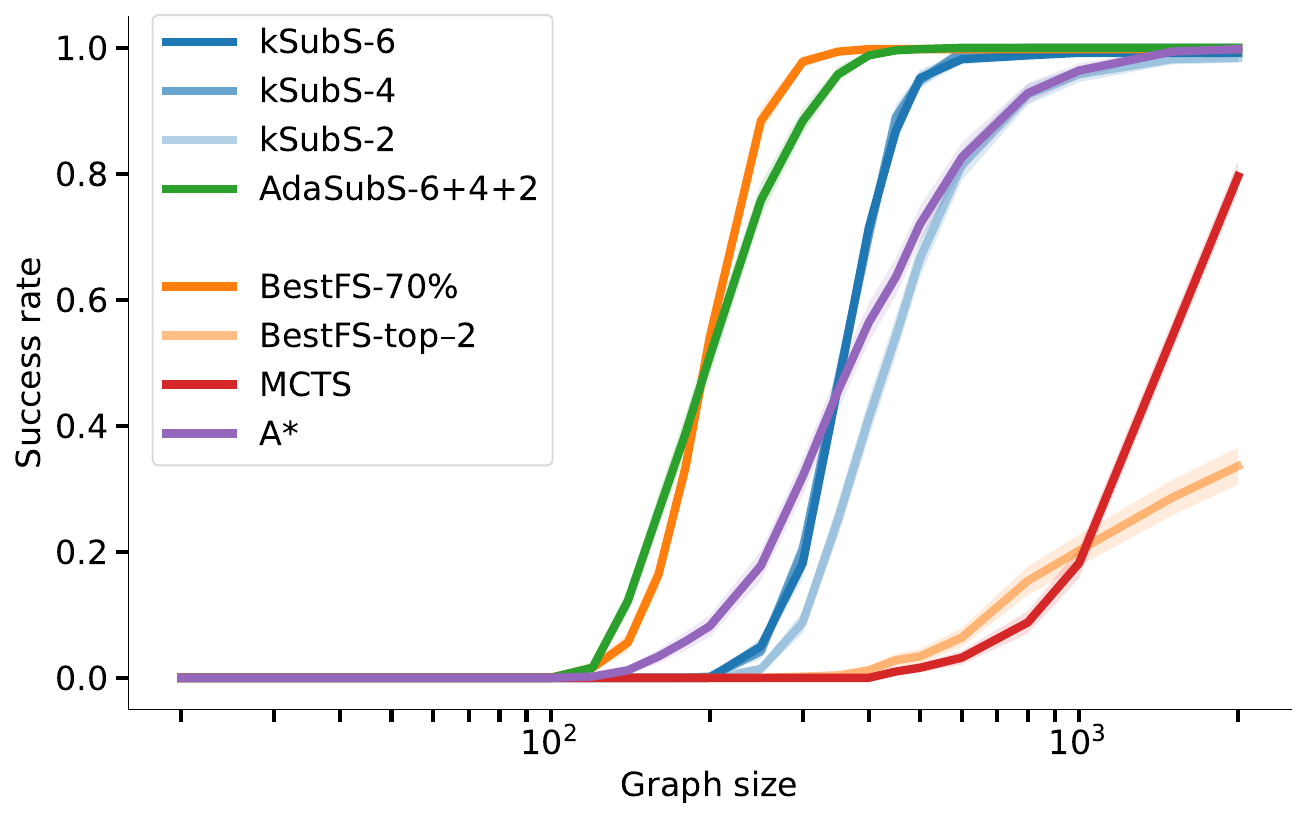}
        \caption{Solving the Rubik's Cube. The light orange line represents the best-performing variant of BestFS that selects a fixed number of actions for each expansion. The solid orange line represents BestFS with actions confidence threshold, which is much more efficient.}
        \label{fig:rubik_top2_pitfall}
    \end{minipage}
    \hfill 
    \begin{minipage}[t]{0.48\linewidth} 
        \centering
        \includegraphics[width=\linewidth]{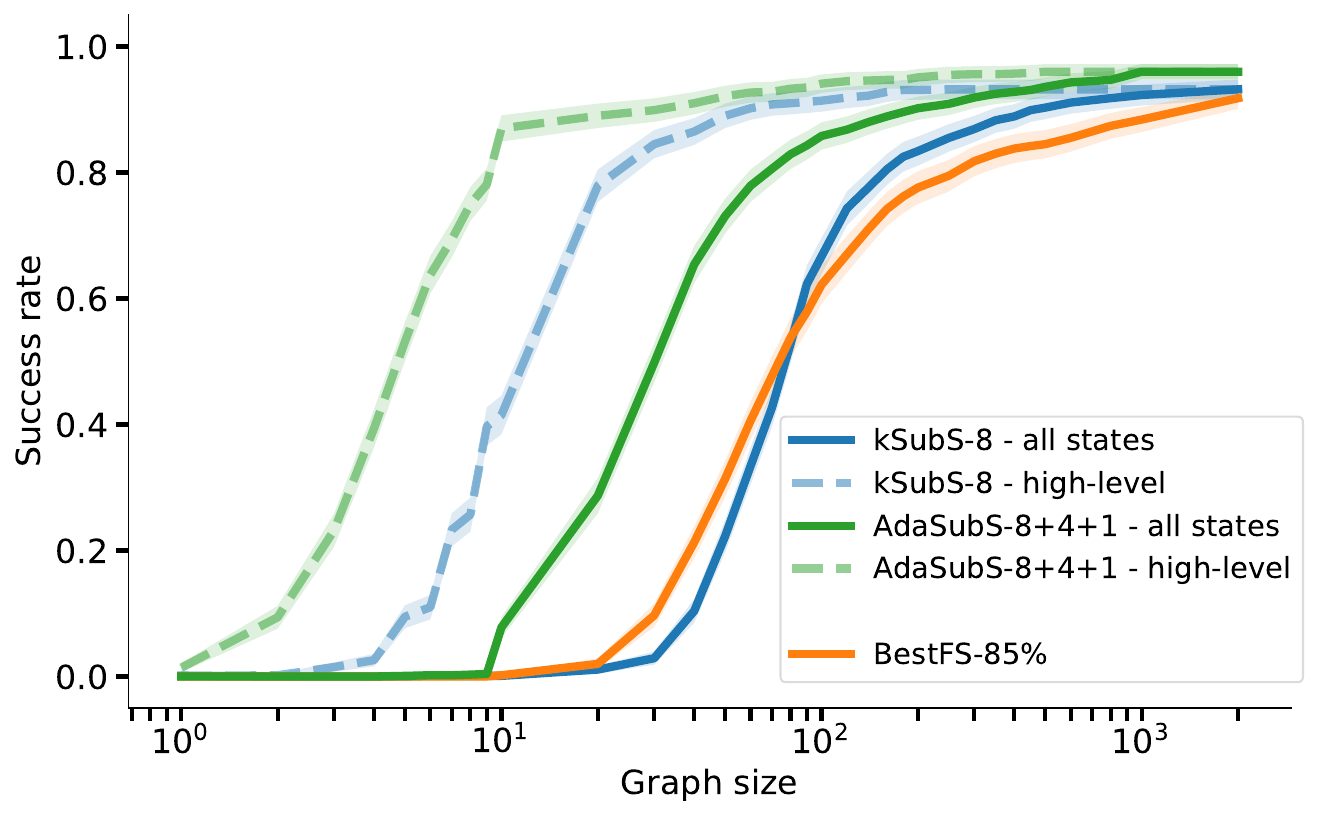}
        \caption{Solving Sokoban. Solid lines correspond to using \textit{complete search budget} as the search tree size metric. Dashed lines correspond to the same runs, but using \textit{sparse search budget} as the search tree size metric. For BestFS, both methods are equivalent.}
        \label{fig:sokoban_budget_pitfall}
    \end{minipage}
\end{figure}

This issue arises in practical settings as well. Figure \ref{fig:sokoban_budget_pitfall} compares subgoal methods and low-level BestFS on the Sokoban environment. The dashed line represents the same runs but evaluated with the sparse search budget instead of the complete search budget. For BestFS, both budget measures are equivalent. The figure clearly demonstrates that while kSubS and \BestFS{} visit a similar number of states to solve an instance, the sparse search budget falsely amplifies the difference between the two methods.

\subsection{Baselines}

A common evaluation practice in hierarchical search studies is to compare hierarchical methods against the search algorithm used as the planner \citep{ksubs, ada, hips, hipseps}. While this is generally a good approach, it is critical to ensure that baseline methods are properly tuned to allow for fair comparisons.

Our study shows that the most effective low-level method is \BestFS{} with a confidence threshold. This simple greedy search often performs significantly better than other low-level methods and, in some cases, is competitive with subgoal methods. However, if we were to follow prior works such as \citep{ksubs, ada} and restrict our comparisons to variants of BestFS that select a fixed number of actions in each node expansion, without employing a confidence threshold (see Appendix \ref{appendix:algorithms_bestfs} for detailed definitions and analysis), we would artificially widen the gap between BestFS and subgoal methods. As noted in Appendix \ref{appendix:algorithms_bestfs}, the performance of \BestFS{} is highly sensitive to the confidence threshold, and proper tuning is essential. Nevertheless, we advocate for using \BestFS{} with a confidence threshold as a standard baseline in evaluations of hierarchical methods.

\subsection{Code Quality}

\begin{figure}[ht]
    \centering
    \begin{minipage}[b]{0.40\textwidth}
        \centering
        \includegraphics[width=\textwidth]{images/random_rubik.png}
    \end{minipage}
    \hfill
    \begin{minipage}[b]{0.48\textwidth}
        \centering
        \includegraphics[width=\textwidth]{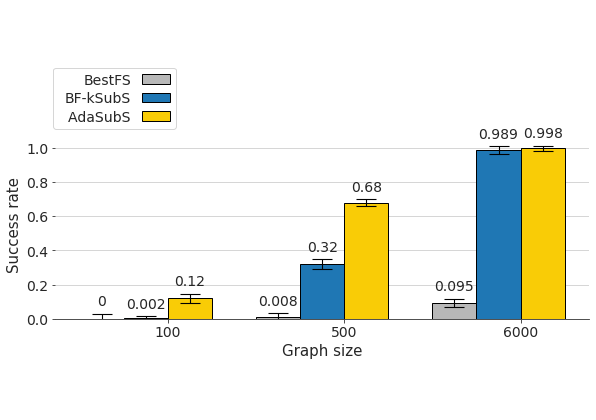}
    \end{minipage}
    \caption{Solving the Rubik's Cube. Components are trained on reverse random shuffles. The left chart present our results, while the right presents results of the same experiment from \citep{ada}.}
    \label{fig:rubik_compare_pitfall}
\end{figure}

While our results generally align with the findings of \citep{ksubs, ada}, we observed some notable differences. Most strikingly, when components were trained on reverse random shuffles of the Rubik's Cube, our models demonstrated significantly better performance. In particular, \citep{ada} reports that both kSubS and AdaSubS substantially outperform \BestFS{}. However, in our experiments, these methods perform similarly, with only minor differences between them (see Figure \ref{fig:rubik_compare_pitfall}).

For this study, we re-implemented all algorithms from scratch, using up-to-date libraries and carefully tuning hyperparameters. Our experiments revealed that low-level methods are highly sensitive to the quality of the value function, whereas subgoal-based methods are more resilient (Section \ref{sec:analysis_noise}). We hypothesize that the discrepancy in performance compared to \citep{ksubs, ada} may stem from insufficient training of the value function in their implementation, leading to the observed performance gap.

Using the original implementations of kSubS and AdaSubS, which is a common practice, would replicate the same limitation. This shows the importance of re-implementing algorithms independently and carefully tuning their components, ensuring that evaluations are not biased by potential shortcomings in the original implementations.

\newpage
\section{Proof Of The Search Advancement formula}\label{appendix:noise_theorem_proof}

\begin{theorem}[Search advancement formula, complete statement]
Let $g_k: S\to \mathcal P(S)$ be a stochastic \mbox{$k$-subgoal} generator that, given a state $s\in S$ samples a set of $b$ subgoals $\{s_i\}$ such that the distances $d(s_i, s)$ are independent, uniformly distributed in the interval $[-k;k]$. Let \mbox{$V:S\to\mathbb R$} be a value function with approximation error uniformly distributed in the interval $[-\sigma;\sigma]$.

Then, after $n$ iterations of search, the expected total progress toward the goal is:
\begin{equation}\label{eq:noise_main_appendix}
    \mathbb E_{Adv} = \frac{nb}{4\sigma k} \int_{-k}^k x\left( \int_{-\sigma}^{\sigma} \tilde u(x+h)^{b-1}\mathrm dh \right) \mathrm dx,
\end{equation}
where $\tilde u(x)$ is CDF of the sum of two uniform variables $U(-k,k)+U(-\sigma,\sigma)$.
Additionally, if we approximate that sum as $U(-k-\sigma,k+\sigma)$, we get
\begin{equation}\label{eq:noise_approximated_appendix}
    \mathbb E_{Adv} \approx \frac{n\left((k+\sigma)^b(bk^2+bk\sigma-2k\sigma-2\sigma^2) + \sigma^{b}(2k\sigma+bk\sigma+2\sigma^2)-k^b(bk^2)\right)}{(b+1) (b+2) k \sigma (k+\sigma)^{b-1}}
\end{equation}
\end{theorem}

\begin{proof}
Let $A_1, \dots, A_b$ be independent and identically distributed (i.i.d.) random variables sampled from $U(-k, k)$, and let $B_1, \dots, B_b $ be i.i.d. random variables sampled from $U(-\sigma, \sigma) $. Denote the CDF of the sum $A_i + B_i $ as $\tilde{u}(x) $, and its corresponding probability density function (PDF) as $p(x) = \tilde{u}'(x) $. Let $I = \arg\max_{i} (A_i + B_i) $.

We now define the cumulative likelihood of selecting the largest sum among the subgoals:
$$
CLS(x) = \mathbb{P}\left( \forall_{1 \leq i \leq b} \, A_i + B_i < x \right).
$$
Since the $A_i $'s and $B_i $'s are independent, it follows that $CLS(x) = \tilde{u}(x)^b $, which represents the cumulative distribution of the largest sum $A_i + B_i $. Differentiating this expression gives the PDF of the largest sum:
$$
PLS(x) = CLS'(x) = b \cdot \tilde{u}(x)^{b-1} \cdot p(x).
$$

Now, consider the event that $A_I = x $, which is equivalent to the event that the maximum $\max_i (A_i + B_i) = x + h $ for some $h \in [-\sigma, \sigma] $ and $B_I = h $. Given that $\max_i (A_i + B_i) = x + h $, there are $p(x+h) \cdot 4\sigma k $ possible values of $B_I $, since $A_I \in [-k, k] $ and $B_I \in [-\sigma, \sigma] $. Therefore, the PDF of this variable is
$$
q(x) = \int_{-\sigma}^{\sigma} \frac{PLS(x+h)}{p(x+h) \cdot 4\sigma k} \, \mathrm{d}h = \int_{-\sigma}^{\sigma} \frac{b \cdot \tilde{u}(x+h)^{b-1}}{4\sigma k} \, \mathrm{d}h.
$$

Thus, the expected value of $A_I $, which represents the progress in each step, is given by
$$
\mathbb{E}[A_I] = \int_{-k}^{k} x q(x) \, \mathrm{d}x = \frac{b}{4\sigma k} \int_{-k}^{k} x \left( \int_{-\sigma}^{\sigma} \tilde{u}(x+h)^{b-1} \, \mathrm{d}h \right) \mathrm{d}x.
$$

If we model the search process as advancing to the best subgoal in each iteration, the total expected progress after $n $ iterations is
$$
\mathbb{E}_{Adv} = n \mathbb{E}[A_I] = \frac{nb}{4\sigma k} \int_{-k}^{k} x \left( \int_{-\sigma}^{\sigma} \tilde{u}(x+h)^{b-1} \, \mathrm{d}h \right) \mathrm{d}x.
$$

Finally, by approximating the PDF $p(x) \approx \frac{1}{2k + 2\sigma} \mathds 1_{[-k-\sigma, k+\sigma]} $, and substituting this approximation into the previous expression, we arrive at the closed-form approximation:
$$
\mathbb{E}_{Adv} \approx \frac{n\left( (k+\sigma)^b (bk^2 + bk\sigma - 2k\sigma - 2\sigma^2) + \sigma^{b} (2k\sigma + bk\sigma + 2\sigma^2) - k^b (bk^2) \right)}{(b+1)(b+2) k \sigma (k+\sigma)^{b-1}}.
$$
\end{proof}

\newpage
\section{Proof Of The Densification Of The Action Space Theorem}\label{appendix:actions_theorem_proof}

In Section \ref{sec:analysis_complex_actions}, we showed experimentally that both in the mathematical INT environment and Rubik’s Cube with multiplied action space the advantage of subgoal methods is significant. We attributed those benefits to the ability of subgoal methods to use states as actions and the reduced diversity in low-level search. And indeed, we can prove in general that as the action space gets more complex, the diversity of top actions drops.

To give an illustrative example, in the Rubik’s Cube experiment, to model the increasingly complex action space, for an arbitrary state we can view the training data as a ground-truth density function $f$ over an interval $[0, 1]$, that is split evenly between the actions (i.e. into 12 intervals of length 1/12). Then, we can define arbitrarily dense action spaces $A_n$ consisting of $n$ points distributed evenly in the domain. For instance, $A_{12}$ corresponds to the standard Rubik’s Cube action space, while $A_{1200}$ corresponds to the variant multiplied 100 times. Our theorem confirms that the actions selected by the policy gets less diverse as the complexity of the action space increases, up to the extreme of converging to a single point as $n$ approaches infinity. In practice, it is even more general, since the data-driven action distribution $f$ may also model smooth interpolation between actions.

While this is rather intuitive when the learned distributions are perfect, it may seem that approximation errors, induced both by the limited training data and the policy network can actually improve diversity. We show that the result holds even in presence of arbitrarily large approximation errors, which is a bit counter-intuitive.

Formally, the theorem is as follows:

\begin{theorem}[Densification of the action space]\label{thm:complex_actions_appendix}
Fix any state $s$ from the state space $S$. Let $f : A \to [0, 1]$ be the action distribution induced by the data-collecting policy for the state $s$. Assume that $f$ is continuous and has a unique maximum. For clarity, assume $A=[0,1]$.

Consider a sequence of increasingly dense discrete action spaces $A_n := \{i / n\}_{i=0}^n\subset A$.
Let $\rho_n : S \times A_n \to [0, 1]$ be a family of policies that learn the distribution $f|_{A_n}$ over actions, with uniform approximation error $U(-E, E)$, where $E\in\mathbb R_+$.
Let $r_n$ be the range of the top $K$ actions according to the probabilities estimated by $\rho_n$. Then

$$\lim_{n \to \infty} \mathbb{E}[r_n] = 0.$$
\end{theorem}

Intuitively, this theorem states that as the action space become more dense and complex, the actions sampled for search become increasingly less diverse, which strongly impedes successful planning. Note that this analysis is strictly more general than the experiment in Section \ref{sec:analysis_complex_actions} with the Rubik's Cube environment, where we simply copied the available actions. Here we model the complexity by adding dense intermediate actions, which leads to a similar conclusion.

While we assume a one-dimensional action domain for clarity, it is straightforward to generalize the proof to cover arbitrarily high-dimensional action spaces.

Firstly, we shall prove the following key lemma.

\begin{lemma}\label{lemma:complex_actions}  
Let \( f : [0, 1] \to \mathbb{R} \) be a continuous function with a unique maximum. Let \(\{a_n\}\) be a partition of the interval \([0, 1]\) into \(n\) uniformly spaced points, i.e., \(a_{n, i} = \frac{i}{n}\) for \(i = 0, 1, \ldots, n\). Define \(e_{n,i}\) as i.i.d. samples from a uniform distribution \(U(-E, E)\). For a fixed \(n\), let \(r_n \in \mathbb{R}\) denote the smallest interval length such that the points in \(\{a_n\}\) corresponding to the top \(K\) values of \(f(a_{n,i}) + e_{n,i}\) are contained within this interval. Then
\[
\lim_{n \to \infty} \mathbb{E}[r_n] = 0.
\]
\end{lemma}

\begin{proof}
Define \(p_{n,i,k}\) as the probability that \(f(a_{n,i}) + e_{n,i}\) is the \(k\)-th highest value among all points in \(\{a_n\}\). Let $m$ be the unique point such that $f(m)$ is maximal. Without loss of generality, we may assume that $m=0$.

Let \(d_{n,k}\) denote the expected distance of the \(k\)-th highest point from \(0\), expressed as
\[
d_{n,k} := \sum_{i=0}^n p_{n,i,k}a_{n,i}.
\]
For sufficiently large \(n\), it holds that \(r_n \leq d_{n,1} + \ldots + d_{n,K} \leq K d_{n,K}\). Thus, it suffices to prove that \(\lim_{n \to \infty} d_{n,K} = 0\).
 
Fix \(\alpha \in (0, 1)\) such that $f(a_{n,\alpha n}) \geq f(a_{n,\alpha' n})$ for each $\alpha'>\alpha$. Since $f$ is continuous and $m=0$ is the unique maximum of $f$, there exist such $\alpha$ arbitrarily close to $0$. Let \(q_{n,\alpha}\) be the probability that \(f(a_{n,\alpha n}) + e_{n,\alpha n}\) is among the top \(K\) values. Since $m$ is a unique maximum, there exists $0<\beta<\alpha$ such that \(f(a_{n,\beta n}) > f(a_{n,\alpha n})\). Therefore, if at least \(K\) points \(a_{n,i}\) with \(i / n < \beta\) satisfy \(e_{n,i} > E - (f(a_{n,\beta n}) - f(a_{n,\alpha n}))\), then \(f(a_{n,\alpha n}) + e_{n,\alpha n}\) cannot be among the top \(K\). The probability of this event is a strict upper bound on \(q_{n,\alpha}\).

The events \(e_{n,i} > E - (f(a_{n,\beta n}) - f(a_{n,\alpha n}))\) are pairwise independent, each occurring with probability
\[
c := \frac{f(a_{n,\beta n}) - f(a_{n,\alpha n})}{2E} > 0.
\]
For sufficiently large \(n\), the probability that at most \(K\) of the \(\beta n\) trials succeed is bounded by
\[
1 - K \binom{\beta n}{K} (1 - c)^{\beta n}.
\]
Using the asymptotic behavior of binomial coefficients and exponential terms, it follows that
\begin{equation}\label{eq:q_limit}
\lim_{n \to \infty} n^2 q_{n,\alpha} = 0,
\end{equation}
with convergence that is exponential.

Using the definition of \(d_{n,K}\), decompose it as
\[
d_{n,K} = \sum_{i=0}^n p_{n,i,K} a_{n,i} = \sum_{i=0}^{\alpha n} p_{n,i,K} a_{n,i} + \sum_{i=\alpha n}^n p_{n,i,K} a_{n,i}.
\]

For \(i \geq \alpha n\), since we know that $f(a_{n,\alpha n}) \geq f(a_{n,\alpha' n})$ for each $\alpha'>\alpha$, we can bound \(p_{n,i,K}\) by \(p_{n,\alpha n,K}\) for sufficiently large \(n\). Therefore
\[
\sum_{i=\alpha n}^n p_{n,i,K} a_{n,i} \leq (1 - \alpha)n p_{n,\alpha n, K}.
\]
Since \(p_{n,\alpha n, K} \leq q_{n,\alpha}\), it follows that
\[
(1 - \alpha) n^2 p_{n,\alpha n,K} \leq (1 - \alpha) n^2 q_{n,\alpha}.
\]
According to Equation \ref{eq:q_limit}, this term converges to \(0\).

For \(i \leq \alpha n\), observe that \(a_{n,i} < \alpha\) and the probabilities \(p_{n,i,K}\) sum to at most \(1\). Thus
\[
\sum_{i=0}^{\alpha n} p_{n,i,K} a_{n,i} \leq \alpha.
\]

Combining these bounds, we have
\[
\lim_{n \to \infty} d_{n,K} \leq \alpha.
\]
Since \(\alpha > 0\) was an arbitrarily small constant, it follows that \(\lim_{n \to \infty} d_{n,K} = 0\).

By the relation \(r_n \leq K d_{n,K}\) and the fact that \(\lim_{n \to \infty} d_{n,K} = 0\), we conclude that
\[
\lim_{n \to \infty} \mathbb{E}[r_n] = 0.
\]
\end{proof}

Now, Theorem \ref{thm:complex_actions_appendix} is a straightforward implication of Lemma \ref{lemma:complex_actions}, applied to the sequence of policies $\rho_n$ and increasingly dense action spaces $A_n$.

%% file: resources/example_Beginner.txt
bwooyryoorgrgbgyyygbyrrwryggygbgwwbbwrwooywgobbrowrbwo
yoboyworogbygbgyyygygrrwrygwrwbgwwbbrgrooywgobbrowrbwo
ooyryoowbgyggbgyyywrwrrwrygrgrbgwwbbgbyooywgobbrowrbwo
ooyryoygggybgbbyyrrrwyrrgwwogrwgwbbbgbyooywgowbrowrbwo
ooyryorbbgywgbbyyrgyrwrrwrwygrggwgbbgbyooywgobwoowrbwo
rwbryorbbyywobboyrgyrwrrwrwygoggwgbbwoggoboyybwoowrggy
owbryorbbbywwbbryrgyrwrrwrwygyggggbgogwyooybgbwoowryoo
rrobywbobgyrwbbryrygywrrwrwogwggggbgbywyooybgbwoowryoo
rrybyrbowgyrwbbryrygowrrwroggobggggwbywwooobgbwyowyyob
rrobyrboogyrwbbryrygywrywrbgbggggwgowywrooybgbwoowwyob
rwgbyrbooyyrobbbyrygywrywrbgbrggrwgowogyobwrybwoowwogg
bbroyworgygyobbbyrgbrwrywrbwogggrwgoyyryobwrybwoowwogg
grooyworgrgybbbbyrgbrwrywrbwoggggwgowyyroyybrbwoowwyob
oogryrgwogbrbbbbyrwogwrywrbwyygggwgorgyroyybrbwoowwyob
growyoorgwogbbbbyrwyywrywrbrgygggwgogbrroyybrbwoowwyob
owgryrgoowyybbbbyrrgywrywrbgbrgggwgowogroyybrbwoowwyob
rgoryrgoogyywbboyrrgywrywrbgbbggowgyyrwboorygbwoowwwbb
boyryrgoooyygbbryrrgywrywrbgbbggbwgwrbyyorgowbwoowwgwo
roywyrwooybrybyogrbgyorygrbgbbggbwgwrbgyorgobwworwwywo
wwroyoorybgyybyogrgbborygrbrbgggbwgwybryorgobwworwwywo
bwrryorryoybgbgryywbboryorbrbgggbwgwybyyorgowgwoowwgwo
rrbrywyorwbbgbgryyrbgoryorbybyggbwgwoybyorgowgwoowwgwo
yrroyrrwbrbggbgryyybyoryorboybggbwgwwbbyorgowgwoowwgwo
roywyrbrrybygbgryyoyboryorbwbbggbwgwrbgyorgowgwoowwgwo
woyryrgrrrgyybbygyrybwrybrbwbbggbwgwrbgyoogogowoowwowo
goyoyrgrryyrgbgybywybrrygrbwbbggbwgwrboyoogoorwowwwbwo
yrroyrgogrbogbgybyyyrrrygrbwybggbwgwwbbyoogoorwowwwbwo
yrroyygobrbogbgybyyyorrwgrowgwggywbbgbbrooroorwgwwybww
goyoyrbyryyogbgybywgwrrwgrogbbggywbbrborooroorwgwwybww
goroyrbyryyogbgybywgyrrrgrrbybbgbggwwboyoogoorwwwwwbwo
bbwoyrbyrryoobggbywgyrrrgrrbyobgwggbgywoobooorwwwwwygy
wrrbyybobgywobggbyryorrrgrrwgybgwggbbyooobooorwwwwwygy
gogbyybobyywgbgybyryorrrgrrwgwbgrggroboyooboorwwwwwbwy
gobbyybooyywgbgybyrygrrygrbwrrgggwbgybowoowoorwowwrbwr
byooyogbbybogbgybyyywrrygrbryggggwbgwrrwoowoorwowwrbwr
bywoyygbbybogbgybyyyorrrgrrwgrbgygggbrroooooorwwwwwbww
gobbyybywyyogbgybywgrrrrgrrbrrbgygggybooooooorwwwwwbww
bbgyyowybwgrgbgybybrrrrrgrrybobgygggyyooooooorwwwwwbww
obgoyooybygwbbgygrbrryrrwrrybobgygggyyboowoorbwwrwwgww
oooyybbogbrrbbgygryboyrrwrryybbgygggygwoowoorbwwrwwgww
yooyybwogrgrrbgbbybborrrgrryybbgygggygbooyooorwwwwwwww
yooyybygrrgrrbwbbwgrbrrbrrowybogygggygbooyooogbywwwwww
yyygyorbogrbrbwbbwwybrrbrroygbogygggrgrooyooogbywwwwww
rgybyyooywybrbwbbwygbrrbrrorgrogyggggrbooyooogbywwwwww
rgybyyrogwyyrbobbobbogrryrrygrbgyggggrbooyooobwwwwwwww
yyggyorbrgrbrbobbowyygrryrrbbobgygggygrooyooobwwwwwwww
yyggyooobgrbrbwbbwygwrryrryrbobgyrggygrooyooogbbwwwwww
ogyoyybogygwrbwbbwrborryrryygrbgyrgggrbooyooogbbwwwwww
ogyoyyybryggrbobbboyybrrrrrbgrbgyggggrbooyooowwwwwwwww
yoobygryyoyyrbobbbbgrbrrrrrgrbbgygggyggooyooowwwwwwwww
yorbyrryroyyrbobbbbgwbrwrrwgbgggrgybygggoyooowwowwowwy
rrroyyybryggrbobbboyybrwrrwbgwggrgybgbggoyooowwowwowwy
rrooygybgyggrbobbboyrbryrrrwrbggybggybgooyooowwywwwwww
oggrybroyybgrbobbbyggbryrrroyrggybggwrbooyooowwywwwwww
oggrybogbybyrbobbrgyrgrrybryyrwgywggwrbooyooogobwwwwww
orogygbbggyrrbobbryyrgrrybrwrbwgywggybyooyooogobwwwwww
orogygrorgygrbobbbygybryrrrbrbbgygggybyooyooowwwwwwwww
rgooyrrgoygyrbobbbbrbbryrrrybybgyggggygooyooowwwwwwwww
rorgygorobrbrbobbbybybryrrrgygbgygggygyooyooowwwwwwwww
roygyyorrbrbrbobbbybwbrwrrwgbgggygygogygoyroowwowwowwy
yyroyrrgoogyrbobbbbrbbrwrrwybwggygyggbggoyroowwowwowwy
yyroygrggogyrbobbbbrrbrrrrowygbgyyggybgooyooowwbwwwwww
rggyygyorybgrbobbbogybrrrrobrrbgyyggwygooyooowwbwwwwww
rggyygbbyybrrbobbyyrogrrobrbrrwgywggwygooyooogobwwwwww
byrbygyggyrorbobbybrrgrrobrwygwgywggybrooyooogobwwwwww
byrbygyooyrgrbobbbogbbrrrrryygggygggybrooyooowwwwwwwww
rgoyyobbyybrrbobbbyrgbrrrrrogbggygggyygooyooowwwwwwwww
ooygybrybyygrbobbbybrbrrrrryrgggygggogbooyooowwwwwwwww
yoybybrybgobybbyrbwbrwrrwrryrgggygggogroogoooowwywwbww
rbyyyobbywbrybbyrbyrgwrrwrrogrggyggggoboogoooowwywwbww
obygyobbyyywrbbbbrrrgyrrbrrogrggyggggobooyoooywwwwwwww
bgobybyoyrrgrbbbbrogryrrbrrgobggygggyywooyoooywwwwwwww
bgobybrbgrryrbwbbwbyorrgrrryobogyyggyywooyooogggwwwwww
obggybbbryywrbwbbwrryrrgrrrbyoogyyggyobooyooogggwwwwww
obggybboyyyrrbbbbbygrrrrrrrgyoggygggyobooyooowwwwwwwww
bgooybybgygrrbbbbbgyorrrrrryobggygggyyrooyooowwwwwwwww
ogoyybrbgbrybbgbbrbyoorryrryobggygggyywoowoowgwwrwwrww
obggyboyryywbbgbbrbryorryrrbyoggygggyoboowoowgwwrwwrww
bbgoybyyrwgrybbybbgryrrrrrrbyoggygggyoooogooowwwwwwbww
gbrbyyboyyooybbybbwgrrrrrrrgryggygggbyooogooowwwwwwbww
yyybyyboyboowbbwbbwgrrrrrrrgrgggbggrogoyooboowwwwwwgyy
bbyoyyyyywgrwbbwbbgrgrrrrrrogoggbggrbooyooboowwwwwwgyy
obroyyyyyygrbbbbbbgrgrrrrrrogyggygggbyboooooowwwwwwwww
yooyybyyrgrgbbbbbbogyrrrrrrbybggygggygroooooowwwwwwwww
yyyyyorboogybbbbbbbybrrrrrrygrggyggggrgoooooowwwwwwwww
ryybyyooybybbbbbbbygrrrrrrrgrgggygggogyoooooowwwwwwwww
gygbyyooyyybybbrbbygrrrrrrrgrwggwggwooooogooywwwwwwbbb
gyyyyogbooooybbrbbyybrrrrrrygrggwggwgrwoogooywwwwwwbbb
ryoyyogboboobbbbbbyybrrrrrryggggyggywgyroogoowwwwwwwwr
oooyybrygwgybbbbbbboorrrrrryybggyggyyggroogoowwwwwwwwr
oogyyrryywgybbbbbbboorrbrrgbyyyggyggrggwoowoowwowwrwwr
ryoyyoyrgboobbbbbbbyyrrbrrgrggyggyggwgywoowoowwowwrwwr
ryyyybyrgboobbbbbbbyorrrrrryyrggggggggyoooooowwwwwwwww
yyrryygbybyobbbbbbyyrrrrrrrggyggggggboooooooowwwwwwwww
grybyyyyryyrbbbbbbggyrrrrrrbooggggggbyooooooowwwwwwwww
oggbyyyyryyrrbbgbbggyrrrrrrbowggwggwoobooyooowwwwwwybb
oggyyybyrgrybbybbrogybrryrrbowggwggwooyoowoowgwwrwwrbb
byoyygrygogybbybbrbowbrryrrooyggwggwgryoowoowgwwrwwrbb
byobygyygyyrgbbobbgowrrrrrrooyggwggwgrrooyoobwwwwwwybb
oggyyybbygrrgbbobbyyrrrrrrrgowggwggwooyooyoobwwwwwwybb
oggyyybbyyrrbbbbbbyyrrrrrrrgooggggggyyboooooowwwwwwwww
oggyyybbygrrgbbobbyyrrrrrrrgowggwggwooyooyoobwwwwwwybb
bggyyyybyoggbbrbbroyryrrbrrgowggwggwooyoowoowywwrwwrbb
yybbygyygoyrbbrbbrgowyrrbrrooyggwggwoggoowoowywwrwwrbb
gybyygbygrrrybbobbyowrrrrrrooyggwggwogyoobooywwwwwwgbb
bggyyygybogyybbobbrrrrrrrrryowggwggwooyoobooywwwwwwgbb
oyoyyygybggybbbbbbrrrrrrrrryobggggggybyoooooowwwwwwwww
oyooyyyybbbgbbgbbyorryrrgrryobggggggybwoowoowrwwrwwrww
yooyyybyoorrbbgbbyyobyrrgrrybwggggggbbgoowoowrwwrwwrww
yooyyygyorgyrbbobbrobrrrrrrybwggggggbbbooyooywwwwwwgww
gyyyyooyorobrbbobbybwrrrrrrbbbggggggrgyooyooywwwwwwgww
yyyyyoyyoorrbbobbbgbwyrrorrbbbggggggrggoowoowywwrwwrww
yyyyyyooygbwbbobbbbbbyrrorrrggggggggorroowoowywwrwwrww
oyyoyyyyybbbbbobbbrggyrrorrorrgggggggbwoowoowywwrwwrww
ryyyyyoyybobbbbbbbyggrrrrrrorrgggggggbyoooooowwwwwwwww
rggyyyoyyyobybbrbbyggrrrrrrorwggwggwoogoobooywwwwwwbbb
oyryygyygyggybbrbborwrrrrrroogggwggwyoboobooywwwwwwbbb
ryyyygyygbggbbbbbborwrrrrrroooggyggrbbyoooyoowwwwwwwwg
yyryyyggyorwbbbbbbooorrrrrrbbyggyggrbggoooyoowwwwwwwwg
yyryyyggyrrwybbybbooorrrrrrbbgggwggwyoboogoogwwwwwwobb
gyygyyyyroooybbybbbbgrrrrrryobggwggwrrwoogoogwwwwwwobb
yggyyyryybbgybbybbyobrrrrrrrrwggwggwooooogoogwwwwwwobb
yybyyyryyobgbbbbbbyobrrrrrrrryggggggoggoooooowwwwwwwww
yybyyrryrobgbbbbbbyowrrwrrwggrggrggyyggyooboowwowwowwo
ryyyyyrrbyowbbbbbbggrrrwrrwyggggrggyobgyooboowwowwowwo
rybyyyrroyowbbbbbbggyrryrrbgrygggyggobgoooooowwrwwwwww
ryrryyoybggybbbbbbgryrryrrbobggggyggyowoooooowwrwwwwww
ryyryyoybggybbbbbbgrrrrwrrwygoggbgggbowyooroowwowwowwy
orryyybyygrrbbbbbbygorrwrrwbowggbgggggyyooroowwowwowwy
byoyyryyrygobbbbbbbowrrwrrwggyggbggggrryooroowwowwowwy
byryyyyygygobbbbbbboorrrrrrybgggggggyrroooooowwwwwwwww
yybyyygyrboobbbbbbybgrrrrrryrrggggggygooooooowwwwwwwww
yybyyybbobowbbwbbwrryrrbrrggrryggrggygoooooooggywwwwww
byoyybyybygobbwbbwbowrrbrrgrryyggrgggrrooooooggywwwwww
bbyyybyybwgowbwwbwbowrrbrrgrrbygyrgorooroogooggywwwggy
yybyybbbybowwbwwbwrrbrrbrrgrooygyrgowgoroogooggywwwggy
yybyybryrboywbbwbbbbgrrrrrryooggyggowgoroogoowwwwwwggy
bbryyyyyrwgowbbwbbboyrrrrrrbbgggyggoyooroogoowwwwwwggy
gyoyyyyyrrgobbbbbbboyrrrrrrbbygggggggryoooooowwwwwwwww
yygyyyryoboybbbbbbbbyrrrrrrgryggggggrgooooooowwwwwwwww
ryyyyyoygbbybbbbbbgryrrrrrrrgoggggggboyoooooowwwwwwwww
oyryyygyygrybbbbbbrgorrrrrrboyggggggbbyoooooowwwwwwwww
gyoyyyyyrrgobbbbbbboyrrrrrrbbygggggggryoooooowwwwwwwww
gyoyyybborgwbbwbbwrrbrrorryybyyggrgggryooooooggbwwwwww
oyoyybgybgrybbwbbwrgwrrorryrrbyggrggybyooooooggbwwwwww
bbgyybgybwrywbwwbwrgwrrorryrroygyrgoyoobooyooggbwwwggb
gybyybbbgrgwwbwwbwrrorrorryyooygyrgowrybooyooggbwwwggb
gybyybyyrrggwbbwbbooyrrrrrrbooggyggowrybooyoowwwwwwggb
bbryyygyywrywbbwbbrggrrrrrrooyggyggoboobooyoowwwwwwggb
yyoyyygyyrrybbbbbbrggrrrrrroobggggggybboooooowwwwwwwww
yyoyyyoggrrybbybbggrrgrrrrrwobwggwggybbooooooybbwwwwww
yyoyyyoggrrybbyrrrgrrgrrwggwobwggoooybbooobbgbwwbwwyww
yyoyyyryyrrbbbwrrwwgggrrgrroobggggooybbooobbgowwbwwyww
yyoyyyryyrrbbbwbbgwgggrrrrwoobggggrrybbooogooybowwwwww
yyoyyyoggrrybbybbrgrwgrrwgroobbggyrrybbooogoobwgwwwwww
yyoyyyoggrrybbywgrgrwgrryrroobbgggooybbooobbrgwwwwwbww
yyoyyyryyrrgbbwwgwyggrrrrrwoobggggooybbooobbrgbowwwbww
yyoyyyryyrrgbbwbbryggrrrwgwoobgggrrwybbooogoobwgwwbwwo
yyoyyyogrrrybbybbrgrwgrgyrwgobwggbrwybbooogoogwrwwbwwo
yyoyyyogrrrybbyyrwgrwgrgbrwgobwgggooybbooobbrrbowwwgww
yyoyyywyyrrrbbbyrobggrrrwgwoobgggrooybbooobbrgwgwwwgww
yyoyyywyyrrrbbbbbrbggrrryrooobgggwgwybboooroogwgwwwwwg
yyoyyyogwrrybbybbwgrogrrbrygobwggggwybboooroorbrwwwwwg
yyoyyyogwrrybbybrygrogrrggwgobwggrooybbooobbwrwgbwwrww
yyoyyyyyyrrrbbwbrgggggrrwrooobgggwooybbooobbwrwgbwwrww
yyoyyyyyyrrrbbwbbwggggrrbrgoobgggwroybbooowoorbrwwwwwg
yyyyyyyyogggbbwbbwoobgrrbrgybbgggwrorrrooowoorbrwwwwwg
yyyyyyoyyoobbbwbbwybbgrrbrgrrrgggwrogggooowoorbrwwwwwg
yyyyyyrgwooybbybbobrgbrrygbrrrbggrrogggooowoobwwwwwwwg
yyyyyyrgwooybbyygbbrgbrrrrorrrbggwoogggooobbowwgwwwbww
yyyyyybyyoowbbwyggrbbrrrorgrrrgggwoogggooobbowbrwwwbww
yyyyyybyyoowbbwbborbbrrryggrrrgggorggggooowoobwwwwbwwr
yyyyyyrgoooybbybbbbrgbrgrrywrrwggbrggggooowoowwowwbwwr
yyyyyyrgoooybbyrrybrgbrgbrgwrrwggwoogggooobbbobrwwwwww
yyyyyyyyyooobbbrrrbbbrrrgggrrrgggooogggooobbbwwwwwwwww
yyyyyyyyyooobbbbbbbbbrrrrrrrrrgggggggggoooooowwwwwwwww
yyyyyyyyybbbbbbbbbrrrrrrrrrgggggggggooooooooowwwwwwwww
yyyyyyyyyrrrbbbbbbgggrrrrrroooggggggbbboooooowwwwwwwww
yyyyyyyyygggbbbbbbooorrrrrrbbbggggggrrroooooowwwwwwwww
yyyyyyyyyooobbbbbbbbbrrrrrrrrrgggggggggoooooowwwwwwwww
yyyyyyyyybbbbbbbbbrrrrrrrrrgggggggggooooooooowwwwwwwww

%% file: resources/example_CFOP.txt
bryoyrorryyygboywogybbrgbgrwwoogwgygbbrooyobrwrwgwbgww
brooyoorbyyygboywogyybrrbgrowgwgywogwbrboywbrwrbgwggwr
brooyoowwyybgbrywoyrryrggbbbwgrgywogwbrboywbryoogwggwr
oobwyrwooyrrgbrywobwgyrggbbwbrrgywogyybboywbryoogwggwr
oobwyrorryrygboywogybbrwbggwbrogyoogyybboywbrwrwgwggwr
oobwyworgyrygboywogywbrgbgroowogbgyrrybroybbrwrwgwbgwy
oworyogwbgywgboywooowbrgbgrrybogbgyryryroybbrwrwgwbgwy
growywboooowgboyworybbrgbgryryogbgyrgywroybbrwrwgwbgwy
bwgoyroworybgboywoyrybrgbgrgywogbgyroowroybbrwrwgwbgwy
bwgoyroobrywgbrywwbbygrrrgyoywwgboyroowroybbrgoggwbgwy
bwgoyrwrwryggboywgrgbgrbyryoywogbbyroowroybbrowogwbgwy
rwgyyrwrwygrwbygogbgborbwryoywogbbyroogrogbborwogwbywy
gwyyyrwrwygrwbyyogbgborbwryoyrogwbygggoooborbrwogwbrbw
wygrywwrybgbwbyyogoyrorbwryggoogwbygygrooborbrwogwbrbw
owgrywwryggbybywogoyrorbwryggwogbbyrooyrogbbrrwogwbbwy
owrrybwryggbybywogoyoorbwrybogyggrbwyoywoggbrrwbgwrbwo
oworybwryggbybywogoyborrwrorybbgowggyoybogrbrrwggwwbwy
bogrybwryogbwbyoogoyborrwroryybgwwgbrbyboorgyrwggwwgyw
ywbrybwryggbobybogoyborrwrorywbgywggrbrgobyoyrwggwwowo
wryrywybboybobybogryworrwrorbrbgywggggbgobyoyrwggwwowo
yrwbyrbwyrywobybogrbrorrwroggbbgywggoybgobyoyrwggwwowo
bbywyryrwrbrobybogggborrwrooybbgywggrywgobyoyrwggwwowo
ybybyrwrwborobbgyrbgbwrryrooybbgywggryogogyorgwgowwwwo
rbygyrorwgobyborbrygbbrrwrooybbgywggrywgooyogbwgwwwywo
ogrrybwryygbyborbroybbrrwrorywbgywgggobgooyogbwgwwwywo
wrorygybroybyborbrrywbrrwrogobbgywggygbgooyogbwgwwwywo
ogrrybwryygbyborbroybbrrwrorywbgywgggobgooyogbwgwwwywo
ogbryrwroygbyborbroygbrwwrowbrggygywyobboorogbwywwgywg
wrorygorboygyborbrwbrbrwwroyobggygywygbboorogbwywwgywg
wrrryboryoygyborbrwbobrgwrbbywogyyggggbgooyogbwrwwwywo
rbyryrwroggbyborbroygbrgwrbwboogyyggbywgooyogbwrwwwywo
rbgrygwrbggbyborbroyrbrwwroyowggbgyooywrooyogbwywwgywb
wrrrybbggoyryborbryowbrwwrooywggbgyoggbrooyogbwywwgywb
brwgyrgbryowyborbroywbrwwroggbggbgyooyrrooyogbwywwgywb
brygyrgboyowyborbroywbrrwrrbboggygggbyrgooyogbwwwwwywo
ggbbyroryoywyborbrbbobrrwrrbyrggygggyowgooyogbwwwwwywo
obgrygyrbbboyborbrbyrbrrwrryowggygggoywgooyogbwwwwwywo
obyrygyrobboyborbrbygbrgwrbwygoggyggoywwoowogbwrwwrywr
yrorybogybygyborbrwygbrgwrboywoggyggbbowoowogbwrwwrywr
yrgrygogbbygyborbrwyrbrrwrryooggyggwyboboooogbwwwwwywb
ggbrygyroyboyborbrbygbrrwrrwyrggyggwyooboooogbwwwwwywb
bgogyrgryyooyborbrybobrrwrrbygggyggwwyrboooogbwwwwwywb
gywgyrgryooogbobbrybobrrwrrbybggwggyobwooygorbwwwwwyyr
gggryyyrwybogbobbrbybbrrwrrobwggwggyoooooygorbwwwwwyyr
bgyryyyrwyboyborbrbybbrrwrrobgggggggoyrooooogbwwwwwyww
yrbrygwyybybyborbrobgbrrwrroyrggggggyboooooogbwwwwwyww
wryyyrygbobgyborbroyrbrrwrryboggggggbybooooogbwwwwwyww
wryyyrrogobbybwrbwwborryrrrybogggbggbybooooogggywwwyww
rywoyrgrywboybwrbwyborryrrrbybgggbggobbooooogggywwwyww
rywoyrbgbwbyybrrbgoyrbrryrryybggggggobbooooogowwwwwyww
borgyybrwoyrybrrbgyybbrryrrobbggggggwbyooooogowwwwwyww
bgbryowyryybybrrbgobbbrryrrwbyggggggoyrooooogowwwwwyww
wrbyygrobobbybrrbgwbybrryrroyrggggggyybooooogowwwwwyww
rywoyrbgbwbyybrrbgoyrbrryrryybggggggobbooooogowwwwwyww
gywoyrbgbrywbbbgryryrorrbrryybggggggobyoowoooowwbwwyww
boggyybrwryrbbbgryyyborrbrrobyggggggrywoowoooowwbwwyww
bgbryowygyybbbbgryobyorrbrrrywggggggryroowoooowwbwwyww
ogboyobygbbyybrybgobybrryrrrywggggggrywooroobowwwwwrww
booyyggobobyybrybgrywbrryrrrywggggggbbyooroobowwwwwrww
boorygyobyyobbbgrybywyrrgrrrywggggggbbroowooorwwbwwyww
ogboyobrybbrbbbgryyyoyrrgrrbywggggggrywoowooorwwbwwyww
ygbyyogryrbybbrbbgryobrryrrbywggggggrybooooooowwwwwwww
gyyrygyobryobbrbbgbywbrryrrrybggggggrbyooooooowwwwwwww
yrgoyybgybywbbrbbgrybbrryrrrbyggggggryoooooooowwwwwwww
gyyrygyobryobbrbbgbywbrryrrrybggggggrbyooooooowwwwwwww
gyyryggroryobbwbbwybbrryrrwyyboggbggrbyooooooggrwwwwww
grgryyogyybbbbwbbwyybrryrrwrbyoggbggryoooooooggrwwwwww
grgryyrobybybbgbbobywyrryrrrbyggggggryooooooobwwwwwwww
rrgoyrbygbywbbgbborbyyrryrrryoggggggybyoooooobwwwwwwww
rrgyyryygwgoybbbbbbbywrrwrrryoggggggybbooooorowwowwyww
grgryyryyybbybbbbbwgowrrwrrbbyggggggryoooooorowwowwyww
rrgoyyoyybyybbbbbbggorrrrrrbbyggggggryyoooooowwwwwwwww
grgryyryyybbybbbbbwgowrrwrrbbyggggggryoooooorowwowwyww
yggryyryygbbrbbgbbwgowrrwrrbbwggwggyoorooyrooowwowwyyb
wwyryyryygbbgbbybbwgowrrwrrbbbggyggyrooooooyrowwowwgrg
wworyrryrgbbgbbybbwgwwrwwrgggbggbyybyooyooyyrowoowogrr
bbbryrryrobbwbbwbbwgwwrwwrgggrggryygyyyyoorooowoowoggy
bbrryyryyobbwbbwbbwgbwrrwrrrrgggyggyyyyooooooowwowwggg
gyyryyryyrbbbbbbbbwgbwrrwrrrrgggggggooyooyooyowwowwoww
yyyyyyyyybbrbbbbbbggbrrrrrrrrgggggggooooooooowwwwwwwww
yyyyyyyyyggbbbbbbbrrgrrrrrroooggggggbbroooooowwwwwwwww
yyyyyyyyyrrgbbbbbbooorrrrrrbbrggggggggboooooowwwwwwwww
yyyyyyyyyooobbbbbbbbrrrrrrrggbggggggrrgoooooowwwwwwwww
yyyyyyyyybbrbbbbbbggbrrrrrrrrgggggggooooooooowwwwwwwww
gyyryyryyrbbbbbbbbwgbwrrwrrrrgggggggooyooyooyowwowwoww
rrgyyyyyywgbbbbbbbrrgwrrwrrooyggggggrbbooyooyowwowwoww
rroyyoyyrwgbbbbbbbrrgwrywryyggoggoggwbbwoywoyowgowrowr
yyryyrroorrgbbbbbbyggwrywrywbboggoggwgbwoywoyowgowrowr
ryyoyyorryggbbbbbbwbbwrywrywgboggoggrrgwoywoyowgowrowr
yyyyyygrrbbybbgbbgrbboryorywgboggoggrrowoowoowwgwwrwwr
yyryyryygrrobbgbbgbbyoryoryrbboggoggwgbwoowoowwgwwrwwr
byroyroygoggrbbrbbwbywrywryrbboggoggwgywoywoyowgowrbwr
oobyyygrrwbyrbbrbbrbbwrywrywgyoggoggoggwoywoyowgowrbwr
gyoryorybrbbrbbrbbwgywrywryoggoggoggwbywoywoyowgowrbwr
yyoyyoyybrrrbbbbbbggyrryrryoggoggoggwbbwoowoowwgwwrwwr
yyyyyyyyyrrrbbbbbbgggrrrrrroooggggggbbboooooowwwwwwwww
yyyyyyyyybbbbbbbbbrrrrrrrrrgggggggggooooooooowwwwwwwww

%% file: resources/example_Kociemba.txt
wwwgyggoybrywbborrrwoorygyybyorgorbbgrrgogwyywbbbwwgoo

wwwgyggogbrywbborrrwworggyyoobygbbrrorrwogbyywbobwygoy

ggwoywggwrwwwbborrooborggyyorrygbbrrbrywogbyywbobwygoy

goggygwwwoobwbborrorrorggyybryygbbrrrwwwogbyywbobwygoy

goggygbyboowwbworwrgyrryoogorybgbwrrrwwwogbyybbrbwygoy

goggygbyboowwbwoogrgyrrywrrorybgbbyyrwwwogorwryybwobbg

woggygwybowoobogwwggygrybrrorybgbbyyrwbwoborrryyrwowbg

rogbygbybgoowbwwoowgygrywrrorybgbbyyrwwwororrgyygwobbg

bbryyobggwgywbwwooorygrywrrrwwbgbbyygoowororrgyygwobbg

bboyywbggwgywbwwooorrgrowrgwbywgyrbbgooooryrrgyygwybbr

wwwyywbggbgybbwrooorrgrowrgwbbwgbrboorroorgoygyygwybyy

wwwyywbggbgybbwgoyorrgroroowbbwgbwrgorroorrbobggywyyyy

wwwyywywybgbbbggogrgoorroorbbbggbgrgorroorrbowwwywyyyy

bbgyywywywgbwbgwogrgoorroorbbyggygryroobororrwwwywybbg

yyyyywywyggbbbgbogrgoorroorbbgggbgrbobrroorrowwwywywww

yyywyyywyrgobbgbogbbgorroorobrggbgrbggbroorrowwwywywww

ywywyyyyybbgbbgbogobrorroorggbggbgrbrgoroorrowwwywywww

ywywyyyyybbgbbgoorobrorrgrbggbggbrrorgoroobogwywwwwwyw

ywrwyryybbbgbbgoorobworwgrwrggrggobbygoyooyogwybwwrwyr

ywwwywyywbbgbbgoorobborrgrrorrbggbggbgoroorogwyywwywyy

gwwoywoywobbobbrggybbwrryrrorrbggbggbgwrowrowoyyowygyy

wwwwywwywroogbbgbbgbborrorrorrbggbggbggroorooyyywwyyyy

wwwyywwwwgbbgbbgbborrorrorrbggbggbggrooroorooyyywwyyyy

wwwyywbbbgbygbygbyooorrrrrrwggwggwggroorooroobbbwwyyyy

wwwyywyyygbbgbbgbbrrorrorrobggbggbggroorooroowwwwwyyyy

wwoyyoyyogbbgbbgbbrrwrryrrybbbggggggyoowoowoowwrwwryyr

wwwyyyyyygbbgbbgbbrrrrrrrrrggbggbggbooooooooowwwwwwyyy

bbbyyyyyywbbwbbwbbrrrrrrrrrggyggyggyooooooooowwwwwwggg

yyyyyyyyybbbbbbbbbrrrrrrrrrgggggggggooooooooowwwwwwwww

%% file: resources/example_random.txt
ybrbyygyybywwboybyoggwrrgoroobwgogrowwobogbgbrgwrwyrrw
ybbbybgywbywwboybyogrwrygoybooogrowgwwoyogwgbrggrwrrrr
ybrbyygyybywwboybyoggwrrgoroobwgogrowwobogbgbrgwrwyrrw
boobyygyyrywbboybyoggwrrgoroowwgrgrrbbwgowbgorgwrwybwy
oyyoyybbgbbwbboybyrywwrrgoroggwgrgrroowgowbgorgwrwybwy
oyyoyyowgbbgbbbybbwrryrorwgwggggrrrroowgowbgowoyrwybwy
yygyywooooowbbbybbbbgyrorwgwrrggrrrrwgggowbgowoyrwybwy
oygwywgooybobbobbwybgyroowgwrrggrrrrwgbgorbgwboyywyrwy
yygyywooooowbbbybbbbgyrorwgwrrggrrrrwgggowbgowoyrwybwy
yygyyooogoowbbbybbbbyyryrwyrgwrgrrrroggwowggowobrwgbww
yygyyooogoowbbbggobbyyryybbrgwrgrrwyoggwowrrrbrwwwowgb
yyyyyyooboowbbbggobbwyroybbrrrwggyrwgggoowgrrbrrwwwwgo
yyyyyyooboowbbbgrrbbwyroggorrrwggybbgggoowyrwwwbgwrowr
yyyyyyrbwoowbbwgrbgybgrbooworroggbbbgggoowyrwywrgwrowr
yyyyyyrbwoowbbwyrwgybgrbgrborroggoowgggoowbbbogywwwrrr
yyyyyyooooowbbbyrrbbbyrrgggyrrgggoowgggoowbbbwwwwwwrrr
yyyyyyooooowbbbbbbbbbyrryrryrrgggggggggoowoowrwwrwwrww
oyyoyyoyybbbbbbbbbyrryrryrrgggggggggoowoowoowrwwrwwrww
yyyyyyyyybbbbbbbbbrrrrrrrrrgggggggggooooooooowwwwwwwww
yyoyyoyyobbbbbbbbbrryrryrrygggggggggwoowoowoowwrwwrwwr
yyyyyyyyybbbbbbbbbrrrrrrrrrgggggggggooooooooowwwwwwwww

%% file: main.bbl
\begin{thebibliography}{87}
\providecommand{\natexlab}[1]{#1}
\providecommand{\url}[1]{\texttt{#1}}
\expandafter\ifx\csname urlstyle\endcsname\relax
  \providecommand{\doi}[1]{doi: #1}\else
  \providecommand{\doi}{doi: \begingroup \urlstyle{rm}\Url}\fi

\bibitem[Achiam et~al.(2017)Achiam, Held, Tamar, and Abbeel]{DBLP:conf/icml/AchiamHTA17}
Joshua Achiam, David Held, Aviv Tamar, and Pieter Abbeel.
\newblock Constrained policy optimization.
\newblock In Doina Precup and Yee~Whye Teh (eds.), \emph{Proceedings of the 34th International Conference on Machine Learning, {ICML} 2017, Sydney, NSW, Australia, 6-11 August 2017}, volume~70 of \emph{Proceedings of Machine Learning Research}, pp.\  22--31. {PMLR}, 2017.
\newblock URL \url{http://proceedings.mlr.press/v70/achiam17a.html}.

\bibitem[Andrychowicz et~al.(2020)Andrychowicz, Raichuk, Stanczyk, Orsini, Girgin, Marinier, Hussenot, Geist, Pietquin, Michalski, Gelly, and Bachem]{what_matters}
Marcin Andrychowicz, Anton Raichuk, Piotr Stanczyk, Manu Orsini, Sertan Girgin, Rapha{\"{e}}l Marinier, L{\'{e}}onard Hussenot, Matthieu Geist, Olivier Pietquin, Marcin Michalski, Sylvain Gelly, and Olivier Bachem.
\newblock What matters in on-policy reinforcement learning? {A} large-scale empirical study.
\newblock \emph{CoRR}, abs/2006.05990, 2020.
\newblock URL \url{https://arxiv.org/abs/2006.05990}.

\bibitem[Applegate et~al.(2006)Applegate, Bixby, Chv{\'a}tal, and Cook]{applegate2006traveling}
David~L Applegate, Robert~E Bixby, V{\'a}clav Chv{\'a}tal, and William~J Cook.
\newblock \emph{The Traveling Salesman Problem: A Computational Study}.
\newblock Princeton University Press, 2006.

\bibitem[Baker et~al.(2009)Baker, Saxe, and Tenenbaum]{baker2009action}
Chris~L Baker, Rebecca Saxe, and Joshua~B Tenenbaum.
\newblock Action understanding as inverse planning.
\newblock \emph{Cognition}, 113\penalty0 (3):\penalty0 329--349, 2009.

\bibitem[Bengio et~al.(2021)Bengio, Lodi, and Prouvost]{bengio_combinatorial_optimization}
Yoshua Bengio, Andrea Lodi, and Antoine Prouvost.
\newblock Learning combinatorial optimization algorithms over graphs.
\newblock In \emph{Advances in Neural Information Processing Systems}, 2021.

\bibitem[Black et~al.(2024)Black, Nakamoto, Atreya, Walke, Finn, Kumar, and Levine]{DBLP:conf/iclr/BlackNAWFKL24}
Kevin Black, Mitsuhiko Nakamoto, Pranav Atreya, Homer~Rich Walke, Chelsea Finn, Aviral Kumar, and Sergey Levine.
\newblock Zero-shot robotic manipulation with pre-trained image-editing diffusion models.
\newblock In \emph{The Twelfth International Conference on Learning Representations, {ICLR} 2024, Vienna, Austria, May 7-11, 2024}. OpenReview.net, 2024.
\newblock URL \url{https://openreview.net/forum?id=c0chJTSbci}.

\bibitem[Bruck \& Goodman(1987)Bruck and Goodman]{DBLP:conf/nips/BruckG87}
Jehoshua Bruck and Joseph~W. Goodman.
\newblock On the power of neural networks for solving hard problems.
\newblock In Dana~Z. Anderson (ed.), \emph{Neural Information Processing Systems, Denver, Colorado, USA, 1987}, pp.\  137--143. American Institue of Physics, 1987.
\newblock URL \url{http://papers.nips.cc/paper/70-on-the-power-of-neural-networks-for-solving-hard-problems}.

\bibitem[Brunetto \& Trunda(2017)Brunetto and Trunda]{heuristics_rubic}
Robert Brunetto and Otakar Trunda.
\newblock Deep heuristic-learning in the rubik's cube domain: An experimental evaluation.
\newblock In Jaroslava Hlav{\'{a}}cov{\'{a}} (ed.), \emph{Proceedings of the 17th Conference on Information Technologies - Applications and Theory {(ITAT} 2017), Martinsk{\'{e}} hole, Slovakia, September 22-26, 2017}, volume 1885 of \emph{{CEUR} Workshop Proceedings}, pp.\  57--64. CEUR-WS.org, 2017.
\newblock URL \url{https://ceur-ws.org/Vol-1885/57.pdf}.

\bibitem[Campbell et~al.(2002)Campbell, Jr., and Hsu]{deepblue}
Murray Campbell, A.~Joseph~Hoane Jr., and Feng{-}Hsiung Hsu.
\newblock Deep blue.
\newblock \emph{Artif. Intell.}, 134\penalty0 (1-2):\penalty0 57--83, 2002.
\newblock \doi{10.1016/S0004-3702(01)00129-1}.
\newblock URL \url{https://doi.org/10.1016/S0004-3702(01)00129-1}.

\bibitem[Chen et~al.(2024)Chen, Deng, Kawaguchi, G{\"{u}}l{\c{c}}ehre, and Ahn]{DBLP:journals/corr/abs-2401-02644}
Chang Chen, Fei Deng, Kenji Kawaguchi, {\c{C}}aglar G{\"{u}}l{\c{c}}ehre, and Sungjin Ahn.
\newblock Simple hierarchical planning with diffusion.
\newblock \emph{CoRR}, abs/2401.02644, 2024.
\newblock \doi{10.48550/ARXIV.2401.02644}.
\newblock URL \url{https://doi.org/10.48550/arXiv.2401.02644}.

\bibitem[Chen et~al.(2021)Chen, Lu, Rajeswaran, Lee, Grover, Laskin, Abbeel, Srinivas, and Mordatch]{DBLP:conf/nips/ChenLRLGLASM21}
Lili Chen, Kevin Lu, Aravind Rajeswaran, Kimin Lee, Aditya Grover, Michael Laskin, Pieter Abbeel, Aravind Srinivas, and Igor Mordatch.
\newblock Decision transformer: Reinforcement learning via sequence modeling.
\newblock In Marc'Aurelio Ranzato, Alina Beygelzimer, Yann~N. Dauphin, Percy Liang, and Jennifer~Wortman Vaughan (eds.), \emph{Advances in Neural Information Processing Systems 34: Annual Conference on Neural Information Processing Systems 2021, NeurIPS 2021, December 6-14, 2021, virtual}, pp.\  15084--15097, 2021.
\newblock URL \url{https://proceedings.neurips.cc/paper/2021/hash/7f489f642a0ddb10272b5c31057f0663-Abstract.html}.

\bibitem[Choset et~al.(2005)Choset, Lynch, Hutchinson, Kantor, Burgard, Kavraki, and Thrun]{choset2005principles}
Howie Choset, Kevin~M. Lynch, Seth Hutchinson, George Kantor, Wolfram Burgard, Lydia~E. Kavraki, and Sebastian Thrun.
\newblock \emph{Principles of Robot Motion: Theory, Algorithms, and Implementations}.
\newblock MIT Press, Cambridge, MA, 2005.
\newblock ISBN 978-0-262-03327-5.

\bibitem[Collaboration et~al.(2023)Collaboration, O'Neill, Rehman, Maddukuri, Gupta, Padalkar, Lee, Pooley, Gupta, Mandlekar, Jain, Tung, Bewley, Herzog, Irpan, Khazatsky, Rai, Gupta, Wang, Singh, Garg, Kembhavi, Xie, Brohan, Raffin, Sharma, Yavary, Jain, Balakrishna, Wahid, Burgess-Limerick, Kim, Schölkopf, Wulfe, Ichter, Lu, Xu, Le, Finn, Wang, Xu, Chi, Huang, Chan, Agia, Pan, Fu, Devin, Xu, Morton, Driess, Chen, Pathak, Shah, Büchler, Jayaraman, Kalashnikov, Sadigh, Johns, Foster, Liu, Ceola, Xia, Zhao, Stulp, Zhou, Sukhatme, Salhotra, Yan, Feng, Schiavi, Berseth, Kahn, Wang, Su, Fang, Shi, Bao, Amor, Christensen, Furuta, Walke, Fang, Ha, Mordatch, Radosavovic, Leal, Liang, Abou-Chakra, Kim, Drake, Peters, Schneider, Hsu, Bohg, Bingham, Wu, Gao, Hu, Wu, Wu, Sun, Luo, Gu, Tan, Oh, Wu, Lu, Yang, Malik, Silvério, Hejna, Booher, Tompson, Yang, Salvador, Lim, Han, Wang, Rao, Pertsch, Hausman, Go, Gopalakrishnan, Goldberg, Byrne, Oslund, Kawaharazuka, Black, Lin, Zhang, Ehsani, Lekkala, Ellis, Rana,
  Srinivasan, Fang, Singh, Zeng, Hatch, Hsu, Itti, Chen, Pinto, Fei-Fei, Tan, Fan, Ott, Lee, Weihs, Chen, Lepert, Memmel, Tomizuka, Itkina, Castro, Spero, Du, Ahn, Yip, Zhang, Ding, Heo, Srirama, Sharma, Kim, Kanazawa, Hansen, Heess, Joshi, Suenderhauf, Liu, Palo, Shafiullah, Mees, Kroemer, Bastani, Sanketi, Miller, Yin, Wohlhart, Xu, Fagan, Mitrano, Sermanet, Abbeel, Sundaresan, Chen, Vuong, Rafailov, Tian, Doshi, Mart{'i}n-Mart{'i}n, Baijal, Scalise, Hendrix, Lin, Qian, Zhang, Mendonca, Shah, Hoque, Julian, Bustamante, Kirmani, Levine, Lin, Moore, Bahl, Dass, Sonawani, Song, Xu, Haldar, Karamcheti, Adebola, Guist, Nasiriany, Schaal, Welker, Tian, Ramamoorthy, Dasari, Belkhale, Park, Nair, Mirchandani, Osa, Gupta, Harada, Matsushima, Xiao, Kollar, Yu, Ding, Davchev, Zhao, Armstrong, Darrell, Chung, Jain, Vanhoucke, Zhan, Zhou, Burgard, Chen, Wang, Zhu, Geng, Liu, Liangwei, Li, Lu, Ma, Kim, Chebotar, Zhou, Zhu, Wu, Xu, Wang, Bisk, Cho, Lee, Cui, Cao, Wu, Tang, Zhu, Zhang, Jiang, Li, Li, Iwasawa, Matsuo, Ma,
  Xu, Cui, Zhang, and Lin]{open_x_embodiment_rt_x_2023}
Open X-Embodiment Collaboration, Abby O'Neill, Abdul Rehman, Abhiram Maddukuri, Abhishek Gupta, Abhishek Padalkar, Abraham Lee, Acorn Pooley, Agrim Gupta, Ajay Mandlekar, Ajinkya Jain, Albert Tung, Alex Bewley, Alex Herzog, Alex Irpan, Alexander Khazatsky, Anant Rai, Anchit Gupta, Andrew Wang, Anikait Singh, Animesh Garg, Aniruddha Kembhavi, Annie Xie, Anthony Brohan, Antonin Raffin, Archit Sharma, Arefeh Yavary, Arhan Jain, Ashwin Balakrishna, Ayzaan Wahid, Ben Burgess-Limerick, Beomjoon Kim, Bernhard Schölkopf, Blake Wulfe, Brian Ichter, Cewu Lu, Charles Xu, Charlotte Le, Chelsea Finn, Chen Wang, Chenfeng Xu, Cheng Chi, Chenguang Huang, Christine Chan, Christopher Agia, Chuer Pan, Chuyuan Fu, Coline Devin, Danfei Xu, Daniel Morton, Danny Driess, Daphne Chen, Deepak Pathak, Dhruv Shah, Dieter Büchler, Dinesh Jayaraman, Dmitry Kalashnikov, Dorsa Sadigh, Edward Johns, Ethan Foster, Fangchen Liu, Federico Ceola, Fei Xia, Feiyu Zhao, Freek Stulp, Gaoyue Zhou, Gaurav~S. Sukhatme, Gautam Salhotra, Ge~Yan,
  Gilbert Feng, Giulio Schiavi, Glen Berseth, Gregory Kahn, Guanzhi Wang, Hao Su, Hao-Shu Fang, Haochen Shi, Henghui Bao, Heni~Ben Amor, Henrik~I Christensen, Hiroki Furuta, Homer Walke, Hongjie Fang, Huy Ha, Igor Mordatch, Ilija Radosavovic, Isabel Leal, Jacky Liang, Jad Abou-Chakra, Jaehyung Kim, Jaimyn Drake, Jan Peters, Jan Schneider, Jasmine Hsu, Jeannette Bohg, Jeffrey Bingham, Jeffrey Wu, Jensen Gao, Jiaheng Hu, Jiajun Wu, Jialin Wu, Jiankai Sun, Jianlan Luo, Jiayuan Gu, Jie Tan, Jihoon Oh, Jimmy Wu, Jingpei Lu, Jingyun Yang, Jitendra Malik, João Silvério, Joey Hejna, Jonathan Booher, Jonathan Tompson, Jonathan Yang, Jordi Salvador, Joseph~J. Lim, Junhyek Han, Kaiyuan Wang, Kanishka Rao, Karl Pertsch, Karol Hausman, Keegan Go, Keerthana Gopalakrishnan, Ken Goldberg, Kendra Byrne, Kenneth Oslund, Kento Kawaharazuka, Kevin Black, Kevin Lin, Kevin Zhang, Kiana Ehsani, Kiran Lekkala, Kirsty Ellis, Krishan Rana, Krishnan Srinivasan, Kuan Fang, Kunal~Pratap Singh, Kuo-Hao Zeng, Kyle Hatch, Kyle Hsu,
  Laurent Itti, Lawrence~Yunliang Chen, Lerrel Pinto, Li~Fei-Fei, Liam Tan, Linxi~"Jim" Fan, Lionel Ott, Lisa Lee, Luca Weihs, Magnum Chen, Marion Lepert, Marius Memmel, Masayoshi Tomizuka, Masha Itkina, Mateo~Guaman Castro, Max Spero, Maximilian Du, Michael Ahn, Michael~C. Yip, Mingtong Zhang, Mingyu Ding, Minho Heo, Mohan~Kumar Srirama, Mohit Sharma, Moo~Jin Kim, Naoaki Kanazawa, Nicklas Hansen, Nicolas Heess, Nikhil~J Joshi, Niko Suenderhauf, Ning Liu, Norman~Di Palo, Nur Muhammad~Mahi Shafiullah, Oier Mees, Oliver Kroemer, Osbert Bastani, Pannag~R Sanketi, Patrick~"Tree" Miller, Patrick Yin, Paul Wohlhart, Peng Xu, Peter~David Fagan, Peter Mitrano, Pierre Sermanet, Pieter Abbeel, Priya Sundaresan, Qiuyu Chen, Quan Vuong, Rafael Rafailov, Ran Tian, Ria Doshi, Roberto Mart{'i}n-Mart{'i}n, Rohan Baijal, Rosario Scalise, Rose Hendrix, Roy Lin, Runjia Qian, Ruohan Zhang, Russell Mendonca, Rutav Shah, Ryan Hoque, Ryan Julian, Samuel Bustamante, Sean Kirmani, Sergey Levine, Shan Lin, Sherry Moore, Shikhar Bahl,
  Shivin Dass, Shubham Sonawani, Shuran Song, Sichun Xu, Siddhant Haldar, Siddharth Karamcheti, Simeon Adebola, Simon Guist, Soroush Nasiriany, Stefan Schaal, Stefan Welker, Stephen Tian, Subramanian Ramamoorthy, Sudeep Dasari, Suneel Belkhale, Sungjae Park, Suraj Nair, Suvir Mirchandani, Takayuki Osa, Tanmay Gupta, Tatsuya Harada, Tatsuya Matsushima, Ted Xiao, Thomas Kollar, Tianhe Yu, Tianli Ding, Todor Davchev, Tony~Z. Zhao, Travis Armstrong, Trevor Darrell, Trinity Chung, Vidhi Jain, Vincent Vanhoucke, Wei Zhan, Wenxuan Zhou, Wolfram Burgard, Xi~Chen, Xiaolong Wang, Xinghao Zhu, Xinyang Geng, Xiyuan Liu, Xu~Liangwei, Xuanlin Li, Yao Lu, Yecheng~Jason Ma, Yejin Kim, Yevgen Chebotar, Yifan Zhou, Yifeng Zhu, Yilin Wu, Ying Xu, Yixuan Wang, Yonatan Bisk, Yoonyoung Cho, Youngwoon Lee, Yuchen Cui, Yue Cao, Yueh-Hua Wu, Yujin Tang, Yuke Zhu, Yunchu Zhang, Yunfan Jiang, Yunshuang Li, Yunzhu Li, Yusuke Iwasawa, Yutaka Matsuo, Zehan Ma, Zhuo Xu, Zichen~Jeff Cui, Zichen Zhang, and Zipeng Lin.
\newblock Open {X-E}mbodiment: Robotic learning datasets and {RT-X} models.
\newblock \url{https://arxiv.org/abs/2310.08864}, 2023.

\bibitem[Cormen et~al.(2009)Cormen, Leiserson, Rivest, and Stein]{10.5555/1614191}
Thomas~H. Cormen, Charles~E. Leiserson, Ronald~L. Rivest, and Clifford Stein.
\newblock \emph{Introduction to Algorithms, Third Edition}.
\newblock The MIT Press, 3rd edition, 2009.
\newblock ISBN 0262033844.

\bibitem[Culberson(1997)]{pspace_soko}
Joseph~C. Culberson.
\newblock Sokoban is pspace-complete.
\newblock 1997.
\newblock URL \url{https://api.semanticscholar.org/CorpusID:61114368}.

\bibitem[Czechowski et~al.(2021)Czechowski, Odrzyg{\'{o}}zdz, Zbysinski, Zawalski, Olejnik, Wu, Kucinski, and Milos]{ksubs}
Konrad Czechowski, Tomasz Odrzyg{\'{o}}zdz, Marek Zbysinski, Michal Zawalski, Krzysztof Olejnik, Yuhuai Wu, Lukasz Kucinski, and Piotr Milos.
\newblock Subgoal search for complex reasoning tasks.
\newblock In Marc'Aurelio Ranzato, Alina Beygelzimer, Yann~N. Dauphin, Percy Liang, and Jennifer~Wortman Vaughan (eds.), \emph{Advances in Neural Information Processing Systems 34: Annual Conference on Neural Information Processing Systems 2021, NeurIPS 2021, December 6-14, 2021, virtual}, pp.\  624--638, 2021.
\newblock URL \url{https://proceedings.neurips.cc/paper/2021/hash/05d8cccb5f47e5072f0a05b5f514941a-Abstract.html}.

\bibitem[Demaine et~al.(2018)Demaine, Eisenstat, and Rudoy]{Demain}
Erik~D. Demaine, Sarah Eisenstat, and Mikhail Rudoy.
\newblock Solving the rubik's cube optimally is np-complete.
\newblock Schloss Dagstuhl – Leibniz-Zentrum für Informatik, 2018.
\newblock \doi{10.4230/LIPICS.STACS.2018.24}.
\newblock URL \url{https://drops.dagstuhl.de/entities/document/10.4230/LIPIcs.STACS.2018.24}.

\bibitem[Devlin et~al.(2019)Devlin, Chang, Lee, and Toutanova]{bert}
Jacob Devlin, Ming-Wei Chang, Kenton Lee, and Kristina Toutanova.
\newblock {BERT}: Pre-training of deep bidirectional transformers for language understanding.
\newblock In Jill Burstein, Christy Doran, and Thamar Solorio (eds.), \emph{Proceedings of the 2019 Conference of the North {A}merican Chapter of the Association for Computational Linguistics: Human Language Technologies, Volume 1 (Long and Short Papers)}, pp.\  4171--4186, Minneapolis, Minnesota, June 2019. Association for Computational Linguistics.
\newblock \doi{10.18653/v1/N19-1423}.
\newblock URL \url{https://aclanthology.org/N19-1423}.

\bibitem[Dosovitskiy et~al.(2017)Dosovitskiy, Ros, Codevilla, Lopez, and Koltun]{Dosovitskiy17}
Alexey Dosovitskiy, German Ros, Felipe Codevilla, Antonio Lopez, and Vladlen Koltun.
\newblock {CARLA}: {An} open urban driving simulator.
\newblock In \emph{Proceedings of the 1st Annual Conference on Robot Learning}, pp.\  1--16, 2017.

\bibitem[Dulac{-}Arnold et~al.(2015)Dulac{-}Arnold, Evans, Sunehag, and Coppin]{DBLP:journals/corr/Dulac-ArnoldESC15}
Gabriel Dulac{-}Arnold, Richard Evans, Peter Sunehag, and Ben Coppin.
\newblock Reinforcement learning in large discrete action spaces.
\newblock \emph{CoRR}, abs/1512.07679, 2015.
\newblock URL \url{http://arxiv.org/abs/1512.07679}.

\bibitem[Edmonds et~al.(2017)Edmonds, Gao, Xie, Liu, Qi, Zhu, Rothrock, and Zhu]{8206196}
Mark Edmonds, Feng Gao, Xu~Xie, Hangxin Liu, Siyuan Qi, Yixin Zhu, Brandon Rothrock, and Song-Chun Zhu.
\newblock Feeling the force: Integrating force and pose for fluent discovery through imitation learning to open medicine bottles.
\newblock In \emph{2017 IEEE/RSJ International Conference on Intelligent Robots and Systems (IROS)}, pp.\  3530--3537, 2017.
\newblock \doi{10.1109/IROS.2017.8206196}.

\bibitem[Eysenbach et~al.(2019)Eysenbach, Salakhutdinov, and Levine]{sorb}
Ben Eysenbach, Russ~R Salakhutdinov, and Sergey Levine.
\newblock Search on the replay buffer: Bridging planning and reinforcement learning.
\newblock In H.~Wallach, H.~Larochelle, A.~Beygelzimer, F.~d\textquotesingle Alch\'{e}-Buc, E.~Fox, and R.~Garnett (eds.), \emph{Advances in Neural Information Processing Systems}, volume~32. Curran Associates, Inc., 2019.
\newblock URL \url{https://proceedings.neurips.cc/paper_files/paper/2019/file/5c48ff18e0a47baaf81d8b8ea51eec92-Paper.pdf}.

\bibitem[Fatemi et~al.(2021)Fatemi, Killian, Subramanian, and Ghassemi]{DBLP:conf/nips/FatemiKSG21}
Mehdi Fatemi, Taylor~W. Killian, Jayakumar Subramanian, and Marzyeh Ghassemi.
\newblock Medical dead-ends and learning to identify high-risk states and treatments.
\newblock In Marc'Aurelio Ranzato, Alina Beygelzimer, Yann~N. Dauphin, Percy Liang, and Jennifer~Wortman Vaughan (eds.), \emph{Advances in Neural Information Processing Systems 34: Annual Conference on Neural Information Processing Systems 2021, NeurIPS 2021, December 6-14, 2021, virtual}, pp.\  4856--4870, 2021.
\newblock URL \url{https://proceedings.neurips.cc/paper/2021/hash/26405399c51ad7b13b504e74eb7c696c-Abstract.html}.

\bibitem[Feng et~al.(2022)Feng, Gomes, and Selman]{feng2022left}
Dieqiao Feng, Carla~P Gomes, and Bart Selman.
\newblock Left heavy tails and the effectiveness of the policy and value networks in {DNN}-based best-first search for sokoban planning.
\newblock In Alice~H. Oh, Alekh Agarwal, Danielle Belgrave, and Kyunghyun Cho (eds.), \emph{Advances in Neural Information Processing Systems}, 2022.
\newblock URL \url{https://openreview.net/forum?id=b6to5kfFhQh}.

\bibitem[Fickinger et~al.(2022)Fickinger, Cohen, Russell, and Amos]{fickinger2022crossdomain}
Arnaud Fickinger, Samuel Cohen, Stuart Russell, and Brandon Amos.
\newblock Cross-domain imitation learning via optimal transport.
\newblock In \emph{International Conference on Learning Representations}, 2022.
\newblock URL \url{https://openreview.net/forum?id=xP3cPq2hQC}.

\bibitem[Fishbach \& Dhar(2005)Fishbach and Dhar]{Fishbach2005}
Ayelet Fishbach and Ravi Dhar.
\newblock Goals as excuses or guides: The liberating effect of perceived goal progress on choice.
\newblock \emph{Journal of Consumer Research}, 32\penalty0 (3):\penalty0 370--377, 2005.

\bibitem[Fu et~al.(2020)Fu, Kumar, Nachum, Tucker, and Levine]{DBLP:journals/corr/abs-2004-07219}
Justin Fu, Aviral Kumar, Ofir Nachum, George Tucker, and Sergey Levine.
\newblock {D4RL:} datasets for deep data-driven reinforcement learning.
\newblock \emph{CoRR}, abs/2004.07219, 2020.
\newblock URL \url{https://arxiv.org/abs/2004.07219}.

\bibitem[Ghavamzadeh \& Mahadevan(2003)Ghavamzadeh and Mahadevan]{DBLP:conf/icml/GhavamzadehM03}
Mohammad Ghavamzadeh and Sridhar Mahadevan.
\newblock Hierarchical policy gradient algorithms.
\newblock In Tom Fawcett and Nina Mishra (eds.), \emph{Machine Learning, Proceedings of the Twentieth International Conference {(ICML} 2003), August 21-24, 2003, Washington, DC, {USA}}, pp.\  226--233. {AAAI} Press, 2003.
\newblock URL \url{http://www.aaai.org/Library/ICML/2003/icml03-032.php}.

\bibitem[Grauman et~al.(2022)Grauman, Westbury, Byrne, Chavis, Furnari, Girdhar, Hamburger, Jiang, Liu, Liu, Martin, Nagarajan, Radosavovic, Ramakrishnan, Ryan, Sharma, Wray, Xu, Xu, Zhao, Bansal, Batra, Cartillier, Crane, Do, Doulaty, Erapalli, Feichtenhofer, Fragomeni, Fu, Gebreselasie, Gonzalez, Hillis, Huang, Huang, Jia, Khoo, Kolar, Kottur, Kumar, Landini, Li, Li, Li, Mangalam, Modhugu, Munro, Murrell, Nishiyasu, Price, Puentes, Ramazanova, Sari, Somasundaram, Southerland, Sugano, Tao, Vo, Wang, Wu, Yagi, Zhao, Zhu, Arbelaez, Crandall, Damen, Farinella, Fuegen, Ghanem, Ithapu, Jawahar, Joo, Kitani, Li, Newcombe, Oliva, Park, Rehg, Sato, Shi, Shou, Torralba, Torresani, Yan, and Malik]{grauman2022ego4d}
Kristen Grauman, Andrew Westbury, Eugene Byrne, Zachary Chavis, Antonino Furnari, Rohit Girdhar, Jackson Hamburger, Hao Jiang, Miao Liu, Xingyu Liu, Miguel Martin, Tushar Nagarajan, Ilija Radosavovic, Santhosh~Kumar Ramakrishnan, Fiona Ryan, Jayant Sharma, Michael Wray, Mengmeng Xu, Eric~Zhongcong Xu, Chen Zhao, Siddhant Bansal, Dhruv Batra, Vincent Cartillier, Sean Crane, Tien Do, Morrie Doulaty, Akshay Erapalli, Christoph Feichtenhofer, Adriano Fragomeni, Qichen Fu, Abrham Gebreselasie, Cristina Gonzalez, James Hillis, Xuhua Huang, Yifei Huang, Wenqi Jia, Weslie Khoo, Jachym Kolar, Satwik Kottur, Anurag Kumar, Federico Landini, Chao Li, Yanghao Li, Zhenqiang Li, Karttikeya Mangalam, Raghava Modhugu, Jonathan Munro, Tullie Murrell, Takumi Nishiyasu, Will Price, Paola~Ruiz Puentes, Merey Ramazanova, Leda Sari, Kiran Somasundaram, Audrey Southerland, Yusuke Sugano, Ruijie Tao, Minh Vo, Yuchen Wang, Xindi Wu, Takuma Yagi, Ziwei Zhao, Yunyi Zhu, Pablo Arbelaez, David Crandall, Dima Damen, Giovanni~Maria
  Farinella, Christian Fuegen, Bernard Ghanem, Vamsi~Krishna Ithapu, C.~V. Jawahar, Hanbyul Joo, Kris Kitani, Haizhou Li, Richard Newcombe, Aude Oliva, Hyun~Soo Park, James~M. Rehg, Yoichi Sato, Jianbo Shi, Mike~Zheng Shou, Antonio Torralba, Lorenzo Torresani, Mingfei Yan, and Jitendra Malik.
\newblock Ego4d: Around the world in 3,000 hours of egocentric video, 2022.

\bibitem[Guez et~al.(2018)Guez, Mirza, Gregor, Kabra, Racaniere, Weber, Raposo, Santoro, Orseau, Eccles, Wayne, Silver, Lillicrap, and Valdes]{boxobanlevels}
Arthur Guez, Mehdi Mirza, Karol Gregor, Rishabh Kabra, Sebastien Racaniere, Theophane Weber, David Raposo, Adam Santoro, Laurent Orseau, Tom Eccles, Greg Wayne, David Silver, Timothy Lillicrap, and Victor Valdes.
\newblock An investigation of model-free planning: boxoban levels.
\newblock https://github.com/deepmind/boxoban-levels/, 2018.

\bibitem[Haslum et~al.(2019)Haslum, Lipovetzky, Magazzeni, and Muise]{DBLP:series/synthesis/2019Haslum}
Patrik Haslum, Nir Lipovetzky, Daniele Magazzeni, and Christian Muise.
\newblock \emph{An Introduction to the Planning Domain Definition Language}.
\newblock Synthesis Lectures on Artificial Intelligence and Machine Learning. Morgan {\&} Claypool Publishers, 2019.
\newblock ISBN 978-3-031-00456-8.
\newblock \doi{10.2200/S00900ED2V01Y201902AIM042}.
\newblock URL \url{https://doi.org/10.2200/S00900ED2V01Y201902AIM042}.

\bibitem[Huang et~al.(2019)Huang, Liu, and Su]{DBLP:conf/nips/HuangLS19}
Zhiao Huang, Fangchen Liu, and Hao Su.
\newblock Mapping state space using landmarks for universal goal reaching.
\newblock In Hanna~M. Wallach, Hugo Larochelle, Alina Beygelzimer, Florence d'Alch{\'{e}}{-}Buc, Emily~B. Fox, and Roman Garnett (eds.), \emph{Advances in Neural Information Processing Systems 32: Annual Conference on Neural Information Processing Systems 2019, NeurIPS 2019, December 8-14, 2019, Vancouver, BC, Canada}, pp.\  1940--1950, 2019.
\newblock URL \url{https://proceedings.neurips.cc/paper/2019/hash/3b712de48137572f3849aabd5666a4e3-Abstract.html}.

\bibitem[Hull(1932)]{Hull1932}
Clark~L. Hull.
\newblock The goal gradient hypothesis and maze learning.
\newblock \emph{Psychological Review}, 39\penalty0 (1):\penalty0 25--43, 1932.

\bibitem[James et~al.(2017)James, Konidaris, and Rosman]{James_Konidaris_Rosman_2017}
Steven James, George Konidaris, and Benjamin Rosman.
\newblock An analysis of monte carlo tree search.
\newblock \emph{Proceedings of the AAAI Conference on Artificial Intelligence}, 31\penalty0 (1), Feb. 2017.
\newblock \doi{10.1609/aaai.v31i1.11028}.
\newblock URL \url{https://ojs.aaai.org/index.php/AAAI/article/view/11028}.

\bibitem[Jiang et~al.(2019)Jiang, Gu, Murphy, and Finn]{DBLP:conf/nips/JiangGMF19}
Yiding Jiang, Shixiang Gu, Kevin Murphy, and Chelsea Finn.
\newblock Language as an abstraction for hierarchical deep reinforcement learning.
\newblock In Hanna~M. Wallach, Hugo Larochelle, Alina Beygelzimer, Florence d'Alch{\'{e}}{-}Buc, Emily~B. Fox, and Roman Garnett (eds.), \emph{Advances in Neural Information Processing Systems 32: Annual Conference on Neural Information Processing Systems 2019, NeurIPS 2019, December 8-14, 2019, Vancouver, BC, Canada}, pp.\  9414--9426, 2019.
\newblock URL \url{https://proceedings.neurips.cc/paper/2019/hash/0af787945872196b42c9f73ead2565c8-Abstract.html}.

\bibitem[Kelly et~al.(2019)Kelly, Sidrane, Driggs-Campbell, and Kochenderfer]{8793698}
Michael Kelly, Chelsea Sidrane, Katherine Driggs-Campbell, and Mykel~J. Kochenderfer.
\newblock Hg-dagger: Interactive imitation learning with human experts.
\newblock In \emph{2019 International Conference on Robotics and Automation (ICRA)}, pp.\  8077--8083, 2019.
\newblock \doi{10.1109/ICRA.2019.8793698}.

\bibitem[Kim et~al.(2020)Kim, Gu, Song, Zhao, and Ermon]{pmlr-v119-kim20c}
Kuno Kim, Yihong Gu, Jiaming Song, Shengjia Zhao, and Stefano Ermon.
\newblock Domain adaptive imitation learning.
\newblock In Hal~Daumé III and Aarti Singh (eds.), \emph{Proceedings of the 37th International Conference on Machine Learning}, volume 119 of \emph{Proceedings of Machine Learning Research}, pp.\  5286--5295. PMLR, 13--18 Jul 2020.
\newblock URL \url{https://proceedings.mlr.press/v119/kim20c.html}.

\bibitem[Kim et~al.(2024)Kim, Pertsch, Karamcheti, Xiao, Balakrishna, Nair, Rafailov, Foster, Lam, Sanketi, Vuong, Kollar, Burchfiel, Tedrake, Sadigh, Levine, Liang, and Finn]{kim24openvla}
{Moo Jin} Kim, Karl Pertsch, Siddharth Karamcheti, Ted Xiao, Ashwin Balakrishna, Suraj Nair, Rafael Rafailov, Ethan Foster, Grace Lam, Pannag Sanketi, Quan Vuong, Thomas Kollar, Benjamin Burchfiel, Russ Tedrake, Dorsa Sadigh, Sergey Levine, Percy Liang, and Chelsea Finn.
\newblock Openvla: An open-source vision-language-action model.
\newblock \emph{arXiv preprint arXiv:2406.09246}, 2024.

\bibitem[Kipf \& Welling(2017)Kipf and Welling]{DBLP:conf/iclr/KipfW17}
Thomas~N. Kipf and Max Welling.
\newblock Semi-supervised classification with graph convolutional networks.
\newblock In \emph{5th International Conference on Learning Representations, {ICLR} 2017, Toulon, France, April 24-26, 2017, Conference Track Proceedings}. OpenReview.net, 2017.
\newblock URL \url{https://openreview.net/forum?id=SJU4ayYgl}.

\bibitem[Kiran et~al.(2022)Kiran, Sobh, Talpaert, Mannion, Sallab, Yogamani, and P{\'{e}}rez]{DBLP:journals/tits/KiranSTMSYP22}
B.~Ravi Kiran, Ibrahim Sobh, Victor Talpaert, Patrick Mannion, Ahmad A.~Al Sallab, Senthil~Kumar Yogamani, and Patrick P{\'{e}}rez.
\newblock Deep reinforcement learning for autonomous driving: {A} survey.
\newblock \emph{{IEEE} Trans. Intell. Transp. Syst.}, 23\penalty0 (6):\penalty0 4909--4926, 2022.
\newblock \doi{10.1109/TITS.2021.3054625}.
\newblock URL \url{https://doi.org/10.1109/TITS.2021.3054625}.

\bibitem[Kirk et~al.(2023)Kirk, Zhang, Grefenstette, and Rockt{\"{a}}schel]{DBLP:journals/jair/KirkZGR23}
Robert Kirk, Amy Zhang, Edward Grefenstette, and Tim Rockt{\"{a}}schel.
\newblock A survey of zero-shot generalisation in deep reinforcement learning.
\newblock \emph{J. Artif. Intell. Res.}, 76:\penalty0 201--264, 2023.
\newblock \doi{10.1613/JAIR.1.14174}.
\newblock URL \url{https://doi.org/10.1613/jair.1.14174}.

\bibitem[Kool \& Botvinick(2014)Kool and Botvinick]{Kool2014}
Wouter Kool and Matthew Botvinick.
\newblock A labor/leisure tradeoff in cognitive control.
\newblock \emph{Journal of Experimental Psychology: General}, 143\penalty0 (1):\penalty0 131--141, 2014.

\bibitem[Kujanp{\"{a}}{\"{a}} et~al.(2023{\natexlab{a}})Kujanp{\"{a}}{\"{a}}, Pajarinen, and Ilin]{hips}
Kalle Kujanp{\"{a}}{\"{a}}, Joni Pajarinen, and Alexander Ilin.
\newblock Hierarchical imitation learning with vector quantized models.
\newblock In Andreas Krause, Emma Brunskill, Kyunghyun Cho, Barbara Engelhardt, Sivan Sabato, and Jonathan Scarlett (eds.), \emph{International Conference on Machine Learning, {ICML} 2023, 23-29 July 2023, Honolulu, Hawaii, {USA}}, volume 202 of \emph{Proceedings of Machine Learning Research}, pp.\  17896--17919. {PMLR}, 2023{\natexlab{a}}.
\newblock URL \url{https://proceedings.mlr.press/v202/kujanpaa23a.html}.

\bibitem[Kujanp{\"{a}}{\"{a}} et~al.(2023{\natexlab{b}})Kujanp{\"{a}}{\"{a}}, Pajarinen, and Ilin]{hipseps}
Kalle Kujanp{\"{a}}{\"{a}}, Joni Pajarinen, and Alexander Ilin.
\newblock Hybrid search for efficient planning with completeness guarantees.
\newblock \emph{CoRR}, abs/2310.12819, 2023{\natexlab{b}}.
\newblock \doi{10.48550/ARXIV.2310.12819}.
\newblock URL \url{https://doi.org/10.48550/arXiv.2310.12819}.

\bibitem[Kumar et~al.(2022)Kumar, Hong, Singh, and Levine]{ORL_vs_BC}
Aviral Kumar, Joey Hong, Anikait Singh, and Sergey Levine.
\newblock When should we prefer offline reinforcement learning over behavioral cloning?
\newblock \emph{CoRR}, abs/2204.05618, 2022.
\newblock \doi{10.48550/ARXIV.2204.05618}.
\newblock URL \url{https://doi.org/10.48550/arXiv.2204.05618}.

\bibitem[LaValle(2006)]{lavalle2006planning}
Steven~M LaValle.
\newblock \emph{Planning algorithms}.
\newblock Cambridge university press, 2006.

\bibitem[Lee et~al.(2022)Lee, Kim, Jang, and Kim]{DBLP:conf/nips/LeeKJK22}
Seungjae Lee, Jigang Kim, Inkyu Jang, and H.~Jin Kim.
\newblock {DHRL:} {A} graph-based approach for long-horizon and sparse hierarchical reinforcement learning.
\newblock In Sanmi Koyejo, S.~Mohamed, A.~Agarwal, Danielle Belgrave, K.~Cho, and A.~Oh (eds.), \emph{Advances in Neural Information Processing Systems 35: Annual Conference on Neural Information Processing Systems 2022, NeurIPS 2022, New Orleans, LA, USA, November 28 - December 9, 2022}, 2022.
\newblock URL \url{http://papers.nips.cc/paper\_files/paper/2022/hash/58b286aea34a91a3d33e58af0586fa40-Abstract-Conference.html}.

\bibitem[Levine et~al.(2020)Levine, Kumar, Tucker, and Fu]{DBLP:journals/corr/abs-2005-01643}
Sergey Levine, Aviral Kumar, George Tucker, and Justin Fu.
\newblock Offline reinforcement learning: Tutorial, review, and perspectives on open problems.
\newblock \emph{CoRR}, abs/2005.01643, 2020.
\newblock URL \url{https://arxiv.org/abs/2005.01643}.

\bibitem[Levy et~al.(2019)Levy, Konidaris, Jr., and Saenko]{DBLP:conf/iclr/LevyKPS19}
Andrew Levy, George~Dimitri Konidaris, Robert~Platt Jr., and Kate Saenko.
\newblock Learning multi-level hierarchies with hindsight.
\newblock In \emph{7th International Conference on Learning Representations, {ICLR} 2019, New Orleans, LA, USA, May 6-9, 2019}. OpenReview.net, 2019.
\newblock URL \url{https://openreview.net/forum?id=ryzECoAcY7}.

\bibitem[Lewis et~al.(2020)Lewis, Liu, Goyal, Ghazvininejad, Mohamed, Levy, Stoyanov, and Zettlemoyer]{bart}
Mike Lewis, Yinhan Liu, Naman Goyal, Marjan Ghazvininejad, Abdelrahman Mohamed, Omer Levy, Veselin Stoyanov, and Luke Zettlemoyer.
\newblock {BART}: Denoising sequence-to-sequence pre-training for natural language generation, translation, and comprehension.
\newblock In Dan Jurafsky, Joyce Chai, Natalie Schluter, and Joel Tetreault (eds.), \emph{Proceedings of the 58th Annual Meeting of the Association for Computational Linguistics}, pp.\  7871--7880, Online, July 2020. Association for Computational Linguistics.
\newblock \doi{10.18653/v1/2020.acl-main.703}.
\newblock URL \url{https://aclanthology.org/2020.acl-main.703}.

\bibitem[Li et~al.(2022)Li, Peng, and Zhou]{li2022efficient}
Quanyi Li, Zhenghao Peng, and Bolei Zhou.
\newblock Efficient learning of safe driving policy via human-ai copilot optimization.
\newblock In \emph{International Conference on Learning Representations}, 2022.
\newblock URL \url{https://openreview.net/forum?id=0cgU-BZp2ky}.

\bibitem[Mandlekar et~al.(2018)Mandlekar, Zhu, Garg, Booher, Spero, Tung, Gao, Emmons, Gupta, Orbay, Savarese, and Fei{-}Fei]{DBLP:conf/corl/MandlekarZGBSTG18}
Ajay Mandlekar, Yuke Zhu, Animesh Garg, Jonathan Booher, Max Spero, Albert Tung, Julian Gao, John Emmons, Anchit Gupta, Emre Orbay, Silvio Savarese, and Li~Fei{-}Fei.
\newblock {ROBOTURK:} {A} crowdsourcing platform for robotic skill learning through imitation.
\newblock In \emph{2nd Annual Conference on Robot Learning, CoRL 2018, Z{\"{u}}rich, Switzerland, 29-31 October 2018, Proceedings}, volume~87 of \emph{Proceedings of Machine Learning Research}, pp.\  879--893. {PMLR}, 2018.
\newblock URL \url{http://proceedings.mlr.press/v87/mandlekar18a.html}.

\bibitem[McAleer et~al.(2019)McAleer, Agostinelli, Shmakov, and Baldi]{agnostelli}
Stephen McAleer, Forest Agostinelli, Alexander Shmakov, and Pierre Baldi.
\newblock Solving the rubik's cube with approximate policy iteration.
\newblock In \emph{7th International Conference on Learning Representations, {ICLR} 2019, New Orleans, LA, USA, May 6-9, 2019}. OpenReview.net, 2019.
\newblock URL \url{https://openreview.net/forum?id=Hyfn2jCcKm}.

\bibitem[Mees et~al.(2022)Mees, Hermann, and Burgard]{DBLP:journals/ral/MeesHB22}
Oier Mees, Luk{\'{a}}s Hermann, and Wolfram Burgard.
\newblock What matters in language conditioned robotic imitation learning over unstructured data.
\newblock \emph{{IEEE} Robotics Autom. Lett.}, 7\penalty0 (4):\penalty0 11205--11212, 2022.
\newblock \doi{10.1109/LRA.2022.3196123}.
\newblock URL \url{https://doi.org/10.1109/LRA.2022.3196123}.

\bibitem[Mnih et~al.(2015)Mnih, Kavukcuoglu, Silver, Rusu, Veness, Bellemare, Graves, Riedmiller, Fidjeland, Ostrovski, Petersen, et~al.]{Mnih2015}
Volodymyr Mnih, Koray Kavukcuoglu, David Silver, Andrei~A Rusu, Joel Veness, Marc~G Bellemare, Alex Graves, Martin Riedmiller, Andreas~K Fidjeland, Georg Ostrovski, Stig Petersen, et~al.
\newblock Human-level control through deep reinforcement learning.
\newblock \emph{Nature}, 518\penalty0 (7540):\penalty0 529--533, 2015.

\bibitem[Nachum et~al.(2018)Nachum, Gu, Lee, and Levine]{DBLP:conf/nips/NachumGLL18}
Ofir Nachum, Shixiang Gu, Honglak Lee, and Sergey Levine.
\newblock Data-efficient hierarchical reinforcement learning.
\newblock In Samy Bengio, Hanna~M. Wallach, Hugo Larochelle, Kristen Grauman, Nicol{\`{o}} Cesa{-}Bianchi, and Roman Garnett (eds.), \emph{Advances in Neural Information Processing Systems 31: Annual Conference on Neural Information Processing Systems 2018, NeurIPS 2018, December 3-8, 2018, Montr{\'{e}}al, Canada}, pp.\  3307--3317, 2018.
\newblock URL \url{https://proceedings.neurips.cc/paper/2018/hash/e6384711491713d29bc63fc5eeb5ba4f-Abstract.html}.

\bibitem[Nair et~al.(2018)Nair, McGrew, Andrychowicz, Zaremba, and Abbeel]{DBLP:conf/icra/NairMAZA18}
Ashvin Nair, Bob McGrew, Marcin Andrychowicz, Wojciech Zaremba, and Pieter Abbeel.
\newblock Overcoming exploration in reinforcement learning with demonstrations.
\newblock In \emph{2018 {IEEE} International Conference on Robotics and Automation, {ICRA} 2018, Brisbane, Australia, May 21-25, 2018}, pp.\  6292--6299. {IEEE}, 2018.
\newblock \doi{10.1109/ICRA.2018.8463162}.
\newblock URL \url{https://doi.org/10.1109/ICRA.2018.8463162}.

\bibitem[Needleman \& Wunsch(1970)Needleman and Wunsch]{needleman1970general}
Saul~B Needleman and Christian~D Wunsch.
\newblock A general method applicable to the search for similarities in the amino acid sequence of two proteins.
\newblock \emph{Journal of molecular biology}, 48\penalty0 (3):\penalty0 443--453, 1970.

\bibitem[Orseau \& Lelis(2021)Orseau and Lelis]{DBLP:conf/aaai/OrseauL21}
Laurent Orseau and Levi H.~S. Lelis.
\newblock Policy-guided heuristic search with guarantees.
\newblock In \emph{Thirty-Fifth {AAAI} Conference on Artificial Intelligence, {AAAI} 2021, Thirty-Third Conference on Innovative Applications of Artificial Intelligence, {IAAI} 2021, The Eleventh Symposium on Educational Advances in Artificial Intelligence, {EAAI} 2021, Virtual Event, February 2-9, 2021}, pp.\  12382--12390. {AAAI} Press, 2021.
\newblock \doi{10.1609/AAAI.V35I14.17469}.
\newblock URL \url{https://doi.org/10.1609/aaai.v35i14.17469}.

\bibitem[Orseau et~al.(2023)Orseau, Hutter, and Lelis]{ijcai2023p0624}
Laurent Orseau, Marcus Hutter, and Levi H.~S. Lelis.
\newblock Levin tree search with context models.
\newblock In Edith Elkind (ed.), \emph{Proceedings of the Thirty-Second International Joint Conference on Artificial Intelligence, {IJCAI-23}}, pp.\  5622--5630. International Joint Conferences on Artificial Intelligence Organization, 8 2023.
\newblock \doi{10.24963/ijcai.2023/624}.
\newblock URL \url{https://doi.org/10.24963/ijcai.2023/624}.
\newblock Main Track.

\bibitem[Panov \& Skrynnik(2018)Panov and Skrynnik]{DBLP:journals/corr/abs-1806-05292}
Aleksandr~I. Panov and Aleksey Skrynnik.
\newblock Automatic formation of the structure of abstract machines in hierarchical reinforcement learning with state clustering.
\newblock \emph{CoRR}, abs/1806.05292, 2018.
\newblock URL \url{http://arxiv.org/abs/1806.05292}.

\bibitem[Park et~al.(2023)Park, Ghosh, Eysenbach, and Levine]{DBLP:conf/nips/ParkGEL23}
Seohong Park, Dibya Ghosh, Benjamin Eysenbach, and Sergey Levine.
\newblock {HIQL:} offline goal-conditioned {RL} with latent states as actions.
\newblock In Alice Oh, Tristan Naumann, Amir Globerson, Kate Saenko, Moritz Hardt, and Sergey Levine (eds.), \emph{Advances in Neural Information Processing Systems 36: Annual Conference on Neural Information Processing Systems 2023, NeurIPS 2023, New Orleans, LA, USA, December 10 - 16, 2023}, 2023.
\newblock URL \url{http://papers.nips.cc/paper\_files/paper/2023/hash/6d7c4a0727e089ed6cdd3151cbe8d8ba-Abstract-Conference.html}.

\bibitem[Pertsch et~al.(2020)Pertsch, Rybkin, Ebert, Zhou, Jayaraman, Finn, and Levine]{DBLP:conf/nips/PertschREZJFL20}
Karl Pertsch, Oleh Rybkin, Frederik Ebert, Shenghao Zhou, Dinesh Jayaraman, Chelsea Finn, and Sergey Levine.
\newblock Long-horizon visual planning with goal-conditioned hierarchical predictors.
\newblock In Hugo Larochelle, Marc'Aurelio Ranzato, Raia Hadsell, Maria{-}Florina Balcan, and Hsuan{-}Tien Lin (eds.), \emph{Advances in Neural Information Processing Systems 33: Annual Conference on Neural Information Processing Systems 2020, NeurIPS 2020, December 6-12, 2020, virtual}, 2020.
\newblock URL \url{https://proceedings.neurips.cc/paper/2020/hash/c8d3a760ebab631565f8509d84b3b3f1-Abstract.html}.

\bibitem[Ratner \& Warmuth(1986)Ratner and Warmuth]{npuzzle}
Daniel Ratner and Manfred~K. Warmuth.
\newblock Finding a shortest solution for the {N} {\texttimes} {N} extension of the 15-puzzle is intractable.
\newblock In Tom Kehler (ed.), \emph{Proceedings of the 5th National Conference on Artificial Intelligence. Philadelphia, PA, USA, August 11-15, 1986. Volume 1: Science}, pp.\  168--172. Morgan Kaufmann, 1986.
\newblock URL \url{http://www.aaai.org/Library/AAAI/1986/aaai86-027.php}.

\bibitem[Ross et~al.(2011)Ross, Gordon, and Bagnell]{dagger}
St{\'{e}}phane Ross, Geoffrey~J. Gordon, and Drew Bagnell.
\newblock A reduction of imitation learning and structured prediction to no-regret online learning.
\newblock In Geoffrey~J. Gordon, David~B. Dunson, and Miroslav Dud{\'{\i}}k (eds.), \emph{Proceedings of the Fourteenth International Conference on Artificial Intelligence and Statistics, {AISTATS} 2011, Fort Lauderdale, USA, April 11-13, 2011}, volume~15 of \emph{{JMLR} Proceedings}, pp.\  627--635. JMLR.org, 2011.
\newblock URL \url{http://proceedings.mlr.press/v15/ross11a/ross11a.pdf}.

\bibitem[Russell \& Norvig(2009)Russell and Norvig]{10.5555/1671238}
Stuart Russell and Peter Norvig.
\newblock \emph{Artificial Intelligence: A Modern Approach}.
\newblock Prentice Hall Press, USA, 3rd edition, 2009.
\newblock ISBN 0136042597.

\bibitem[Russell \& Norvig(2020)Russell and Norvig]{russel_norvig}
Stuart Russell and Peter Norvig.
\newblock \emph{Artificial Intelligence: {A} Modern Approach (4th Edition)}.
\newblock Pearson, 2020.
\newblock ISBN 9780134610993.
\newblock URL \url{http://aima.cs.berkeley.edu/}.

\bibitem[Sahni(1974)]{DBLP:journals/siamcomp/Sahni74}
Sartaj Sahni.
\newblock Computationally related problems.
\newblock \emph{{SIAM} J. Comput.}, 3\penalty0 (4):\penalty0 262--279, 1974.
\newblock \doi{10.1137/0203021}.
\newblock URL \url{https://doi.org/10.1137/0203021}.

\bibitem[Schulman et~al.(2017)Schulman, Wolski, Dhariwal, Radford, and Klimov]{schulman2017proximal}
John Schulman, Filip Wolski, Prafulla Dhariwal, Alec Radford, and Oleg Klimov.
\newblock Proximal policy optimization algorithms, 2017.

\bibitem[Shen et~al.(2021)Shen, Liu, He, Zhang, Xu, Yu, and Cui]{DBLP:journals/corr/abs-2108-13624}
Zheyan Shen, Jiashuo Liu, Yue He, Xingxuan Zhang, Renzhe Xu, Han Yu, and Peng Cui.
\newblock Towards out-of-distribution generalization: {A} survey.
\newblock \emph{CoRR}, abs/2108.13624, 2021.
\newblock URL \url{https://arxiv.org/abs/2108.13624}.

\bibitem[Silver et~al.(2016)Silver, Huang, Maddison, Guez, Sifre, van~den Driessche, Schrittwieser, Antonoglou, Panneershelvam, Lanctot, Dieleman, Grewe, Nham, Kalchbrenner, Sutskever, Lillicrap, Leach, Kavukcuoglu, Graepel, and Hassabis]{alphago}
David Silver, Aja Huang, Chris~J. Maddison, Arthur Guez, Laurent Sifre, George van~den Driessche, Julian Schrittwieser, Ioannis Antonoglou, Vedavyas Panneershelvam, Marc Lanctot, Sander Dieleman, Dominik Grewe, John Nham, Nal Kalchbrenner, Ilya Sutskever, Timothy~P. Lillicrap, Madeleine Leach, Koray Kavukcuoglu, Thore Graepel, and Demis Hassabis.
\newblock Mastering the game of go with deep neural networks and tree search.
\newblock \emph{Nat.}, 529\penalty0 (7587):\penalty0 484--489, 2016.
\newblock \doi{10.1038/NATURE16961}.
\newblock URL \url{https://doi.org/10.1038/nature16961}.

\bibitem[Silver et~al.(2018)Silver, Hubert, Schrittwieser, Antonoglou, Lai, Guez, Lanctot, Sifre, Kumaran, Graepel, Lillicrap, Simonyan, and Hassabis]{alphazero}
David Silver, Thomas Hubert, Julian Schrittwieser, Ioannis Antonoglou, Matthew Lai, Arthur Guez, Marc Lanctot, Laurent Sifre, Dharshan Kumaran, Thore Graepel, Timothy Lillicrap, Karen Simonyan, and Demis Hassabis.
\newblock A general reinforcement learning algorithm that masters chess, shogi, and go through self-play.
\newblock \emph{Science}, 362\penalty0 (6419):\penalty0 1140--1144, 2018.
\newblock \doi{10.1126/science.aar6404}.
\newblock URL \url{https://www.science.org/doi/abs/10.1126/science.aar6404}.

\bibitem[Singmaster(1981)]{singmaster1981notes}
David Singmaster.
\newblock \emph{Notes on Rubik's Magic Cube}.
\newblock Enslow Publishers, 1981.

\bibitem[Smith \& Waterman(1981)Smith and Waterman]{smith1981identification}
Temple~F Smith and Michael~S Waterman.
\newblock Identification of common molecular subsequences.
\newblock \emph{Journal of molecular biology}, 147\penalty0 (1):\penalty0 195--197, 1981.

\bibitem[Sun et~al.(2020)Sun, Kretzschmar, Dotiwalla, Chouard, Patnaik, Tsui, Guo, Zhou, Chai, Caine, Vasudevan, Han, Ngiam, Zhao, Timofeev, Ettinger, Krivokon, Gao, Joshi, Zhang, Shlens, Chen, and Anguelov]{DBLP:conf/cvpr/SunKDCPTGZCCVHN20}
Pei Sun, Henrik Kretzschmar, Xerxes Dotiwalla, Aurelien Chouard, Vijaysai Patnaik, Paul Tsui, James Guo, Yin Zhou, Yuning Chai, Benjamin Caine, Vijay Vasudevan, Wei Han, Jiquan Ngiam, Hang Zhao, Aleksei Timofeev, Scott Ettinger, Maxim Krivokon, Amy Gao, Aditya Joshi, Yu~Zhang, Jonathon Shlens, Zhifeng Chen, and Dragomir Anguelov.
\newblock Scalability in perception for autonomous driving: Waymo open dataset.
\newblock In \emph{2020 {IEEE/CVF} Conference on Computer Vision and Pattern Recognition, {CVPR} 2020, Seattle, WA, USA, June 13-19, 2020}, pp.\  2443--2451. Computer Vision Foundation / {IEEE}, 2020.
\newblock \doi{10.1109/CVPR42600.2020.00252}.
\newblock URL \url{https://openaccess.thecvf.com/content\_CVPR\_2020/html/Sun\_Scalability\_in\_Perception\_for\_Autonomous\_Driving\_Waymo\_Open\_Dataset\_CVPR\_2020\_paper.html}.

\bibitem[Sutton \& Barto(1998)Sutton and Barto]{DBLP:books/lib/SuttonB98}
Richard~S. Sutton and Andrew~G. Barto.
\newblock \emph{Reinforcement learning - an introduction}.
\newblock Adaptive computation and machine learning. {MIT} Press, 1998.
\newblock ISBN 978-0-262-19398-6.
\newblock URL \url{https://www.worldcat.org/oclc/37293240}.

\bibitem[Sutton et~al.(1999)Sutton, Precup, and Singh]{DBLP:journals/ai/SuttonPS99}
Richard~S. Sutton, Doina Precup, and Satinder Singh.
\newblock Between mdps and semi-mdps: {A} framework for temporal abstraction in reinforcement learning.
\newblock \emph{Artif. Intell.}, 112\penalty0 (1-2):\penalty0 181--211, 1999.
\newblock \doi{10.1016/S0004-3702(99)00052-1}.
\newblock URL \url{https://doi.org/10.1016/S0004-3702(99)00052-1}.

\bibitem[Trinh et~al.(2024)Trinh, Wu, Le, He, and Luong]{AlphaGeometryTrinh2024}
Trieu Trinh, Yuhuai Wu, Quoc Le, He~He, and Thang Luong.
\newblock Solving olympiad geometry without human demonstrations.
\newblock \emph{Nature}, 2024.
\newblock \doi{10.1038/s41586-023-06747-5}.

\bibitem[van~den Oord et~al.(2017)van~den Oord, Vinyals, and Kavukcuoglu]{DBLP:conf/nips/OordVK17}
A{\"{a}}ron van~den Oord, Oriol Vinyals, and Koray Kavukcuoglu.
\newblock Neural discrete representation learning.
\newblock In Isabelle Guyon, Ulrike von Luxburg, Samy Bengio, Hanna~M. Wallach, Rob Fergus, S.~V.~N. Vishwanathan, and Roman Garnett (eds.), \emph{Advances in Neural Information Processing Systems 30: Annual Conference on Neural Information Processing Systems 2017, December 4-9, 2017, Long Beach, CA, {USA}}, pp.\  6306--6315, 2017.
\newblock URL \url{https://proceedings.neurips.cc/paper/2017/hash/7a98af17e63a0ac09ce2e96d03992fbc-Abstract.html}.

\bibitem[Veness et~al.(2009)Veness, Silver, Blair, and Uther]{NIPS2009_389bc7bb}
Joel Veness, David Silver, Alan Blair, and William Uther.
\newblock Bootstrapping from game tree search.
\newblock In Y.~Bengio, D.~Schuurmans, J.~Lafferty, C.~Williams, and A.~Culotta (eds.), \emph{Advances in Neural Information Processing Systems}, volume~22. Curran Associates, Inc., 2009.
\newblock URL \url{https://proceedings.neurips.cc/paper_files/paper/2009/file/389bc7bb1e1c2a5e7e147703232a88f6-Paper.pdf}.

\bibitem[Vinyals et~al.(2019)Vinyals, Babuschkin, Czarnecki, Mathieu, Dudzik, Chung, Choi, Powell, Ewalds, Georgiev, Oh, Horgan, Kroiss, Danihelka, Huang, Sifre, Cai, Agapiou, Jaderberg, Vezhnevets, Leblond, Pohlen, Dalibard, Budden, Sulsky, Molloy, Paine, G{\"{u}}l{\c{c}}ehre, Wang, Pfaff, Wu, Ring, Yogatama, W{\"{u}}nsch, McKinney, Smith, Schaul, Lillicrap, Kavukcuoglu, Hassabis, Apps, and Silver]{starcraft}
Oriol Vinyals, Igor Babuschkin, Wojciech~M. Czarnecki, Micha{\"{e}}l Mathieu, Andrew Dudzik, Junyoung Chung, David~H. Choi, Richard Powell, Timo Ewalds, Petko Georgiev, Junhyuk Oh, Dan Horgan, Manuel Kroiss, Ivo Danihelka, Aja Huang, Laurent Sifre, Trevor Cai, John~P. Agapiou, Max Jaderberg, Alexander~Sasha Vezhnevets, R{\'{e}}mi Leblond, Tobias Pohlen, Valentin Dalibard, David Budden, Yury Sulsky, James Molloy, Tom~Le Paine, {\c{C}}aglar G{\"{u}}l{\c{c}}ehre, Ziyu Wang, Tobias Pfaff, Yuhuai Wu, Roman Ring, Dani Yogatama, Dario W{\"{u}}nsch, Katrina McKinney, Oliver Smith, Tom Schaul, Timothy~P. Lillicrap, Koray Kavukcuoglu, Demis Hassabis, Chris Apps, and David Silver.
\newblock Grandmaster level in starcraft {II} using multi-agent reinforcement learning.
\newblock \emph{Nat.}, 575\penalty0 (7782):\penalty0 350--354, 2019.
\newblock \doi{10.1038/S41586-019-1724-Z}.
\newblock URL \url{https://doi.org/10.1038/s41586-019-1724-z}.

\bibitem[Walke et~al.(2023)Walke, Black, Lee, Kim, Du, Zheng, Zhao, Hansen-Estruch, Vuong, He, Myers, Fang, Finn, and Levine]{walke2023bridgedata}
Homer Walke, Kevin Black, Abraham Lee, Moo~Jin Kim, Max Du, Chongyi Zheng, Tony Zhao, Philippe Hansen-Estruch, Quan Vuong, Andre He, Vivek Myers, Kuan Fang, Chelsea Finn, and Sergey Levine.
\newblock Bridgedata v2: A dataset for robot learning at scale.
\newblock In \emph{Conference on Robot Learning (CoRL)}, 2023.

\bibitem[Wu et~al.(2021)Wu, Jiang, Ba, and Grosse]{int}
Yuhuai Wu, Albert Jiang, Jimmy Ba, and Roger~Baker Grosse.
\newblock {\{}INT{\}}: An inequality benchmark for evaluating generalization in theorem proving.
\newblock In \emph{International Conference on Learning Representations}, 2021.
\newblock URL \url{https://openreview.net/forum?id=O6LPudowNQm}.

\bibitem[Yang et~al.(2018)Yang, Merrick, Jin, and Abbass]{DBLP:journals/tnn/YangMJA18}
Zhaoyang Yang, Kathryn~E. Merrick, Lianwen Jin, and Hussein~A. Abbass.
\newblock Hierarchical deep reinforcement learning for continuous action control.
\newblock \emph{{IEEE} Trans. Neural Networks Learn. Syst.}, 29\penalty0 (11):\penalty0 5174--5184, 2018.
\newblock \doi{10.1109/TNNLS.2018.2805379}.
\newblock URL \url{https://doi.org/10.1109/TNNLS.2018.2805379}.

\bibitem[Yonetani et~al.(2021)Yonetani, Taniai, Barekatain, Nishimura, and Kanezaki]{DBLP:conf/icml/YonetaniTBNK21}
Ryo Yonetani, Tatsunori Taniai, Mohammadamin Barekatain, Mai Nishimura, and Asako Kanezaki.
\newblock Path planning using neural a* search.
\newblock In Marina Meila and Tong Zhang (eds.), \emph{Proceedings of the 38th International Conference on Machine Learning, {ICML} 2021, 18-24 July 2021, Virtual Event}, volume 139 of \emph{Proceedings of Machine Learning Research}, pp.\  12029--12039. {PMLR}, 2021.
\newblock URL \url{http://proceedings.mlr.press/v139/yonetani21a.html}.

\bibitem[Zawalski et~al.(2023)Zawalski, Tyrolski, Czechowski, Odrzyg{\'{o}}zdz, Stachura, Piekos, Wu, Kucinski, and Milos]{ada}
Michal Zawalski, Michal Tyrolski, Konrad Czechowski, Tomasz Odrzyg{\'{o}}zdz, Damian Stachura, Piotr Piekos, Yuhuai Wu, Lukasz Kucinski, and Piotr Milos.
\newblock Fast and precise: Adjusting planning horizon with adaptive subgoal search.
\newblock In \emph{The Eleventh International Conference on Learning Representations, {ICLR} 2023, Kigali, Rwanda, May 1-5, 2023}. OpenReview.net, 2023.
\newblock URL \url{https://openreview.net/pdf?id=7JsGYvjE88d}.

\bibitem[Zhang \& Cho(2017)Zhang and Cho]{10.5555/3298483.3298654}
Jiakai Zhang and Kyunghyun Cho.
\newblock Query-efficient imitation learning for end-to-end simulated driving.
\newblock In \emph{Proceedings of the Thirty-First AAAI Conference on Artificial Intelligence}, AAAI'17, pp.\  2891–2897. AAAI Press, 2017.

\end{thebibliography}
